%% file: SWARM_submission.tex
\documentclass[fleqn,10pt,twocolumn]{SWARM17}

\usepackage{algorithm}
\usepackage[noend]{algpseudocode}
\usepackage{amssymb,amsmath}
\usepackage{float}
\usepackage{graphicx}
\usepackage{subcaption}
\usepackage{tikz}
\usepackage{xcolor}
\usepackage{mathtools} 

\algnewcommand{\IfThenElse}[3]{
	\State \algorithmicif\ #1\ \algorithmicthen\ #2\\ \algorithmicelse\ #3}
\algnewcommand{\IfThen}[2]{
	\State \algorithmicif\ #1\ \algorithmicthen\ #2}

\def\sthree{1.732050808}

\title{Phototactic Supersmarticles}

\author{Sarah Cannon${}^1$, Joshua J.~Daymude${}^2$, William Savoie${}^3$, Ross Warkentin${}^3$, Shengkai Li${}^{3\dagger}$,\\ Daniel I.~Goldman${}^3$, Dana Randall${}^1$, and Andr\'{e}a W.~Richa${}^2$}

\speaker{Shengkai Li} 

\affils{${}^1$College of Computing, Georgia Institute of Technology, Atlanta, Georgia\\
(E-mail: sarah.cannon@gatech.edu, randall@cc.gatech.edu)\\
${}^2$Computer Science, CIDSE, Arizona State University, Tempe, Arizona\\
(E-mail: \{jdaymude,aricha\}@asu.edu)\\
${}^3$School of Physics, Georgia Institute of Technology, Atlanta, Georgia\\
(E-mail: \{wsavoie,rwarkentin3,shengkaili\}@gatech.edu, daniel.goldman@physics.gatech.edu)
}

\abstract{{\it Smarticles}, or {\it smart active particles}, are small robots equipped with only basic movement and sensing abilities that are incapable of rotating or displacing individually. We study the ensemble behavior of smarticles, i.e., the behavior a collective of these very simple computational elements can achieve, and how such behavior can be implemented using minimal programming. We show that an ensemble of smarticles constrained to remain close to one another (which we call a {\it supersmarticle}), achieves directed locomotion toward or away from a light source, a phenomenon known as {\it phototaxing}.
We present  experimental and theoretical models of phototactic supersmarticles that collectively move with a directed displacement in response to light.
The motion of the supersmarticle is approximately Brownian, and is a result of chaotic interactions among smarticles. The system can be directed by introducing asymmetries among the individual smarticle's behavior, in our case by varying activity levels in response to light, resulting in supersmarticle biased motion.
}


\begin{document}
\maketitle

\section{Introduction} \label{sec:intro}

In developing a system of {\it programmable matter}, one hopes to create a material or substance that can utilize user input or stimuli from its environment to change its physical properties in a programmable fashion. Many such systems have been proposed (e.g., smart materials, synthetic cells, and modular and swarm robotics) and each attempts to perform tasks subject to domain-specific capabilities and constraints. In this paper, we are interested in {\it active programmable matter}, where the energy input takes place directly at the scale of each active (matter) particle and allows for self-propelled movement\footnote{As opposed to passive programmable matter systems such as DNA computing and tile self-assembly.}
~\cite{Ramaswamy2010}. We investigate how such a system can achieve {\it directed locomotion}, wherein the individual particles move together as a collective in a desired direction.

Specifically, we consider active programmable matter ensembles composed of particles that individually are incapable of  locomotion. However, when constrained to remain in close proximity to other particles, we show that the overall ensemble can generate movement. Moreover, external stimuli  that  introduce asymmetries into the system with regards to individual particle activity can be used to produce a mode of directed displacement, either towards or away from a light source, known as {\it phototaxing}. We investigate this phenomenon through both experimental and theoretical models.

We show, in Section~\ref{sec:physical}, that phototaxing emerges in testbed experiments on a constrained collection of {\it smarticles} (that we call a {\it supersmarticle}). A smarticle is a small, 3-link, planar robot, developed by Goldman's group, equipped with sensing abilities but is incapable of rotating or displacing individually. A supersmarticle is a collection of smarticles enclosed by an unanchored rigid ring. One can think of a supersmarticle as a ``robot made of robots'' which achieves capabilities greater than any individual smarticle; phototaxing is one such capability. 

To investigate phototaxing from a theoretical perspective, we utilize previous work on {\it self-organizing particle systems}, that abstractly describes programmable matter as a collection of simple computational elements ({\it particles}) with limited memory that each execute fully distributed, local, and asynchronous algorithms to solve system-wide problems of movement, configuration, and coordination (e.g.,~\cite{Derakhshandeh2015}).
Recent work applying stochastic approaches to algorithms for self-organizing particle systems has yielded surprisingly fruitful results, producing local algorithms that are robust, nearly oblivious, and truly decentralized.
This approach was initially applied to develop an algorithm for {\it compression} in self-organizing particle systems under the assumptions of the geometric amoebot model~\cite{Cannon2016}. To solve the compression problem, particles gather as tightly together as possible, as in a sphere or its equivalent in the presence of some underlying geometry. This phenomenon is observed in natural systems (e.g., fire ants forming floating rafts~\cite{Mlot2011}).

\subsection{Our Results}
In this paper, we demonstrate how one can create a phototaxing particle ensemble by giving an algorithm for an abstract particle system under the amoebot model in which phototaxing is observed. It is achieved, rather remarkably, with just one very subtle modification to the compression algorithm: particles become more (or less) active when they sense light. In Section~\ref{sec:alg}, we formally prove that phototaxing occurs for systems of two and three particles; we also present simulation results of our algorithms for much larger systems that demonstrate the same behavior.  We note that in the amoebot model, unlike smarticles, individual particles are capable of movement, but this will be undirected regardless of how active they become in response to a light source.  In contrast, we show that groups of particles can  achieve directed displacement in response to light in the theoretical model, similar to the smarticles.

Both the physical and theoretical systems we consider have three components: (1) individual particles move regularly with no sense of direction, (2) there is a constraint ensuring the particles remain in close proximity to one another, and (3) particles' activity changes in response to light. In both cases, these basic requirements suffice to produce phototaxing. Perhaps the most surprising result is that phototaxing can be achieved without all particles knowing the direction of the light source; the occlusion of light by individual particles suffices for the ensemble as a whole to ``know'' where the light is and move accordingly, entirely via local distributed algorithms. We posit that more generally, many other systems with these three features should also be phototactic.

The remainder of this paper is organized as follows: In Section~\ref{sec:related}, we present a brief overview of related work. Section~\ref{sec:prelims}, we describe the physical smarticles and the theoretical abstractions that we will use in this paper. Section~\ref{sec:physical} presents our experimental testbed results on phototactic supersmarticles, which were the inspiration for the theoretical and simulation analysis on an abstraction of smarticle ensembles that we present  in Section~\ref{sec:alg}. We present our concluding remarks, including directions for future work, in Section~\ref{sec:conclude}.

\subsection{Related Work} \label{sec:related}

When examining the recently proposed and realized systems of programmable matter, one can distinguish between \emph{passive} (e.g., \cite{Cheung2011,Woods2013,Angluin2006})
and \emph{active} (e.g.,~\cite{Cieliebak2012,Rubenstein2014,Chazelle2009,Yim2007})
systems. Our work falls within the latter, which distinguishes itself from passive systems due to self-propelled motion at the particle level.
Examples of active programmable matter systems include \emph{swarm robotics}, various other models of modular robotics, and the \emph{amoebot model}, which defines our computational framework (detailed in Section~\ref{subsec:amoebot}).

Swarm robotics systems usually involve a collection of autonomous robots that move freely in space with limited sensing and communication ranges. These systems can perform a variety of tasks including gathering (e.g.,~\cite{Cieliebak2012}), shape formation (e.g.,~\cite{Rubenstein2014}), and imitating the collective behavior of natural systems (e.g.,~\cite{Chazelle2009}); however, the individual robots are more complex and have more powerful computational capabilities than those we consider.
\emph{Modular self-reconfigurable robotic systems} focus on motion planning and control of kinematic robots to achieve dynamic morphology~(e.g., \cite{Yim2007}).
The \emph{nubot model}~\cite{Woods2013-nubot}
seeks to provide a formal framework for rigorous algorithmic study of molecular programming systems.

In our physical experiments, our supersmarticles achieve phototaxing by changing the behavior of individual smarticles in response to light, making some of them inactive. We believe that the inactive smarticles can be approximated as a loose extension of the boundary, and one whose collision model is softer than the normal rigid boundary. This is consistent with previous work done with randomly diffusing self-propelled particles in~\cite{Dauchot2017,Kardar2015}. These studies investigated systems of self-propelled active particles enclosed in a boundary. The boundary's perimeter was divided into two sections, each composed of distinct materials, one half with a softer potential and the other half a more rigid potential. They found the applied pressure on the soft boundary from the particles was larger than on the more rigid boundary. We utilize this emergent response resulting from physical interactions, shown previously in simulation, in experiment to generate directed motion from a collection of individually non-motile robotic units.

\section{Preliminaries} \label{sec:prelims}

We begin by describing both the physical smarticles and the theoretical abstractions.

\subsection{Smarticles} \label{subsec:smarticles}

In order to explore emergent phenomena that result from collections of entities with limited mobility and sensing, we developed what we are calling ``smarticles.'' {\it Smarticles}, or {\it smart particles}, are small 14 x 2.5 x 3 cm robots which can change their shape in situ, but are incapable of rotating or displacing individually. Each smarticle is a three-link, two revolute joint, planar robot where only the center link is in contact with the ground.

Each smarticle consists of two Power HD-1440A MicroServos, a MEMS analog omnidirectional microphone, two photoresistors, a current sensing resistor, and a re-programmable Arduino Pro Mini 328-3.3V/8MHz, which handles the ADC and servo control. The two servos control the smarticle's two outer links, allowing the smarticle to fully explore its two-dimensional configuration space. The microphone, and pair of photoresistors, represent two channels through which we can send basic commands: using varying frequency ranges of sound or controlling levels of light. The current sensing resistor detects current draw from the servos, and thus the torque they are experiencing, allowing each smarticle to sense its own stress state. The links of the smarticles were 3D printed, ensuring uniform construction between all smarticles. Each smarticle is capable of performing predefined shape changes in the joint space. When we place a collection of smarticles placed inside an unanchored ring, we call this a supersmarticle.

\begin{figure}
	\centering
	\includegraphics[width=175pt]{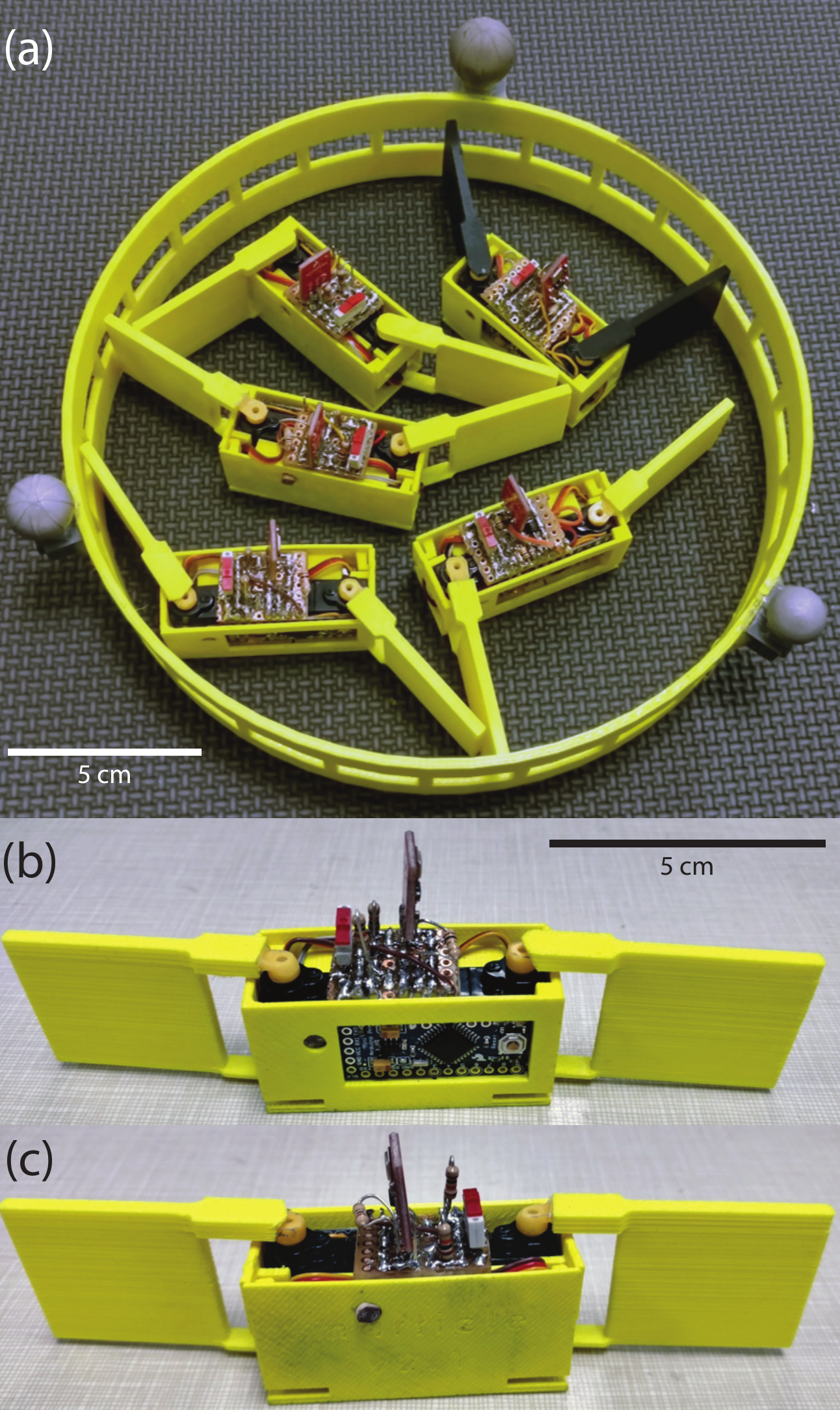}
	\caption{(a) A supersmarticle composed of 5 individual smarticles. A single smarticle, as viewed from the (b) front and (c) rear.}
	\label{fig:expPic}
\end{figure}

\subsection{The geometric amoebot model} \label{subsec:amoebot}

\begin{figure}
	\centering
	\begin{subfigure}{.45\columnwidth}
		\centering
		\begin{tikzpicture}[scale=0.6]
		\clip (0.5,-0.25) rectangle (5.5,3.25);
		\foreach \i in {0,...,10} \draw[black,line width=.5pt] (\i*\sthree / 2,-5)--(\i*\sthree / 2,5);
		\foreach \i in {-10,...,10}
		{
			\draw[black,line width=.5pt] (0,\i)--(5*\sthree,\i + 5);
			\draw[black,line width=.5pt] (0,\i)--(5*\sthree,\i - 5);
		}
		\foreach \i in {0,2,...,10}
		\foreach \j in {-5,...,5}
		\draw[fill] (\i*\sthree / 2,\j) circle (0.13);
		\foreach \i in {1,3,...,10}
		\foreach \j in {-4.5,...,4.5}
		\draw[fill] (\i*\sthree / 2,\j) circle (0.13);
		\end{tikzpicture}
		\caption{}
		\label{fig:modelgrid}
	\end{subfigure}%
	\begin{subfigure}{.45\columnwidth}
		\centering
		\includegraphics[scale=0.45]{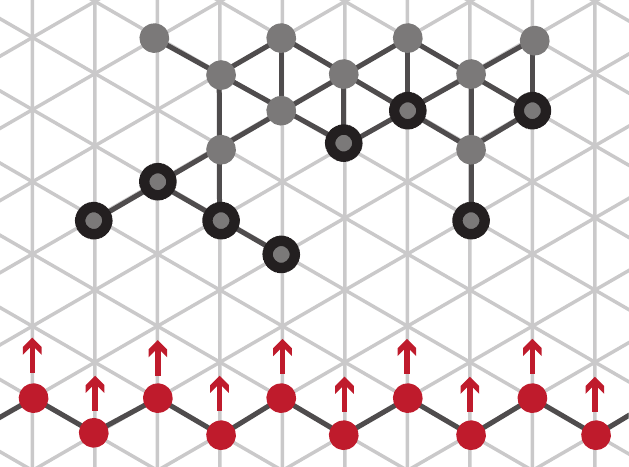}
		\caption{}
		\label{fig:modelheadlabel}
	\end{subfigure}
	\vspace{-3mm}
	\caption{(a) A section of the triangular lattice $\Gamma$. (b) An example particle system with some point light sources broadcasting upward along lattice lines; the particles that sense the light are outlined, while all others are occluded.}
	\label{fig:model}
\end{figure}
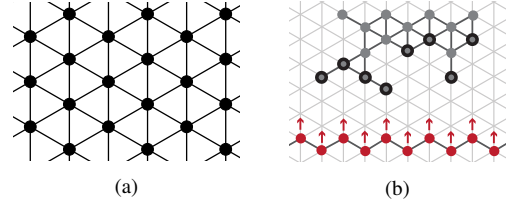

In order to study smarticle systems from a more formal perspective, we turn to self-organizing particle systems, which abstract away from specific instantiations of programmable matter to a more general model. This approach describes programmable matter as a collection of simple computational elements ({\it particles}) with limited memory that each execute fully distributed, local, asynchronous algorithms to solve system-wide problems of movement, configuration, and coordination.
In the {\it geometric amoebot model}, space is modeled as the infinite triangular lattice $\Gamma$ (Fig.~\ref{fig:model}a). Each particle occupies a distinct lattice point and can move along lattice edges. Each particle is anonymous, and there is no shared coordinate system or global sense of orientation. Particles interact only if they are {\it neighbors}, that is, if they occupy adjacent vertices of the lattice.
Every particle has a constant-size, local, memory which both it and its neighbors are able to read from and write to for communication. Due to the limitation of constant-size memory, particles cannot know the total number of particles in the system or any estimate of this quantity.
We assume that any conflicts (of movement or shared memory writes) are resolved arbitrarily.
Full model details can be found in~\cite{Cannon2016}.

For phototaxing, we furthermore assume that each particle can sense light, and that particles can occlude that light. More specifically, we consider point light sources that broadcast along rays in $\Gamma$. The light from a source is sensed only by the first particle in the lattice line along which the light is shining, and not by any other particles that may be in that lattice line (Fig.~\ref{fig:model}b).

\subsection{Compression} \label{subsec:compression}

Our local distributed algorithm for phototaxing under the assumptions of the geometric amoebot model uses the stochastic compression algorithm of Cannon et al.~\cite{Cannon2016} as a subroutine, so we present a high-level summary here. We assume particles start in a simply connected configuration, and  we design algorithms that ensure they stay simply connected.
Variants of the compression algorithm in~\cite{Cannon2016} produce a variety of other useful behaviors, including expansion over as wide an area as possible, coating an arbitrarily shaped surface, spanning fixed sites, and forming shortcut bridges~\cite{AndresArroyo2017} (a behavior also observed in army ants~\cite{Reid2015}); here, we show another variant produces phototaxing. For all of these problems, tools from Markov chain analysis and distributed algorithms allow us to relate local and global optimal behavior.

The stochastic algorithm in \cite{Cannon2016} achieves compression by favoring moves that increase the number of edges in the particle system configuration, where an {\it edge} of a configuration is an edge of $\Gamma$ where both endpoints are occupied by particles. Since the total number of particles stays fixed, maximizing the number of edges within a configuration is equivalent to minimizing the number of edges on the perimeter.
The compression algorithm takes as input a parameter $\lambda$ that controls how desirable it is for a particle to have neighbors, where $\lambda > 1$ favors configurations with more neighboring pairs of particles and thus more edges.
The distributed compression algorithm ensures the system converges to a distribution that favors having more edges using a {\it Metropolis filter}~\cite{Metropolis1953,Hastings1970}, a tool from Markov chain analysis that allows local probabilities of moves to be set so that global convergence to a certain distribution occurs. Specifically, our algorithms incorporate carefully chosen probabilities for particle moves so that the system converges to a stationary distribution $\pi$ over particle system configurations $\sigma$ where $\pi(\sigma) \sim \lambda^{e(\sigma)}$, where $e(\sigma)$ is the number of edges of configuration $\sigma$.
When $\lambda > 1$, this leads directly to distribution $\pi$ placing the most weight on configurations with the most edges, which provably are the most compressed configurations.
In particular, for any $\lambda > 2+\sqrt{2} \sim 3.42$, under $\pi$ all but an exponentially small fraction of particle configurations will be compressed.
This means the Markov chain, and thus the associated distributed local algorithm, converge to a distribution over particle configurations where with very high probability compression has been achieved.

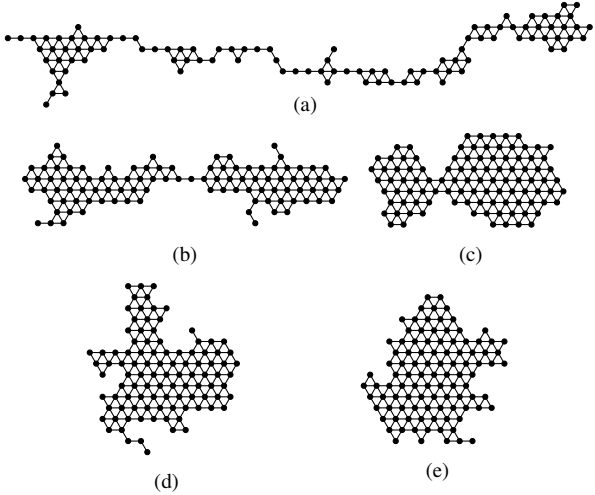
\begin{figure}[t]
	\centering
	\begin{subfigure}{\columnwidth}
		\centering
		\input{line100_bias4_1mil_rotated.txt} \vspace{-4mm}
		\caption{}
	\end{subfigure}\\[1ex]
	\begin{subfigure}[b]{.55\columnwidth}
		\centering
		\input{line100_bias4_2mil_rotated.txt} \vspace{-4mm}
		\caption{}
	\end{subfigure}%
	\begin{subfigure}[b]{.4\columnwidth}
		\centering
		\input{line100_bias4_3mil_rotated.txt}
		\caption{}
	\end{subfigure}\\[1ex]
	\begin{subfigure}{.45\columnwidth}
		\centering
		\input{line100_bias4_4mil_rotated.txt}
		\caption{}
	\end{subfigure}%
	\begin{subfigure}{.45\columnwidth}
		\centering
		\input{line100_bias4_5mil_rotated.txt}
		\caption{}
	\end{subfigure}
	\vspace{-2mm}
	\caption{The compression algorithm for 100 particles initially in a line after (a) 1 million, (b) 2 million, (c) 3million, (d) 4 million and (e) 5 million iterations of Markov chain $\mathcal{M}$ with bias $\lambda = 4$.}
	\label{fig:markovcomp}
\end{figure}

Algorithm~\ref{alg:particles-compression} is a simplified, high level description of the local distributed algorithm executed by each particle in order to achieve system-level compression~\cite{Cannon2016}; parameter $\lambda$, the input to the compression algorithm, is known by each particle.
A simulated asynchronous execution of this compression algorithm is shown in Fig.~\ref{fig:markovcomp}.

\begin{algorithm}
	\begin{algorithmic}[1]
		\State Let $\ell$ denote $P$'s current location; choose neighboring location $\ell'$ uniformly at random from the six possible choices in $\Gamma$.
		\If {$\ell'$ is unoccupied and certain local connectivity conditions hold in the neighborhood of $\ell \cup \ell'$}
		\State Generate a random number $q \in (0,1)$.
		\State Let $e$ be the number of other particles adjacent to location $\ell$ and $e'$ be the number adjacent to $\ell'$.
		\IfThen{$q < \lambda^{e' - e}$}{Move to $\ell'$.} \label{algstate:particles-compression-lambda}
		\Else{}  Remain at $\ell$.
		\EndIf
	\end{algorithmic}
	\caption{(Compression for particle $P$)}
	\label{alg:particles-compression}
\end{algorithm}

To analyze the limiting behavior of the algorithm, we assume each particle activates and executes Algorithm~\ref{alg:particles-compression} at a time drawn randomly from a Poisson distribution. 
This has the benefit of indirectly ensuring that our particle activations are fair, in the sense that for any particle $P$ and any time $t$, $P$ will always be activated at least once after $t$.
Further details regarding resolutions of the conflicts (of movement or shared memory writes) that arise when nearby particles are activated at nearly the same time are available in~\cite{Cannon2016}; most importantly, these efforts ensure that for the formal analysis, we may assume that at most one particle is active (performing a bounded amount of computation and at most one movement) at a time. This follows the standard asynchronous model of computation~\cite{lynch96}, which greatly simplifies analysis.
In particular, one can define the (centralized) Markov chain $\mathcal{M}$ associated with Algorithm~\ref{alg:particles-compression} as follows: $\mathcal{M}$ picks a particle 
uniformly at random
and then executes the steps of Algorithm~\ref{alg:particles-compression} for that particle. This enables the use of techniques from Markov chain analysis to prove guarantees about the behavior of the system when each particle is independently executing Algorithm~\ref{alg:particles-compression}; we now summarize those guarantees.


{\bf Theorem 1:}
Consider a self-organizing particle system under the geometric amoebot model where each particle individually executes Algorithm~\ref{alg:particles-compression} with some fixed $\lambda > 2+\sqrt{2}$. The particle system will always remain simply connected and will converge to a distribution over  configurations $\pi(\sigma) \sim \lambda^{e(\sigma)}$ where with all but exponentially small probability the  system is compressed.

\section{Physical Phototactic Supersmarticles} \label{sec:physical}

\begin{figure}
	\centering
	\includegraphics[width=125pt]{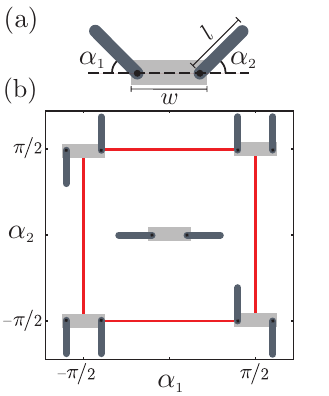}
	\vspace{-2mm}\caption{(a) Configuration space of a single smarticle defined by the angles $\alpha_1$ and $\alpha_2$ between the outer and inner links. (b) The square gait with certain configurations from the trajectory illustrated.}
	\label{fig:squareGait}
\end{figure}

In this section, we describe the supersmarticle displacement experiments and their results. For each experiment, we place the supersmarticle, i.e., the smarticles and ring, on a level plane and each smarticle performs a gait, where a gait is a closed periodic trajectory in the joint space of a smarticle. The smarticles used in the experiments were programmed to exhibit two behavioral states: one where the smarticle servos traced a square drawn in the 2-dimensional joint space as seen in Fig.~\ref{fig:squareGait}, called the active state, and another where the servos were held at a fixed position such that all links of the smarticle were parallel, called the inactive state. Smarticles will persist in the active state until either photoresistors, one found on either side of the smarticle, detect light above a certain threshold. When above the threshold the smarticle will persist in the inactive state until the light level sensed by either of its photoresistors drop below the threshold and will becomes active again. 

The experimental setup is planar, and hence the smarticle nearest to the light source occludes light from reaching other smarticles behind it inside the supersmarticle. Given the light sensor locations on the smarticle body and the geometry of the straight configuration, typically only one smarticle at a time is inactive, occluding the light and keeping the others smarticles below the photoresistor threshold. The occlusion of the light source effectively produces a light gradient across the supersmarticle which provides a decentralized, stigmergic communication method. Each smarticle's behavior is a response to the local environment, which in turn is a tends to affect the local environment of its neighbors.

\subsection{Experimental Methods} \label{subsec:methods}

Two types of experiments were performed: one type where all smarticles remained active and another type with both active and inactive smarticles. In the second experiment, one side of the system is illuminated with a light source, thereby forcing certain smarticles into the inactive state. All experiments were performed in a dark room, so that smarticles only entered the inactive state when subjected to the controlling light source input.

Experimental trials were initiated with the supersmarticle at the center of a $0.4m^2$ test plate and ended when the supersmarticle had translated to an edge of the test plane. When internal supersmarticle configurations exhibited slow displacement rates, trails were cropped at 10 minutes.

Multiple trials were taken with the light source at one of four locations to minimize systematic error. In each light experiment, one light source was placed at the center of an edge of the test plate. The light source was directed towards the nearest exposed photoresistor, thereby rendering a single smarticle within the system inactive.

Trajectories of the supersmarticle center of geometry were recorded using OptiTrack infrared video recording technology, and the data were exported and analyzed in MATLAB using a MSD analysis package~\cite{Tarantino2014}.

\subsection{Experimental Results and Discussion} \label{subsec:results}
\begin{figure}
	\centering
	\includegraphics[width=225pt]{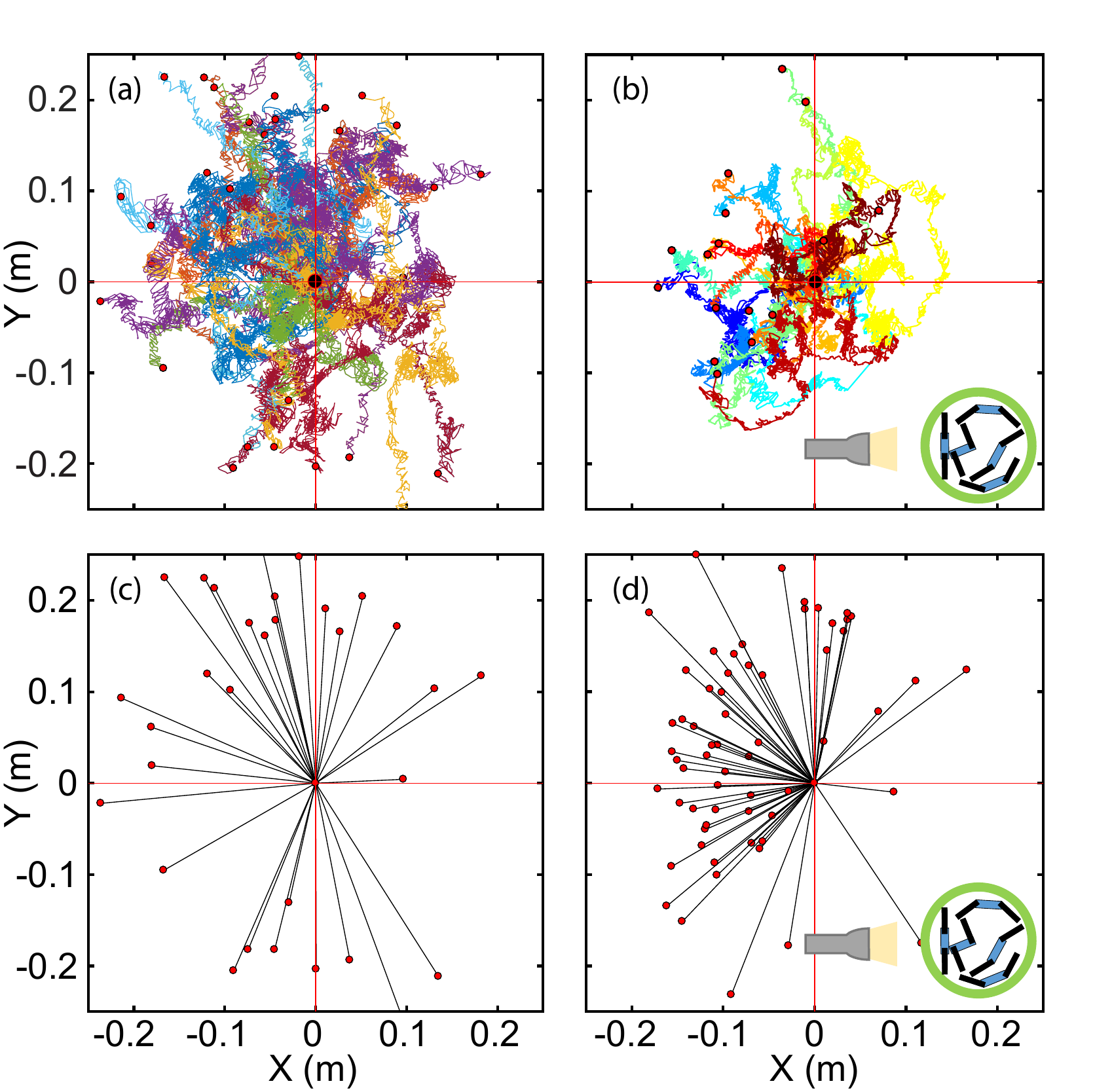}
	\caption{(a) and (b) are trajectories of the supersmarticle's center of geometry for a non-biased motion and light-biased motion respectively. Each colored trajectory represents a separate trial. (c) and (d) are shows lines connecting the initial and final position of the supersmarticle for each experiment. (d) contains all light biased directions data. Trials where the light was not originating from the +x direction were rotated to allow comparison between all trials. Illumination direction is shown via the location of the flashlight with respect to the supersmarticle image. All tracks begin from $(0,0)$ and end at the red circles.}
	\label{fig:expData}
\end{figure}

The supersmarticle's motion was dependent on the activity within the ring. Diffusive behavior was observed in both the control (Fig.~\ref{fig:expData}.(a,c)) and directed experiments (Fig.~\ref{fig:expData}.(b,d)), but the presence of inactive smarticles near the light source introduces a biased drift towards the light. The light-controlled supersmarticle system consistently diffused in the direction of the light source, with an average success rate of $82.3\pm 6.0\%$ across all trials.

Mean squared displacement (MSD) curves are useful for describing the types of diffusive behavior present in a given dynamic system \cite{Berg1983}. The MSD is defined as:
\begin{center}
	$\sigma^2 = \langle\vec{x}\cdot\vec{x}\rangle - \langle\vec{x}\rangle \cdot \langle\vec{x}\rangle = 4Dt^\gamma$
\end{center}
where $\gamma$ is the diffusion coefficient for the system. Free diffusive movement is seen for values of $\gamma = 1$, with $\gamma < 1$ characterizing subdiffusive behavior and $\gamma > 1$ characterizing superdiffusive behavior, or active transport. By fitting a line to the log-log plot of the MSD curve, the slope of the resulting fit will be the diffusion parameter $\gamma$.

The average MSD curve for each set of experimental data was computed, and a linear approximations were fit to the log-log plots. After performing out analyses across all data sets, the mean slope for fully active system was computed to be $0.99 \ m^{2}/s$ and the light-directed systems were $1.04 \pm 0.02 \ m^2/s$.  The application of the light-control algorithm results in a shift in diffusive behavior, from a purely diffusive system to a superdiffusive system where the active transport phenomenon causes the system to propagate towards the light source.

\section{A Phototactic Algorithm} \label{sec:alg}

To complement the physical experiments, we developed a local distributed algorithm for phototaxing in self-organizing particle systems under the abstract geometric amoebot model. We prove that the algorithm causes the system to diffuse in a certain direction in reaction to the light source when there are two or three particles, and we present simulations that demonstrate this same effect for larger particle systems.

We assume the particle system starts in some connected\footnote{The assumption of connectedness can be relaxed, but it simplifies the proofs while maintaining the phototaxing behavior we desire. We can think of connectivity and compression as playing a role analogous to that of the ring in the physical model.} initial configuration $\sigma_0$. For phototaxing to occur, we assume a collection of point light sources (sufficiently far from the particle configuration to not interfere with its motion) broadcast light along lattice lines in the same direction. Specifically, we assume the light sources form an infinite jagged line below all the particles and broadcast light upwards, as in Fig.~\ref{fig:model}b. We define the {\it height} of a particle system to be the $y$-coordinate of its center of mass, where all light sources are assumed to have $y$-coordinate $0$ or $-1/2$; we assume all edges of the triangular lattice are of length 1. We say that phototaxing occurs if there is some fixed number of iterations after which the height of the particle system has strictly increased or strictly decreased in expectation.

Our local distributed algorithm for phototaxing (specifically, for locomotion away from a light source) is remarkably simple; each particle executes Algorithm~\ref{alg:particles-phototaxing} when activated. The choice of $1/4$ in the algorithm is because it seems to work well in practice. Smaller values for that parameter can affect the compression algorithm's execution and cause different structural configurations to emerge, while larger values (that are still less than 1) correspond to even slower locomotion. Conflicts (of movement or shared memory writes) are resolved just as they are for compression; recall from Section~\ref{subsec:compression} this allows us to assume at most one particle is active at a time.

\begin{algorithm}
	\begin{algorithmic}[1]
		\If {$P$ senses light}
		\State $P$ Executes Algorithm~\ref{alg:particles-compression}.
		\Else
		\State $P$ Executes Algorithm~\ref{alg:particles-compression} with probability $\frac{1}{4}$.
		\EndIf
	\end{algorithmic}
	\caption{Phototaxing for a particle $P$}
	\label{alg:particles-phototaxing}
\end{algorithm}

So far, we assumed all particles activate at the same rate; under this assumption, we will see that the particle system achieves the desired phototaxing when all the particles independently execute Algorithm~\ref{alg:particles-phototaxing}. If instead we assume that it is possible for particles' activation rates to change in response to light, as is the case for the physical smarticles of Section~\ref{sec:physical}, then phototaxing can occur when each particle simply executes Algorithm~\ref{alg:particles-compression}. For instance, if particles are four times more likely to activate when exposed to light and each executes Algorithm~\ref{alg:particles-compression} upon activation, this system is equivalent to a system of particles with uniform activation rates executing Algorithm~\ref{alg:particles-phototaxing}.

\begin{figure}[t]
	\centering
	\begin{subfigure}{.45\columnwidth}
		\centering
		\includegraphics[scale=1.2]{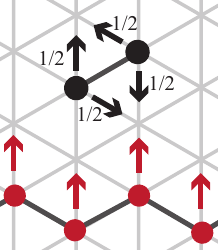}
		\caption{}
	\end{subfigure}%
	\begin{subfigure}{.45\columnwidth}
		\centering
		\includegraphics[scale=1.2]{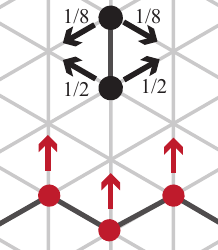}
		\caption{}
	\end{subfigure}
	\vspace{-3mm}\caption{A system of two particles, and the probabilities of each particle's movement if it is activated next. (a) If both particles are exposed to the light source, the expected change in height of the system after one iteration is 0. (b) If one particle occludes the other from the light source,  the expected change in the height of the system after one iteration is $+3/32$.}
	\label{fig:2particles}
\end{figure}

\subsection{Provable Phototaxing for Very Small Systems} \label{subsec:phototaxing-small}

Here, we formally verify the observed phototaxing  for very small systems (with 2 or 3 particles) by proving that when the particles independently execute Algorithm~\ref{alg:particles-phototaxing}, the  system exhibits a drift away from the light source.

We first consider a system of two particles, each activating at the same rate and then executing Algorithm~\ref{alg:particles-compression}. In this case, Algorithm~\ref{alg:particles-phototaxing} simplifies to Algorithm~\ref{alg:particles-phototaxing2}:

\begin{algorithm}
	\begin{algorithmic}[1]
		\State Choose one of the two locations adjacent to both particles uniformly at random; call it $\ell$.
		\IfThenElse{$P$ senses light}{Move to $\ell$}{Move to $\ell$ with probability $\frac{1}{4}$.}
	\end{algorithmic}
	\caption{Phototaxing for a particle $P$: 2 particles}
	\label{alg:particles-phototaxing2}
\end{algorithm}

{\bf Theorem 2:}
For  system of two particles each executing Algorithm~\ref{alg:particles-phototaxing2}, phototaxing occurs. 

\begin{proof}
	We show that after two particle activations, the expected height of the system has increased by at least $+3/64$, which implies the particle system is moving away from the light source.
	Up to translation and reflection, there are two possible states a system of two particles can be in: either both particles are exposed to the light ({\it State 1}), or one particle occludes the other from the light ({\it State 2}); see Fig.~\ref{fig:2particles}. Regardless of state, both particles are equally likely to activate next. In State 1, case analysis shows the expected change in height after one particle activation is 0. Furthermore, with probability $1/2$ the system remains in State 1 and with probability $1/2$ it enters State 2.
	For a particle system in State 2, with probability $1/2$ the occluded particle activates next, and with probability $1/4$ it moves a distance of $-1/2$ in the $y$-direction, causing the height of the system to decrease by $1/4$. With the remaining probability $1/2$, the particle exposed to light is activated and moves a distance of $+1/2$ in the $y$-direction, causing the height of the system to increase by $1/4$.
	Overall, in this case the expected change in the height of the system is:
	\begin{center}
		$\frac{1}{2} \cdot \frac{1}{4} \cdot \left(-\frac{1}{4}\right) + \frac{1}{2} \cdot \left(+\frac{1}{4}\right) = \frac{3}{32}$.
	\end{center}
	
	Beginning in State 1, we condition on the state of the system after one  activation and see that after two  activations the expected height of the system has increased by at least $3/64$. Beginning in State 2, after two particle activations the expected height of the system has increased by at least $3/32 > 3/64$.  This proves the theorem.
\end{proof}

Thus, for systems of two particles each executing Algorithm~\ref{alg:particles-phototaxing2} upon activation, the expected distance from the light sources strictly increases over time, meaning phototaxing provably occurs.

\begin{figure*}
	\centering
	\begin{subfigure}{.33\columnwidth}
		\centering
		\includegraphics[height = 3cm,keepaspectratio,page=1]{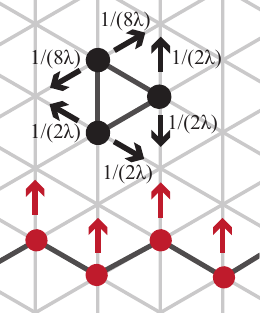}
		\caption{E$[\Delta h] = 0$}
	\end{subfigure}%
	\begin{subfigure}{.3\columnwidth}
		\centering
		\includegraphics[height = 3cm,keepaspectratio,page=6]{light_ex_3particles.pdf}
		\caption{E$[\Delta h] = 0$}
	\end{subfigure}%
	\begin{subfigure}{.3\columnwidth}
		\centering
		\includegraphics[height = 3cm,keepaspectratio,page=2]{light_ex_3particles.pdf}
		\caption{E$[\Delta h] = 0$}
	\end{subfigure}%
	\begin{subfigure}{.3\columnwidth}
		\centering
		\includegraphics[height = 3cm,keepaspectratio,page=7]{light_ex_3particles.pdf}
		\caption{E$[\Delta h] = 0$}
	\end{subfigure}%
	\begin{subfigure}{.28\columnwidth}
		\centering
		\includegraphics[height = 3cm,keepaspectratio,page=3]{light_ex_3particles.pdf}
		\caption{E$[\Delta h] = \frac{1}{48}$}
	\end{subfigure}%
	\begin{subfigure}{.28\columnwidth}
		\centering 
		\includegraphics[height = 3cm,keepaspectratio,page=4]{light_ex_3particles.pdf}
		\caption{E$[\Delta h] = \frac{1}{24}$}
	\end{subfigure}%
	\begin{subfigure}{.28\columnwidth}
		\centering
		\includegraphics[height = 3cm,keepaspectratio,page=5]{light_ex_3particles.pdf}
		\caption{E$[\Delta h] = \frac{1}{24}$}
	\end{subfigure}
	\vspace{-1mm}\caption{The seven possible states for a system of three particles (up to reflection and translation) and the probabilities of each particle's movement if it is activated next; the expected change in the height of the system after one iteration beginning in each of the seven states is also shown.}
	\label{fig:3particles}
\end{figure*}

The same result  holds for systems of three particles, albeit with a slightly slower drift. For systems with exactly three particles, Algorithm~\ref{alg:particles-phototaxing} simplifies to Algorithm~\ref{alg:particles-phototaxing3}, below. Note that the compression bias parameter $\lambda$ and the movement probability filter based on the number of edges in the system (Step~\ref{algstate:particles-compression-lambda} of Algorithm~\ref{alg:particles-compression}) begin to play a role.

\begin{algorithm}
	\begin{algorithmic}[1]
		\State Determine possible valid locations to move to, of which there are at most 2.
		\State For each such location, set move probability to $1/2$.
		\If {move decreases number of edges in system}
		\State Divide move probability by $\lambda$.
		\EndIf
		\If {$P$ does not sense light}
		\State Divide each move probability by $4$.
		\EndIf
		\State Move to a possible valid location with the corresponding move probability; with all remaining probability, don't move.
	\end{algorithmic}
	\caption{Phototaxing for a particle $P$: 3 particles}
	\label{alg:particles-phototaxing3}
\end{algorithm}

{\bf Theorem 3:}
For a system of three particles each executing Algorithm~\ref{alg:particles-phototaxing3} with bias parameter $\lambda > 2+\sqrt{2}$, phototaxing occurs.

\begin{proof}
	We show that after three particle activations the expected height of the system has increased by at least $1/(64\lambda)$.
	Up to translation and reflection, there are seven possible states the particle system could be in; all are shown in Fig.~\ref{fig:3particles}. Doing a case analysis just as for two particles, we see that the expected change in the height of the system after one particle activation is nonnegative in all seven states. For states (e,f,g), the expected increase in height after one particle activation is more than $1/(64\lambda)$, and as expected height is nondecreasing the same holds after three particle activations. For the states (a,b,c,d), the expected change in height after one particle activation is zero, so we consider multiple particle activations at a time. For state (a), after one particle activation there is a positive probability it is in state (e) or state (g), and we can use conditional expectation to calculate that after two particle activations, beginning in state (a), the expected increase in the height of the system is $1/(64\lambda)$. Similarly, beginning in state (b), after two particle activations the expected change in the height of the system is  $1/96$; because $\lambda > 2+\sqrt{2}$, that is, that we are in the regime of compression, then $1/96 > 1/(64\lambda)$.
	
	Beginning in states (c) and (d), it takes at least two particle moves to reach a state where there is a positive expected increase in height after the next particle activation; each reaches state (e) after two particle activations with probability $1/18 + 1/(9\lambda)$ and state (g) after two particle activations with probability $1/18 + 5/(72 \lambda)$. The total expected increase in height after three particle activations starting in either state (c) or state (d) is:
	\begin{center}
		$\left(\frac{1}{18} + \frac{1}{9\lambda} \right)  \frac{1}{48} + \left(\frac{1}{18} + \frac{5}{72\lambda} \right)  \frac{1}{24}= \frac{1}{288} + \frac{1}{192\lambda} > \frac{1}{64\lambda}$,
	\end{center}
	since $\lambda > 3$. Thus, for all possible states the system is in, we have shown that after three particle activations, the height of the system has strictly increased in expectation by at least $1/(64\lambda)$.
\end{proof}

\subsection{Phototaxing Simulations for Larger Systems} \label{subsec:phototaxing-large}

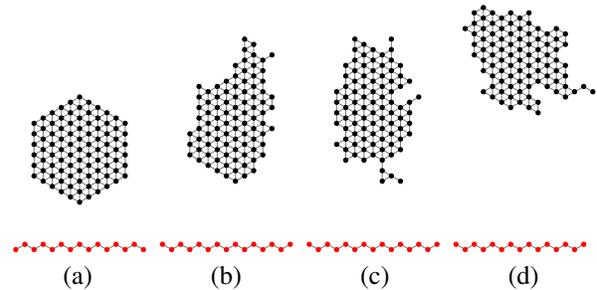
\begin{figure}
	\centering
	\input{hexagonlight_0_modified_cropped.txt}
	\input{hexagonlight_10000000_modified_cropped.txt}
	\input{hexagonlight_20000000_modified_cropped.txt} 
	\input{hexagonlight_30000000_modified_cropped.txt}
	\\ (a) \hspace{14mm} (b) \hspace{14mm} (c) \hspace{14mm} (d)
	\vspace{-3mm}\caption{An execution of Algorithm~\ref{alg:particles-phototaxing} with $\lambda = 4$ for a system of 91 particles, with light sources that shine upwards shown in red, after (a) 0, (b) 10 million, (c) 20 million, and (d) 30 million iterations. Multiple executions all exhibit a drift upwards, as seen here.}
	\label{fig:hexagonlight}
\end{figure}

Algorithm~\ref{alg:particles-phototaxing} can be used to achieve phototaxing for arbitrarily large systems of particles, not just systems with two or three particles.
Simulations for a system with 91 particles can be seen in Fig.~\ref{fig:hexagonlight}. Though the motion is largely random, it is clear there is a general trend away from the light sources. This drift was consistent across all simulations of Algorithm~\ref{alg:particles-phototaxing}.
In all simulations, including the one shown in Fig.~\ref{fig:hexagonlight}, the particle system also exhibited lateral drift of varying magnitude and direction; that drift is not shown in Fig.~\ref{fig:hexagonlight} due to space constraints.

\section{Conclusion} \label{sec:conclude}

This study presented the use of physical and simulated atomic agents incapable of directed motion in confined active matter systems which exhibit locomotion on the collective scale. Moreover, the responses of the individuals of the system to external fields were used to~introduce asymmetries in the system, producing biased locomotion.

Robophysical studies of the supersmarticle system were demonstrated to probabilistically favor the direction of the inactive smarticle, though the physics which drives this behavior has yet to be fully explored. Future work will probe the underlying system dynamics to refine and develop a more comprehensive understanding of the system interactions between active and inactive particles which generate biased locomotion. Physical variables such as the masses of the particles and the confining ring and the friction coefficients are expected to modulate the diffusive properties of the system. Additionally, the interaction behaviors of the particles as they move through their joint space trajectories may lead to various system modes of oscillations characterized by hysteretic displacement loops which lead to the biased locomotion seen in our research.

These physical features will be explored by developing a reduced 1D model of the supersmarticle system in which particles and the confining ring will be restricted to movement along a linear track. This new 1D system will be studied experimentally and through physics-based simulations, with the intent of sweeping the physical and interaction parameter space in order to identify the governing variables which characterize the system dynamics and produce biased locomotion.

We plan to continue to complement the experimental robophysical extensions with rigorous algorithmic studies of the systems to provide a better understanding on how to program collections of smarticles 
to achieve the desired collective behavior.  Here, we extended a known algorithm by changing particles' probabilities of movement; while the simplicity of this approach is one of its strengths, new algorithmic ideas and approaches could provide further insights into phototaxing behavior.

{\bf Acknowledgements:}
S. Cannon: This material is based upon work supported by the National Science Foundation Graduate
Research Fellowship Program under Grant DGE-1148903. 

J. J. Daymude and A. W. Richa: Supported in part by NSF CCF-1422603, CCF-1637393, and CCF-1733680.

D. I. Goldman: Funding provided by NSF PoLS \#0957659 and \#PHY-1205878, and ARO \#W911NF-13-1-0347. 

D. Randall: Supported in part by  NSF CCF-1526900, CCF-1637031, and CCF-1733812.


\bibliographystyle{plain}
\bibliography{ref}



\end{document}

%% file: line100_bias4_1mil_rotated.txt
\begin{tikzpicture}[scale=0.17,rotate=30]
   \draw[fill](0,22.5)circle(0.2);
    \draw(0,22.5)--(0.866025,22);
    \draw[fill](17.3205,10.5)circle(0.2);
    \draw(17.3205,10.5)--(17.3205,11.5);
    \draw[fill](4.33013,20)circle(0.2);
    \draw(4.33013,20)--(5.19615,19.5);
    \draw(4.33013,20)--(5.19615,20.5);
    \draw[fill](2.59808,21)circle(0.2);
    \draw(2.59808,21)--(3.4641,20.5);
    \draw[fill](12.1244,13.5)circle(0.2);
    \draw(12.1244,13.5)--(12.9904,13);
    \draw(12.1244,13.5)--(12.1244,14.5);
    \draw[fill](1.73205,20.5)circle(0.2);
    \draw(1.73205,20.5)--(2.59808,20);
    \draw(1.73205,20.5)--(2.59808,21);
    \draw(1.73205,20.5)--(1.73205,21.5);
    \draw[fill](25.9808,3.5)circle(0.2);
    \draw(25.9808,3.5)--(26.8468,4);
    \draw(25.9808,3.5)--(25.9808,4.5);
    \draw[fill](1.73205,16.5)circle(0.2);
    \draw(1.73205,16.5)--(1.73205,17.5);
    \draw[fill](16.4545,12)circle(0.2);
    \draw(16.4545,12)--(17.3205,11.5);
    \draw[fill](1.73205,21.5)circle(0.2);
    \draw(1.73205,21.5)--(2.59808,21);
    \draw[fill](5.19615,19.5)circle(0.2);
    \draw(5.19615,19.5)--(6.06218,19);
    \draw(5.19615,19.5)--(5.19615,20.5);
    \draw[fill](3.4641,20.5)circle(0.2);
    \draw(3.4641,20.5)--(4.33013,20);
    \draw[fill](6.9282,18.5)circle(0.2);
    \draw(6.9282,18.5)--(7.79423,18);
    \draw[fill](5.19615,18.5)circle(0.2);
    \draw(5.19615,18.5)--(6.06218,18);
    \draw(5.19615,18.5)--(6.06218,19);
    \draw(5.19615,18.5)--(5.19615,19.5);
    \draw[fill](0,16.5)circle(0.2);
    \draw(0,16.5)--(0.866025,17);
    \draw[fill](2.59808,20)circle(0.2);
    \draw(2.59808,20)--(3.4641,19.5);
    \draw(2.59808,20)--(3.4641,20.5);
    \draw(2.59808,20)--(2.59808,21);
    \draw[fill](7.79423,18)circle(0.2);
    \draw(7.79423,18)--(8.66025,17.5);
    \draw[fill](0.866025,17)circle(0.2);
    \draw(0.866025,17)--(1.73205,16.5);
    \draw(0.866025,17)--(1.73205,17.5);
    \draw[fill](2.59808,18)circle(0.2);
    \draw(2.59808,18)--(3.4641,18.5);
    \draw(2.59808,18)--(2.59808,19);
    \draw[fill](5.19615,20.5)circle(0.2);
    \draw[fill](10.3923,14.5)circle(0.2);
    \draw(10.3923,14.5)--(11.2583,14);
    \draw(10.3923,14.5)--(11.2583,15);
    \draw(10.3923,14.5)--(10.3923,15.5);
    \draw[fill](3.4641,18.5)circle(0.2);
    \draw(3.4641,18.5)--(4.33013,18);
    \draw(3.4641,18.5)--(4.33013,19);
    \draw(3.4641,18.5)--(3.4641,19.5);
    \draw[fill](1.73205,17.5)circle(0.2);
    \draw(1.73205,17.5)--(2.59808,18);
    \draw(1.73205,17.5)--(1.73205,18.5);
    \draw[fill](20.7846,7.5)circle(0.2);
    \draw(20.7846,7.5)--(21.6506,7);
    \draw(20.7846,7.5)--(20.7846,8.5);
    \draw[fill](2.59808,19)circle(0.2);
    \draw(2.59808,19)--(3.4641,18.5);
    \draw(2.59808,19)--(3.4641,19.5);
    \draw(2.59808,19)--(2.59808,20);
    \draw[fill](1.73205,18.5)circle(0.2);
    \draw(1.73205,18.5)--(2.59808,18);
    \draw(1.73205,18.5)--(2.59808,19);
    \draw(1.73205,18.5)--(1.73205,19.5);
    \draw[fill](9.52628,16)circle(0.2);
    \draw(9.52628,16)--(10.3923,15.5);
    \draw[fill](12.1244,14.5)circle(0.2);
    \draw[fill](19.0526,8.5)circle(0.2);
    \draw(19.0526,8.5)--(19.9186,8);
    \draw[fill](3.4641,19.5)circle(0.2);
    \draw(3.4641,19.5)--(4.33013,19);
    \draw(3.4641,19.5)--(4.33013,20);
    \draw(3.4641,19.5)--(3.4641,20.5);
    \draw[fill](14.7224,13)circle(0.2);
    \draw(14.7224,13)--(15.5885,12.5);
    \draw[fill](8.66025,17.5)circle(0.2);
    \draw[fill](0.866025,22)circle(0.2);
    \draw(0.866025,22)--(1.73205,21.5);
    \draw[fill](4.33013,18)circle(0.2);
    \draw(4.33013,18)--(5.19615,18.5);
    \draw(4.33013,18)--(4.33013,19);
    \draw[fill](4.33013,19)circle(0.2);
    \draw(4.33013,19)--(5.19615,18.5);
    \draw(4.33013,19)--(5.19615,19.5);
    \draw(4.33013,19)--(4.33013,20);
    \draw[fill](1.73205,19.5)circle(0.2);
    \draw(1.73205,19.5)--(2.59808,19);
    \draw(1.73205,19.5)--(2.59808,20);
    \draw(1.73205,19.5)--(1.73205,20.5);
    \draw[fill](8.66025,16.5)circle(0.2);
    \draw(8.66025,16.5)--(9.52628,16);
    \draw(8.66025,16.5)--(8.66025,17.5);
    \draw[fill](15.5885,12.5)circle(0.2);
    \draw(15.5885,12.5)--(16.4545,12);
    \draw[fill](19.9186,7)circle(0.2);
    \draw(19.9186,7)--(20.7846,7.5);
    \draw(19.9186,7)--(19.9186,8);
    \draw[fill](18.1865,9)circle(0.2);
    \draw(18.1865,9)--(19.0526,8.5);
    \draw[fill](6.06218,18)circle(0.2);
    \draw(6.06218,18)--(6.9282,18.5);
    \draw(6.06218,18)--(6.06218,19);
    \draw[fill](24.2487,4.5)circle(0.2);
    \draw(24.2487,4.5)--(25.1147,4);
    \draw(24.2487,4.5)--(24.2487,5.5);
    \draw[fill](10.3923,15.5)circle(0.2);
    \draw(10.3923,15.5)--(11.2583,15);
    \draw[fill](10.3923,13.5)circle(0.2);
    \draw(10.3923,13.5)--(11.2583,14);
    \draw(10.3923,13.5)--(10.3923,14.5);
    \draw[fill](36.3731,3.5)circle(0.2);
    \draw(36.3731,3.5)--(37.2391,3);
    \draw[fill](17.3205,11.5)circle(0.2);
    \draw[fill](20.7846,8.5)circle(0.2);
    \draw(20.7846,8.5)--(21.6506,9);
    \draw[fill](6.06218,19)circle(0.2);
    \draw(6.06218,19)--(6.9282,18.5);
    \draw[fill](12.9904,13)circle(0.2);
    \draw(12.9904,13)--(13.8564,13.5);
    \draw[fill](30.3109,3)circle(0.2);
    \draw(30.3109,3)--(30.3109,4);
    \draw[fill](22.5167,6.5)circle(0.2);
    \draw(22.5167,6.5)--(23.3827,6);
    \draw[fill](11.2583,15)circle(0.2);
    \draw(11.2583,15)--(12.1244,14.5);
    \draw[fill](13.8564,13.5)circle(0.2);
    \draw(13.8564,13.5)--(14.7224,13);
    \draw[fill](36.3731,0.5)circle(0.2);
    \draw(36.3731,0.5)--(37.2391,0);
    \draw(36.3731,0.5)--(37.2391,1);
    \draw(36.3731,0.5)--(36.3731,1.5);
    \draw[fill](27.7128,3.5)circle(0.2);
    \draw(27.7128,3.5)--(28.5788,3);
    \draw(27.7128,3.5)--(28.5788,4);
    \draw[fill](24.2487,5.5)circle(0.2);
    \draw[fill](33.775,4)circle(0.2);
    \draw(33.775,4)--(34.641,3.5);
    \draw(33.775,4)--(34.641,4.5);
    \draw[fill](37.2391,2)circle(0.2);
    \draw(37.2391,2)--(38.1051,1.5);
    \draw(37.2391,2)--(38.1051,2.5);
    \draw(37.2391,2)--(37.2391,3);
    \draw[fill](23.3827,5)circle(0.2);
    \draw(23.3827,5)--(24.2487,4.5);
    \draw(23.3827,5)--(24.2487,5.5);
    \draw(23.3827,5)--(23.3827,6);
    \draw[fill](36.3731,2.5)circle(0.2);
    \draw(36.3731,2.5)--(37.2391,2);
    \draw(36.3731,2.5)--(37.2391,3);
    \draw(36.3731,2.5)--(36.3731,3.5);
    \draw[fill](38.9711,0)circle(0.2);
    \draw(38.9711,0)--(39.8372,0.5);
    \draw(38.9711,0)--(38.9711,1);
    \draw[fill](38.1051,0.5)circle(0.2);
    \draw(38.1051,0.5)--(38.9711,0);
    \draw(38.1051,0.5)--(38.9711,1);
    \draw(38.1051,0.5)--(38.1051,1.5);
    \draw[fill](30.3109,4)circle(0.2);
    \draw(30.3109,4)--(31.1769,4.5);
    \draw[fill](28.5788,3)circle(0.2);
    \draw(28.5788,3)--(29.4449,2.5);
    \draw(28.5788,3)--(29.4449,3.5);
    \draw(28.5788,3)--(28.5788,4);
    \draw[fill](34.641,3.5)circle(0.2);
    \draw(34.641,3.5)--(35.507,3);
    \draw(34.641,3.5)--(34.641,4.5);
    \draw[fill](19.9186,8)circle(0.2);
    \draw(19.9186,8)--(20.7846,7.5);
    \draw(19.9186,8)--(20.7846,8.5);
    \draw[fill](11.2583,14)circle(0.2);
    \draw(11.2583,14)--(12.1244,13.5);
    \draw(11.2583,14)--(12.1244,14.5);
    \draw(11.2583,14)--(11.2583,15);
    \draw[fill](21.6506,7)circle(0.2);
    \draw(21.6506,7)--(22.5167,6.5);
    \draw[fill](14.7224,12)circle(0.2);
    \draw(14.7224,12)--(15.5885,12.5);
    \draw(14.7224,12)--(14.7224,13);
    \draw[fill](27.7128,2.5)circle(0.2);
    \draw(27.7128,2.5)--(28.5788,3);
    \draw(27.7128,2.5)--(27.7128,3.5);
    \draw[fill](38.9711,2)circle(0.2);
    \draw(38.9711,2)--(39.8372,2.5);
    \draw(38.9711,2)--(38.9711,3);
    \draw[fill](21.6506,9)circle(0.2);
    \draw[fill](34.641,2.5)circle(0.2);
    \draw(34.641,2.5)--(35.507,2);
    \draw(34.641,2.5)--(35.507,3);
    \draw(34.641,2.5)--(34.641,3.5);
    \draw[fill](25.9808,4.5)circle(0.2);
    \draw(25.9808,4.5)--(26.8468,4);
    \draw[fill](34.641,4.5)circle(0.2);
    \draw[fill](38.1051,1.5)circle(0.2);
    \draw(38.1051,1.5)--(38.9711,1);
    \draw(38.1051,1.5)--(38.9711,2);
    \draw(38.1051,1.5)--(38.1051,2.5);
    \draw[fill](37.2391,3)circle(0.2);
    \draw(37.2391,3)--(38.1051,2.5);
    \draw[fill](29.4449,2.5)circle(0.2);
    \draw(29.4449,2.5)--(30.3109,3);
    \draw(29.4449,2.5)--(29.4449,3.5);
    \draw[fill](32.0429,5)circle(0.2);
    \draw(32.0429,5)--(32.909,4.5);
    \draw[fill](32.909,4.5)circle(0.2);
    \draw(32.909,4.5)--(33.775,4);
    \draw[fill](31.1769,4.5)circle(0.2);
    \draw(31.1769,4.5)--(32.0429,4);
    \draw(31.1769,4.5)--(32.0429,5);
    \draw[fill](17.3205,9.5)circle(0.2);
    \draw(17.3205,9.5)--(18.1865,9);
    \draw(17.3205,9.5)--(17.3205,10.5);
    \draw[fill](26.8468,4)circle(0.2);
    \draw(26.8468,4)--(27.7128,3.5);
    \draw[fill](38.1051,2.5)circle(0.2);
    \draw(38.1051,2.5)--(38.9711,2);
    \draw(38.1051,2.5)--(38.9711,3);
    \draw[fill](38.9711,3)circle(0.2);
    \draw(38.9711,3)--(39.8372,2.5);
    \draw[fill](39.8372,2.5)circle(0.2);
    \draw[fill](22.5167,5.5)circle(0.2);
    \draw(22.5167,5.5)--(23.3827,5);
    \draw(22.5167,5.5)--(23.3827,6);
    \draw(22.5167,5.5)--(22.5167,6.5);
    \draw[fill](25.1147,4)circle(0.2);
    \draw(25.1147,4)--(25.9808,3.5);
    \draw(25.1147,4)--(25.9808,4.5);
    \draw[fill](28.5788,4)circle(0.2);
    \draw(28.5788,4)--(29.4449,3.5);
    \draw[fill](35.507,2)circle(0.2);
    \draw(35.507,2)--(36.3731,1.5);
    \draw(35.507,2)--(36.3731,2.5);
    \draw(35.507,2)--(35.507,3);
    \draw[fill](32.0429,4)circle(0.2);
    \draw(32.0429,4)--(32.909,3.5);
    \draw(32.0429,4)--(32.909,4.5);
    \draw(32.0429,4)--(32.0429,5);
    \draw[fill](29.4449,3.5)circle(0.2);
    \draw(29.4449,3.5)--(30.3109,3);
    \draw(29.4449,3.5)--(30.3109,4);
    \draw[fill](35.507,3)circle(0.2);
    \draw(35.507,3)--(36.3731,2.5);
    \draw(35.507,3)--(36.3731,3.5);
    \draw[fill](36.3731,1.5)circle(0.2);
    \draw(36.3731,1.5)--(37.2391,1);
    \draw(36.3731,1.5)--(37.2391,2);
    \draw(36.3731,1.5)--(36.3731,2.5);
    \draw[fill](38.9711,1)circle(0.2);
    \draw(38.9711,1)--(39.8372,0.5);
    \draw(38.9711,1)--(38.9711,2);
    \draw[fill](23.3827,6)circle(0.2);
    \draw(23.3827,6)--(24.2487,5.5);
    \draw[fill](37.2391,1)circle(0.2);
    \draw(37.2391,1)--(38.1051,0.5);
    \draw(37.2391,1)--(38.1051,1.5);
    \draw(37.2391,1)--(37.2391,2);
    \draw[fill](32.909,3.5)circle(0.2);
    \draw(32.909,3.5)--(33.775,4);
    \draw(32.909,3.5)--(32.909,4.5);
    \draw[fill](39.8372,0.5)circle(0.2);
    \draw[fill](37.2391,0)circle(0.2);
    \draw(37.2391,0)--(38.1051,0.5);
    \draw(37.2391,0)--(37.2391,1);
\end{tikzpicture}

%% file: line100_bias4_2mil_rotated.txt
\begin{tikzpicture}[scale=0.17,rotate=30]
    \draw[fill](2.59808,9.5)circle(0.2);
    \draw(2.59808,9.5)--(3.4641,9);
    \draw(2.59808,9.5)--(3.4641,10);
    \draw(2.59808,9.5)--(2.59808,10.5);
    \draw[fill](1.73205,11)circle(0.2);
    \draw(1.73205,11)--(2.59808,10.5);
    \draw(1.73205,11)--(2.59808,11.5);
    \draw(1.73205,11)--(1.73205,12);
    \draw[fill](16.4545,2.5)circle(0.2);
    \draw(16.4545,2.5)--(17.3205,2);
    \draw(16.4545,2.5)--(17.3205,3);
    \draw(16.4545,2.5)--(16.4545,3.5);
    \draw[fill](4.33013,11.5)circle(0.2);
    \draw(4.33013,11.5)--(5.19615,11);
    \draw(4.33013,11.5)--(5.19615,12);
    \draw(4.33013,11.5)--(4.33013,12.5);
    \draw[fill](18.1865,3.5)circle(0.2);
    \draw(18.1865,3.5)--(19.0526,3);
    \draw(18.1865,3.5)--(19.0526,4);
    \draw(18.1865,3.5)--(18.1865,4.5);
    \draw[fill](21.6506,2.5)circle(0.2);
    \draw[fill](20.7846,0)circle(0.2);
    \draw(20.7846,0)--(21.6506,0.5);
    \draw(20.7846,0)--(20.7846,1);
    \draw[fill](1.73205,10)circle(0.2);
    \draw(1.73205,10)--(2.59808,9.5);
    \draw(1.73205,10)--(2.59808,10.5);
    \draw(1.73205,10)--(1.73205,11);
    \draw[fill](3.4641,13)circle(0.2);
    \draw(3.4641,13)--(4.33013,12.5);
    \draw(3.4641,13)--(4.33013,13.5);
    \draw(3.4641,13)--(3.4641,14);
    \draw[fill](10.3923,8)circle(0.2);
    \draw(10.3923,8)--(11.2583,7.5);
    \draw(10.3923,8)--(11.2583,8.5);
    \draw(10.3923,8)--(10.3923,9);
    \draw[fill](11.2583,8.5)circle(0.2);
    \draw[fill](8.66025,9)circle(0.2);
    \draw(8.66025,9)--(9.52628,8.5);
    \draw(8.66025,9)--(9.52628,9.5);
    \draw(8.66025,9)--(8.66025,10);
    \draw[fill](4.33013,13.5)circle(0.2);
    \draw(4.33013,13.5)--(4.33013,14.5);
    \draw[fill](6.06218,9.5)circle(0.2);
    \draw(6.06218,9.5)--(6.9282,9);
    \draw(6.06218,9.5)--(6.9282,10);
    \draw(6.06218,9.5)--(6.06218,10.5);
    \draw[fill](3.4641,10)circle(0.2);
    \draw(3.4641,10)--(4.33013,9.5);
    \draw(3.4641,10)--(4.33013,10.5);
    \draw(3.4641,10)--(3.4641,11);
    \draw[fill](2.59808,12.5)circle(0.2);
    \draw(2.59808,12.5)--(3.4641,12);
    \draw(2.59808,12.5)--(3.4641,13);
    \draw(2.59808,12.5)--(2.59808,13.5);
    \draw[fill](3.4641,12)circle(0.2);
    \draw(3.4641,12)--(4.33013,11.5);
    \draw(3.4641,12)--(4.33013,12.5);
    \draw(3.4641,12)--(3.4641,13);
    \draw[fill](2.59808,11.5)circle(0.2);
    \draw(2.59808,11.5)--(3.4641,11);
    \draw(2.59808,11.5)--(3.4641,12);
    \draw(2.59808,11.5)--(2.59808,12.5);
    \draw[fill](3.4641,14)circle(0.2);
    \draw(3.4641,14)--(4.33013,13.5);
    \draw(3.4641,14)--(4.33013,14.5);
    \draw[fill](13.8564,6)circle(0.2);
    \draw(13.8564,6)--(14.7224,5.5);
    \draw(13.8564,6)--(14.7224,6.5);
    \draw(13.8564,6)--(13.8564,7);
    \draw[fill](4.33013,10.5)circle(0.2);
    \draw(4.33013,10.5)--(5.19615,10);
    \draw(4.33013,10.5)--(5.19615,11);
    \draw(4.33013,10.5)--(4.33013,11.5);
    \draw[fill](15.5885,6)circle(0.2);
    \draw(15.5885,6)--(16.4545,5.5);
    \draw(15.5885,6)--(15.5885,7);
    \draw[fill](4.33013,9.5)circle(0.2);
    \draw(4.33013,9.5)--(5.19615,9);
    \draw(4.33013,9.5)--(5.19615,10);
    \draw(4.33013,9.5)--(4.33013,10.5);
    \draw[fill](7.79423,7.5)circle(0.2);
    \draw(7.79423,7.5)--(8.66025,8);
    \draw(7.79423,7.5)--(7.79423,8.5);
    \draw[fill](18.1865,1.5)circle(0.2);
    \draw(18.1865,1.5)--(19.0526,1);
    \draw(18.1865,1.5)--(19.0526,2);
    \draw(18.1865,1.5)--(18.1865,2.5);
    \draw[fill](8.66025,8)circle(0.2);
    \draw(8.66025,8)--(9.52628,8.5);
    \draw(8.66025,8)--(8.66025,9);
    \draw[fill](11.2583,7.5)circle(0.2);
    \draw(11.2583,7.5)--(12.1244,7);
    \draw(11.2583,7.5)--(11.2583,8.5);
    \draw[fill](21.6506,1.5)circle(0.2);
    \draw(21.6506,1.5)--(22.5167,1);
    \draw(21.6506,1.5)--(21.6506,2.5);
    \draw[fill](6.9282,10)circle(0.2);
    \draw(6.9282,10)--(7.79423,9.5);
    \draw(6.9282,10)--(6.9282,11);
    \draw[fill](4.33013,14.5)circle(0.2);
    \draw[fill](1.73205,13)circle(0.2);
    \draw(1.73205,13)--(2.59808,12.5);
    \draw(1.73205,13)--(2.59808,13.5);
    \draw(1.73205,13)--(1.73205,14);
    \draw[fill](19.0526,4)circle(0.2);
    \draw(19.0526,4)--(19.9186,3.5);
    \draw(19.0526,4)--(19.0526,5);
    \draw[fill](5.19615,12)circle(0.2);
    \draw[fill](3.4641,9)circle(0.2);
    \draw(3.4641,9)--(4.33013,9.5);
    \draw(3.4641,9)--(3.4641,10);
    \draw[fill](2.59808,10.5)circle(0.2);
    \draw(2.59808,10.5)--(3.4641,10);
    \draw(2.59808,10.5)--(3.4641,11);
    \draw(2.59808,10.5)--(2.59808,11.5);
    \draw[fill](5.19615,11)circle(0.2);
    \draw(5.19615,11)--(6.06218,10.5);
    \draw(5.19615,11)--(5.19615,12);
    \draw[fill](0,10)circle(0.2);
    \draw(0,10)--(0.866025,9.5);
    \draw[fill](16.4545,5.5)circle(0.2);
    \draw(16.4545,5.5)--(17.3205,5);
    \draw[fill](1.73205,12)circle(0.2);
    \draw(1.73205,12)--(2.59808,11.5);
    \draw(1.73205,12)--(2.59808,12.5);
    \draw(1.73205,12)--(1.73205,13);
    \draw[fill](20.7846,1)circle(0.2);
    \draw(20.7846,1)--(21.6506,0.5);
    \draw(20.7846,1)--(21.6506,1.5);
    \draw(20.7846,1)--(20.7846,2);
    \draw[fill](12.9904,6.5)circle(0.2);
    \draw(12.9904,6.5)--(13.8564,6);
    \draw(12.9904,6.5)--(13.8564,7);
    \draw[fill](4.33013,12.5)circle(0.2);
    \draw(4.33013,12.5)--(5.19615,12);
    \draw(4.33013,12.5)--(4.33013,13.5);
    \draw[fill](5.19615,10)circle(0.2);
    \draw(5.19615,10)--(6.06218,9.5);
    \draw(5.19615,10)--(6.06218,10.5);
    \draw(5.19615,10)--(5.19615,11);
    \draw[fill](2.59808,13.5)circle(0.2);
    \draw(2.59808,13.5)--(3.4641,13);
    \draw(2.59808,13.5)--(3.4641,14);
    \draw[fill](17.3205,5)circle(0.2);
    \draw(17.3205,5)--(18.1865,4.5);
    \draw[fill](7.79423,8.5)circle(0.2);
    \draw(7.79423,8.5)--(8.66025,8);
    \draw(7.79423,8.5)--(8.66025,9);
    \draw(7.79423,8.5)--(7.79423,9.5);
    \draw[fill](10.3923,10)circle(0.2);
    \draw[fill](5.19615,9)circle(0.2);
    \draw(5.19615,9)--(6.06218,8.5);
    \draw(5.19615,9)--(6.06218,9.5);
    \draw(5.19615,9)--(5.19615,10);
    \draw[fill](15.5885,3)circle(0.2);
    \draw(15.5885,3)--(16.4545,2.5);
    \draw(15.5885,3)--(16.4545,3.5);
    \draw(15.5885,3)--(15.5885,4);
    \draw[fill](1.73205,9)circle(0.2);
    \draw(1.73205,9)--(2.59808,9.5);
    \draw(1.73205,9)--(1.73205,10);
    \draw[fill](14.7224,6.5)circle(0.2);
    \draw(14.7224,6.5)--(15.5885,6);
    \draw(14.7224,6.5)--(15.5885,7);
    \draw(14.7224,6.5)--(14.7224,7.5);
    \draw[fill](0.866025,12.5)circle(0.2);
    \draw(0.866025,12.5)--(1.73205,12);
    \draw(0.866025,12.5)--(1.73205,13);
    \draw(0.866025,12.5)--(0.866025,13.5);
    \draw[fill](0.866025,13.5)circle(0.2);
    \draw(0.866025,13.5)--(1.73205,13);
    \draw(0.866025,13.5)--(1.73205,14);
    \draw[fill](21.6506,0.5)circle(0.2);
    \draw(21.6506,0.5)--(22.5167,1);
    \draw(21.6506,0.5)--(21.6506,1.5);
    \draw[fill](15.5885,4)circle(0.2);
    \draw(15.5885,4)--(16.4545,3.5);
    \draw(15.5885,4)--(16.4545,4.5);
    \draw(15.5885,4)--(15.5885,5);
    \draw[fill](20.7846,2)circle(0.2);
    \draw(20.7846,2)--(21.6506,1.5);
    \draw(20.7846,2)--(21.6506,2.5);
    \draw(20.7846,2)--(20.7846,3);
    \draw[fill](14.7224,4.5)circle(0.2);
    \draw(14.7224,4.5)--(15.5885,4);
    \draw(14.7224,4.5)--(15.5885,5);
    \draw(14.7224,4.5)--(14.7224,5.5);
    \draw[fill](19.9186,0.5)circle(0.2);
    \draw(19.9186,0.5)--(20.7846,0);
    \draw(19.9186,0.5)--(20.7846,1);
    \draw(19.9186,0.5)--(19.9186,1.5);
    \draw[fill](6.06218,8.5)circle(0.2);
    \draw(6.06218,8.5)--(6.9282,8);
    \draw(6.06218,8.5)--(6.9282,9);
    \draw(6.06218,8.5)--(6.06218,9.5);
    \draw[fill](19.0526,3)circle(0.2);
    \draw(19.0526,3)--(19.9186,2.5);
    \draw(19.0526,3)--(19.9186,3.5);
    \draw(19.0526,3)--(19.0526,4);
    \draw[fill](15.5885,5)circle(0.2);
    \draw(15.5885,5)--(16.4545,4.5);
    \draw(15.5885,5)--(16.4545,5.5);
    \draw(15.5885,5)--(15.5885,6);
    \draw[fill](17.3205,4)circle(0.2);
    \draw(17.3205,4)--(18.1865,3.5);
    \draw(17.3205,4)--(18.1865,4.5);
    \draw(17.3205,4)--(17.3205,5);
    \draw[fill](17.3205,1)circle(0.2);
    \draw(17.3205,1)--(18.1865,1.5);
    \draw(17.3205,1)--(17.3205,2);
    \draw[fill](17.3205,3)circle(0.2);
    \draw(17.3205,3)--(18.1865,2.5);
    \draw(17.3205,3)--(18.1865,3.5);
    \draw(17.3205,3)--(17.3205,4);
    \draw[fill](22.5167,1)circle(0.2);
    \draw[fill](6.9282,8)circle(0.2);
    \draw(6.9282,8)--(7.79423,7.5);
    \draw(6.9282,8)--(7.79423,8.5);
    \draw(6.9282,8)--(6.9282,9);
    \draw[fill](20.7846,3)circle(0.2);
    \draw(20.7846,3)--(21.6506,2.5);
    \draw[fill](16.4545,3.5)circle(0.2);
    \draw(16.4545,3.5)--(17.3205,3);
    \draw(16.4545,3.5)--(17.3205,4);
    \draw(16.4545,3.5)--(16.4545,4.5);
    \draw[fill](19.0526,5)circle(0.2);
    \draw(19.0526,5)--(19.0526,6);
    \draw[fill](10.3923,9)circle(0.2);
    \draw(10.3923,9)--(11.2583,8.5);
    \draw(10.3923,9)--(10.3923,10);
    \draw[fill](6.9282,9)circle(0.2);
    \draw(6.9282,9)--(7.79423,8.5);
    \draw(6.9282,9)--(7.79423,9.5);
    \draw(6.9282,9)--(6.9282,10);
    \draw[fill](14.7224,2.5)circle(0.2);
    \draw(14.7224,2.5)--(15.5885,3);
    \draw[fill](14.7224,7.5)circle(0.2);
    \draw(14.7224,7.5)--(15.5885,7);
    \draw[fill](12.9904,5.5)circle(0.2);
    \draw(12.9904,5.5)--(13.8564,5);
    \draw(12.9904,5.5)--(13.8564,6);
    \draw(12.9904,5.5)--(12.9904,6.5);
    \draw[fill](15.5885,7)circle(0.2);
    \draw[fill](19.9186,1.5)circle(0.2);
    \draw(19.9186,1.5)--(20.7846,1);
    \draw(19.9186,1.5)--(20.7846,2);
    \draw(19.9186,1.5)--(19.9186,2.5);
    \draw[fill](19.0526,1)circle(0.2);
    \draw(19.0526,1)--(19.9186,0.5);
    \draw(19.0526,1)--(19.9186,1.5);
    \draw(19.0526,1)--(19.0526,2);
    \draw[fill](18.1865,4.5)circle(0.2);
    \draw(18.1865,4.5)--(19.0526,4);
    \draw(18.1865,4.5)--(19.0526,5);
    \draw[fill](13.8564,7)circle(0.2);
    \draw(13.8564,7)--(14.7224,6.5);
    \draw(13.8564,7)--(14.7224,7.5);
    \draw[fill](16.4545,1.5)circle(0.2);
    \draw(16.4545,1.5)--(17.3205,1);
    \draw(16.4545,1.5)--(17.3205,2);
    \draw(16.4545,1.5)--(16.4545,2.5);
    \draw[fill](12.1244,7)circle(0.2);
    \draw(12.1244,7)--(12.9904,6.5);
    \draw[fill](1.73205,14)circle(0.2);
    \draw(1.73205,14)--(2.59808,13.5);
    \draw[fill](9.52628,9.5)circle(0.2);
    \draw(9.52628,9.5)--(10.3923,9);
    \draw(9.52628,9.5)--(10.3923,10);
    \draw[fill](14.7224,1.5)circle(0.2);
    \draw(14.7224,1.5)--(14.7224,2.5);
    \draw[fill](3.4641,11)circle(0.2);
    \draw(3.4641,11)--(4.33013,10.5);
    \draw(3.4641,11)--(4.33013,11.5);
    \draw(3.4641,11)--(3.4641,12);
    \draw[fill](19.0526,2)circle(0.2);
    \draw(19.0526,2)--(19.9186,1.5);
    \draw(19.0526,2)--(19.9186,2.5);
    \draw(19.0526,2)--(19.0526,3);
    \draw[fill](13.8564,5)circle(0.2);
    \draw(13.8564,5)--(14.7224,4.5);
    \draw(13.8564,5)--(14.7224,5.5);
    \draw(13.8564,5)--(13.8564,6);
    \draw[fill](0.866025,9.5)circle(0.2);
    \draw(0.866025,9.5)--(1.73205,9);
    \draw(0.866025,9.5)--(1.73205,10);
    \draw[fill](7.79423,9.5)circle(0.2);
    \draw(7.79423,9.5)--(8.66025,9);
    \draw(7.79423,9.5)--(8.66025,10);
    \draw[fill](8.66025,10)circle(0.2);
    \draw(8.66025,10)--(9.52628,9.5);
    \draw[fill](6.06218,10.5)circle(0.2);
    \draw(6.06218,10.5)--(6.9282,10);
    \draw(6.06218,10.5)--(6.9282,11);
    \draw[fill](19.9186,2.5)circle(0.2);
    \draw(19.9186,2.5)--(20.7846,2);
    \draw(19.9186,2.5)--(20.7846,3);
    \draw(19.9186,2.5)--(19.9186,3.5);
    \draw[fill](6.9282,11)circle(0.2);
    \draw[fill](17.3205,2)circle(0.2);
    \draw(17.3205,2)--(18.1865,1.5);
    \draw(17.3205,2)--(18.1865,2.5);
    \draw(17.3205,2)--(17.3205,3);
    \draw[fill](16.4545,4.5)circle(0.2);
    \draw(16.4545,4.5)--(17.3205,4);
    \draw(16.4545,4.5)--(17.3205,5);
    \draw(16.4545,4.5)--(16.4545,5.5);
    \draw[fill](19.0526,6)circle(0.2);
    \draw[fill](9.52628,8.5)circle(0.2);
    \draw(9.52628,8.5)--(10.3923,8);
    \draw(9.52628,8.5)--(10.3923,9);
    \draw(9.52628,8.5)--(9.52628,9.5);
    \draw[fill](14.7224,5.5)circle(0.2);
    \draw(14.7224,5.5)--(15.5885,5);
    \draw(14.7224,5.5)--(15.5885,6);
    \draw(14.7224,5.5)--(14.7224,6.5);
    \draw[fill](18.1865,2.5)circle(0.2);
    \draw(18.1865,2.5)--(19.0526,2);
    \draw(18.1865,2.5)--(19.0526,3);
    \draw(18.1865,2.5)--(18.1865,3.5);
    \draw[fill](19.9186,3.5)circle(0.2);
    \draw(19.9186,3.5)--(20.7846,3);
\end{tikzpicture}

%% file: line100_bias4_3mil_rotated.txt
\begin{tikzpicture}[scale=0.17,rotate=30]
    \draw[fill](1.73205,7.5)circle(0.2);
    \draw(1.73205,7.5)--(2.59808,7);
    \draw(1.73205,7.5)--(2.59808,8);
    \draw(1.73205,7.5)--(1.73205,8.5);
    \draw[fill](11.2583,3)circle(0.2);
    \draw(11.2583,3)--(12.1244,2.5);
    \draw(11.2583,3)--(12.1244,3.5);
    \draw(11.2583,3)--(11.2583,4);
    \draw[fill](10.3923,0.5)circle(0.2);
    \draw(10.3923,0.5)--(11.2583,1);
    \draw(10.3923,0.5)--(10.3923,1.5);
    \draw[fill](8.66025,6.5)circle(0.2);
    \draw(8.66025,6.5)--(9.52628,6);
    \draw(8.66025,6.5)--(9.52628,7);
    \draw(8.66025,6.5)--(8.66025,7.5);
    \draw[fill](3.4641,5.5)circle(0.2);
    \draw(3.4641,5.5)--(4.33013,6);
    \draw(3.4641,5.5)--(3.4641,6.5);
    \draw[fill](9.52628,1)circle(0.2);
    \draw(9.52628,1)--(10.3923,0.5);
    \draw(9.52628,1)--(10.3923,1.5);
    \draw(9.52628,1)--(9.52628,2);
    \draw[fill](4.33013,8)circle(0.2);
    \draw(4.33013,8)--(4.33013,9);
    \draw[fill](12.1244,1.5)circle(0.2);
    \draw(12.1244,1.5)--(12.9904,1);
    \draw(12.1244,1.5)--(12.9904,2);
    \draw(12.1244,1.5)--(12.1244,2.5);
    \draw[fill](9.52628,6)circle(0.2);
    \draw(9.52628,6)--(10.3923,5.5);
    \draw(9.52628,6)--(10.3923,6.5);
    \draw(9.52628,6)--(9.52628,7);
    \draw[fill](0.866025,6)circle(0.2);
    \draw(0.866025,6)--(1.73205,5.5);
    \draw(0.866025,6)--(1.73205,6.5);
    \draw(0.866025,6)--(0.866025,7);
    \draw[fill](7.79423,3)circle(0.2);
    \draw(7.79423,3)--(8.66025,2.5);
    \draw(7.79423,3)--(8.66025,3.5);
    \draw(7.79423,3)--(7.79423,4);
    \draw[fill](3.4641,8.5)circle(0.2);
    \draw(3.4641,8.5)--(4.33013,8);
    \draw(3.4641,8.5)--(4.33013,9);
    \draw(3.4641,8.5)--(3.4641,9.5);
    \draw[fill](3.4641,9.5)circle(0.2);
    \draw(3.4641,9.5)--(4.33013,9);
    \draw(3.4641,9.5)--(4.33013,10);
    \draw(3.4641,9.5)--(3.4641,10.5);
    \draw[fill](10.3923,4.5)circle(0.2);
    \draw(10.3923,4.5)--(11.2583,4);
    \draw(10.3923,4.5)--(11.2583,5);
    \draw(10.3923,4.5)--(10.3923,5.5);
    \draw[fill](1.73205,8.5)circle(0.2);
    \draw(1.73205,8.5)--(2.59808,8);
    \draw(1.73205,8.5)--(2.59808,9);
    \draw(1.73205,8.5)--(1.73205,9.5);
    \draw[fill](3.4641,10.5)circle(0.2);
    \draw(3.4641,10.5)--(4.33013,10);
    \draw[fill](9.52628,5)circle(0.2);
    \draw(9.52628,5)--(10.3923,4.5);
    \draw(9.52628,5)--(10.3923,5.5);
    \draw(9.52628,5)--(9.52628,6);
    \draw[fill](10.3923,7.5)circle(0.2);
    \draw(10.3923,7.5)--(11.2583,7);
    \draw[fill](9.52628,8)circle(0.2);
    \draw(9.52628,8)--(10.3923,7.5);
    \draw[fill](13.8564,4.5)circle(0.2);
    \draw[fill](4.33013,6)circle(0.2);
    \draw(4.33013,6)--(5.19615,5.5);
    \draw(4.33013,6)--(5.19615,6.5);
    \draw(4.33013,6)--(4.33013,7);
    \draw[fill](7.79423,7)circle(0.2);
    \draw(7.79423,7)--(8.66025,6.5);
    \draw(7.79423,7)--(8.66025,7.5);
    \draw(7.79423,7)--(7.79423,8);
    \draw[fill](3.4641,7.5)circle(0.2);
    \draw(3.4641,7.5)--(4.33013,7);
    \draw(3.4641,7.5)--(4.33013,8);
    \draw(3.4641,7.5)--(3.4641,8.5);
    \draw[fill](8.66025,1.5)circle(0.2);
    \draw(8.66025,1.5)--(9.52628,1);
    \draw(8.66025,1.5)--(9.52628,2);
    \draw(8.66025,1.5)--(8.66025,2.5);
    \draw[fill](6.9282,5.5)circle(0.2);
    \draw(6.9282,5.5)--(7.79423,5);
    \draw(6.9282,5.5)--(7.79423,6);
    \draw(6.9282,5.5)--(6.9282,6.5);
    \draw[fill](6.9282,2.5)circle(0.2);
    \draw(6.9282,2.5)--(7.79423,2);
    \draw(6.9282,2.5)--(7.79423,3);
    \draw(6.9282,2.5)--(6.9282,3.5);
    \draw[fill](6.9282,1.5)circle(0.2);
    \draw(6.9282,1.5)--(7.79423,1);
    \draw(6.9282,1.5)--(7.79423,2);
    \draw(6.9282,1.5)--(6.9282,2.5);
    \draw[fill](11.2583,7)circle(0.2);
    \draw(11.2583,7)--(12.1244,6.5);
    \draw[fill](1.73205,10.5)circle(0.2);
    \draw(1.73205,10.5)--(2.59808,10);
    \draw[fill](10.3923,3.5)circle(0.2);
    \draw(10.3923,3.5)--(11.2583,3);
    \draw(10.3923,3.5)--(11.2583,4);
    \draw(10.3923,3.5)--(10.3923,4.5);
    \draw[fill](5.19615,4.5)circle(0.2);
    \draw(5.19615,4.5)--(6.06218,4);
    \draw(5.19615,4.5)--(6.06218,5);
    \draw(5.19615,4.5)--(5.19615,5.5);
    \draw[fill](6.06218,7)circle(0.2);
    \draw(6.06218,7)--(6.9282,6.5);
    \draw(6.06218,7)--(6.9282,7.5);
    \draw[fill](2.59808,10)circle(0.2);
    \draw(2.59808,10)--(3.4641,9.5);
    \draw(2.59808,10)--(3.4641,10.5);
    \draw[fill](12.1244,5.5)circle(0.2);
    \draw(12.1244,5.5)--(12.9904,5);
    \draw(12.1244,5.5)--(12.1244,6.5);
    \draw[fill](7.79423,5)circle(0.2);
    \draw(7.79423,5)--(8.66025,4.5);
    \draw(7.79423,5)--(8.66025,5.5);
    \draw(7.79423,5)--(7.79423,6);
    \draw[fill](2.59808,8)circle(0.2);
    \draw(2.59808,8)--(3.4641,7.5);
    \draw(2.59808,8)--(3.4641,8.5);
    \draw(2.59808,8)--(2.59808,9);
    \draw[fill](2.59808,6)circle(0.2);
    \draw(2.59808,6)--(3.4641,5.5);
    \draw(2.59808,6)--(3.4641,6.5);
    \draw(2.59808,6)--(2.59808,7);
    \draw[fill](12.1244,2.5)circle(0.2);
    \draw(12.1244,2.5)--(12.9904,2);
    \draw(12.1244,2.5)--(12.9904,3);
    \draw(12.1244,2.5)--(12.1244,3.5);
    \draw[fill](8.66025,4.5)circle(0.2);
    \draw(8.66025,4.5)--(9.52628,4);
    \draw(8.66025,4.5)--(9.52628,5);
    \draw(8.66025,4.5)--(8.66025,5.5);
    \draw[fill](12.9904,5)circle(0.2);
    \draw(12.9904,5)--(13.8564,4.5);
    \draw[fill](10.3923,1.5)circle(0.2);
    \draw(10.3923,1.5)--(11.2583,1);
    \draw(10.3923,1.5)--(11.2583,2);
    \draw(10.3923,1.5)--(10.3923,2.5);
    \draw[fill](0,6.5)circle(0.2);
    \draw(0,6.5)--(0.866025,6);
    \draw(0,6.5)--(0.866025,7);
    \draw[fill](0.866025,8)circle(0.2);
    \draw(0.866025,8)--(1.73205,7.5);
    \draw(0.866025,8)--(1.73205,8.5);
    \draw(0.866025,8)--(0.866025,9);
    \draw[fill](1.73205,6.5)circle(0.2);
    \draw(1.73205,6.5)--(2.59808,6);
    \draw(1.73205,6.5)--(2.59808,7);
    \draw(1.73205,6.5)--(1.73205,7.5);
    \draw[fill](6.9282,4.5)circle(0.2);
    \draw(6.9282,4.5)--(7.79423,4);
    \draw(6.9282,4.5)--(7.79423,5);
    \draw(6.9282,4.5)--(6.9282,5.5);
    \draw[fill](8.66025,5.5)circle(0.2);
    \draw(8.66025,5.5)--(9.52628,5);
    \draw(8.66025,5.5)--(9.52628,6);
    \draw(8.66025,5.5)--(8.66025,6.5);
    \draw[fill](0.866025,9)circle(0.2);
    \draw(0.866025,9)--(1.73205,8.5);
    \draw(0.866025,9)--(1.73205,9.5);
    \draw(0.866025,9)--(0.866025,10);
    \draw[fill](8.66025,3.5)circle(0.2);
    \draw(8.66025,3.5)--(9.52628,3);
    \draw(8.66025,3.5)--(9.52628,4);
    \draw(8.66025,3.5)--(8.66025,4.5);
    \draw[fill](12.9904,2)circle(0.2);
    \draw(12.9904,2)--(12.9904,3);
    \draw[fill](3.4641,6.5)circle(0.2);
    \draw(3.4641,6.5)--(4.33013,6);
    \draw(3.4641,6.5)--(4.33013,7);
    \draw(3.4641,6.5)--(3.4641,7.5);
    \draw[fill](10.3923,5.5)circle(0.2);
    \draw(10.3923,5.5)--(11.2583,5);
    \draw(10.3923,5.5)--(11.2583,6);
    \draw(10.3923,5.5)--(10.3923,6.5);
    \draw[fill](6.06218,3)circle(0.2);
    \draw(6.06218,3)--(6.9282,2.5);
    \draw(6.06218,3)--(6.9282,3.5);
    \draw(6.06218,3)--(6.06218,4);
    \draw[fill](5.19615,6.5)circle(0.2);
    \draw(5.19615,6.5)--(6.06218,6);
    \draw(5.19615,6.5)--(6.06218,7);
    \draw[fill](6.9282,6.5)circle(0.2);
    \draw(6.9282,6.5)--(7.79423,6);
    \draw(6.9282,6.5)--(7.79423,7);
    \draw(6.9282,6.5)--(6.9282,7.5);
    \draw[fill](9.52628,0)circle(0.2);
    \draw(9.52628,0)--(10.3923,0.5);
    \draw(9.52628,0)--(9.52628,1);
    \draw[fill](11.2583,1)circle(0.2);
    \draw(11.2583,1)--(12.1244,0.5);
    \draw(11.2583,1)--(12.1244,1.5);
    \draw(11.2583,1)--(11.2583,2);
    \draw[fill](2.59808,5)circle(0.2);
    \draw(2.59808,5)--(3.4641,5.5);
    \draw(2.59808,5)--(2.59808,6);
    \draw[fill](0.866025,5)circle(0.2);
    \draw(0.866025,5)--(1.73205,5.5);
    \draw(0.866025,5)--(0.866025,6);
    \draw[fill](0,5.5)circle(0.2);
    \draw(0,5.5)--(0.866025,5);
    \draw(0,5.5)--(0.866025,6);
    \draw(0,5.5)--(0,6.5);
    \draw[fill](7.79423,2)circle(0.2);
    \draw(7.79423,2)--(8.66025,1.5);
    \draw(7.79423,2)--(8.66025,2.5);
    \draw(7.79423,2)--(7.79423,3);
    \draw[fill](12.1244,3.5)circle(0.2);
    \draw(12.1244,3.5)--(12.9904,3);
    \draw(12.1244,3.5)--(12.9904,4);
    \draw(12.1244,3.5)--(12.1244,4.5);
    \draw[fill](4.33013,7)circle(0.2);
    \draw(4.33013,7)--(5.19615,6.5);
    \draw(4.33013,7)--(4.33013,8);
    \draw[fill](12.1244,0.5)circle(0.2);
    \draw(12.1244,0.5)--(12.9904,1);
    \draw(12.1244,0.5)--(12.1244,1.5);
    \draw[fill](4.33013,10)circle(0.2);
    \draw[fill](9.52628,4)circle(0.2);
    \draw(9.52628,4)--(10.3923,3.5);
    \draw(9.52628,4)--(10.3923,4.5);
    \draw(9.52628,4)--(9.52628,5);
    \draw[fill](1.73205,5.5)circle(0.2);
    \draw(1.73205,5.5)--(2.59808,5);
    \draw(1.73205,5.5)--(2.59808,6);
    \draw(1.73205,5.5)--(1.73205,6.5);
    \draw[fill](6.06218,5)circle(0.2);
    \draw(6.06218,5)--(6.9282,4.5);
    \draw(6.06218,5)--(6.9282,5.5);
    \draw(6.06218,5)--(6.06218,6);
    \draw[fill](0.866025,10)circle(0.2);
    \draw(0.866025,10)--(1.73205,9.5);
    \draw(0.866025,10)--(1.73205,10.5);
    \draw[fill](8.66025,2.5)circle(0.2);
    \draw(8.66025,2.5)--(9.52628,2);
    \draw(8.66025,2.5)--(9.52628,3);
    \draw(8.66025,2.5)--(8.66025,3.5);
    \draw[fill](0.866025,7)circle(0.2);
    \draw(0.866025,7)--(1.73205,6.5);
    \draw(0.866025,7)--(1.73205,7.5);
    \draw(0.866025,7)--(0.866025,8);
    \draw[fill](12.9904,3)circle(0.2);
    \draw(12.9904,3)--(12.9904,4);
    \draw[fill](12.1244,4.5)circle(0.2);
    \draw(12.1244,4.5)--(12.9904,4);
    \draw(12.1244,4.5)--(12.9904,5);
    \draw(12.1244,4.5)--(12.1244,5.5);
    \draw[fill](10.3923,6.5)circle(0.2);
    \draw(10.3923,6.5)--(11.2583,6);
    \draw(10.3923,6.5)--(11.2583,7);
    \draw(10.3923,6.5)--(10.3923,7.5);
    \draw[fill](12.1244,6.5)circle(0.2);
    \draw[fill](9.52628,3)circle(0.2);
    \draw(9.52628,3)--(10.3923,2.5);
    \draw(9.52628,3)--(10.3923,3.5);
    \draw(9.52628,3)--(9.52628,4);
    \draw[fill](7.79423,4)circle(0.2);
    \draw(7.79423,4)--(8.66025,3.5);
    \draw(7.79423,4)--(8.66025,4.5);
    \draw(7.79423,4)--(7.79423,5);
    \draw[fill](12.9904,1)circle(0.2);
    \draw(12.9904,1)--(12.9904,2);
    \draw[fill](7.79423,1)circle(0.2);
    \draw(7.79423,1)--(8.66025,0.5);
    \draw(7.79423,1)--(8.66025,1.5);
    \draw(7.79423,1)--(7.79423,2);
    \draw[fill](11.2583,6)circle(0.2);
    \draw(11.2583,6)--(12.1244,5.5);
    \draw(11.2583,6)--(12.1244,6.5);
    \draw(11.2583,6)--(11.2583,7);
    \draw[fill](11.2583,5)circle(0.2);
    \draw(11.2583,5)--(12.1244,4.5);
    \draw(11.2583,5)--(12.1244,5.5);
    \draw(11.2583,5)--(11.2583,6);
    \draw[fill](6.9282,7.5)circle(0.2);
    \draw(6.9282,7.5)--(7.79423,7);
    \draw(6.9282,7.5)--(7.79423,8);
    \draw[fill](10.3923,2.5)circle(0.2);
    \draw(10.3923,2.5)--(11.2583,2);
    \draw(10.3923,2.5)--(11.2583,3);
    \draw(10.3923,2.5)--(10.3923,3.5);
    \draw[fill](6.9282,3.5)circle(0.2);
    \draw(6.9282,3.5)--(7.79423,3);
    \draw(6.9282,3.5)--(7.79423,4);
    \draw(6.9282,3.5)--(6.9282,4.5);
    \draw[fill](7.79423,6)circle(0.2);
    \draw(7.79423,6)--(8.66025,5.5);
    \draw(7.79423,6)--(8.66025,6.5);
    \draw(7.79423,6)--(7.79423,7);
    \draw[fill](9.52628,2)circle(0.2);
    \draw(9.52628,2)--(10.3923,1.5);
    \draw(9.52628,2)--(10.3923,2.5);
    \draw(9.52628,2)--(9.52628,3);
    \draw[fill](6.06218,4)circle(0.2);
    \draw(6.06218,4)--(6.9282,3.5);
    \draw(6.06218,4)--(6.9282,4.5);
    \draw(6.06218,4)--(6.06218,5);
    \draw[fill](2.59808,7)circle(0.2);
    \draw(2.59808,7)--(3.4641,6.5);
    \draw(2.59808,7)--(3.4641,7.5);
    \draw(2.59808,7)--(2.59808,8);
    \draw[fill](1.73205,9.5)circle(0.2);
    \draw(1.73205,9.5)--(2.59808,9);
    \draw(1.73205,9.5)--(2.59808,10);
    \draw(1.73205,9.5)--(1.73205,10.5);
    \draw[fill](8.66025,0.5)circle(0.2);
    \draw(8.66025,0.5)--(9.52628,0);
    \draw(8.66025,0.5)--(9.52628,1);
    \draw(8.66025,0.5)--(8.66025,1.5);
    \draw[fill](12.9904,4)circle(0.2);
    \draw(12.9904,4)--(13.8564,4.5);
    \draw(12.9904,4)--(12.9904,5);
    \draw[fill](6.06218,6)circle(0.2);
    \draw(6.06218,6)--(6.9282,5.5);
    \draw(6.06218,6)--(6.9282,6.5);
    \draw(6.06218,6)--(6.06218,7);
    \draw[fill](2.59808,9)circle(0.2);
    \draw(2.59808,9)--(3.4641,8.5);
    \draw(2.59808,9)--(3.4641,9.5);
    \draw(2.59808,9)--(2.59808,10);
    \draw[fill](8.66025,8.5)circle(0.2);
    \draw(8.66025,8.5)--(9.52628,8);
    \draw[fill](8.66025,7.5)circle(0.2);
    \draw(8.66025,7.5)--(9.52628,7);
    \draw(8.66025,7.5)--(9.52628,8);
    \draw(8.66025,7.5)--(8.66025,8.5);
    \draw[fill](11.2583,2)circle(0.2);
    \draw(11.2583,2)--(12.1244,1.5);
    \draw(11.2583,2)--(12.1244,2.5);
    \draw(11.2583,2)--(11.2583,3);
    \draw[fill](11.2583,4)circle(0.2);
    \draw(11.2583,4)--(12.1244,3.5);
    \draw(11.2583,4)--(12.1244,4.5);
    \draw(11.2583,4)--(11.2583,5);
    \draw[fill](7.79423,8)circle(0.2);
    \draw(7.79423,8)--(8.66025,7.5);
    \draw(7.79423,8)--(8.66025,8.5);
    \draw[fill](4.33013,9)circle(0.2);
    \draw(4.33013,9)--(4.33013,10);
    \draw[fill](5.19615,5.5)circle(0.2);
    \draw(5.19615,5.5)--(6.06218,5);
    \draw(5.19615,5.5)--(6.06218,6);
    \draw(5.19615,5.5)--(5.19615,6.5);
    \draw[fill](9.52628,7)circle(0.2);
    \draw(9.52628,7)--(10.3923,6.5);
    \draw(9.52628,7)--(10.3923,7.5);
    \draw(9.52628,7)--(9.52628,8);
\end{tikzpicture}

%% file: line100_bias4_4mil_rotated.txt
\begin{tikzpicture}[scale=0.17,rotate=30]
    \draw[fill](5.19615,6)circle(0.2);
    \draw(5.19615,6)--(6.06218,5.5);
    \draw(5.19615,6)--(6.06218,6.5);
    \draw(5.19615,6)--(5.19615,7);
    \draw[fill](7.79423,2.5)circle(0.2);
    \draw(7.79423,2.5)--(8.66025,2);
    \draw(7.79423,2.5)--(8.66025,3);
    \draw(7.79423,2.5)--(7.79423,3.5);
    \draw[fill](7.79423,1.5)circle(0.2);
    \draw(7.79423,1.5)--(8.66025,1);
    \draw(7.79423,1.5)--(8.66025,2);
    \draw(7.79423,1.5)--(7.79423,2.5);
    \draw[fill](11.2583,3.5)circle(0.2);
    \draw(11.2583,3.5)--(11.2583,4.5);
    \draw[fill](10.3923,2)circle(0.2);
    \draw(10.3923,2)--(11.2583,2.5);
    \draw(10.3923,2)--(10.3923,3);
    \draw[fill](6.9282,3)circle(0.2);
    \draw(6.9282,3)--(7.79423,2.5);
    \draw(6.9282,3)--(7.79423,3.5);
    \draw(6.9282,3)--(6.9282,4);
    \draw[fill](2.59808,8.5)circle(0.2);
    \draw(2.59808,8.5)--(3.4641,8);
    \draw[fill](6.06218,10.5)circle(0.2);
    \draw(6.06218,10.5)--(6.9282,10);
    \draw(6.06218,10.5)--(6.9282,11);
    \draw[fill](7.79423,4.5)circle(0.2);
    \draw(7.79423,4.5)--(8.66025,4);
    \draw(7.79423,4.5)--(8.66025,5);
    \draw(7.79423,4.5)--(7.79423,5.5);
    \draw[fill](0.866025,3.5)circle(0.2);
    \draw(0.866025,3.5)--(1.73205,3);
    \draw(0.866025,3.5)--(1.73205,4);
    \draw(0.866025,3.5)--(0.866025,4.5);
    \draw[fill](4.33013,5.5)circle(0.2);
    \draw(4.33013,5.5)--(5.19615,5);
    \draw(4.33013,5.5)--(5.19615,6);
    \draw(4.33013,5.5)--(4.33013,6.5);
    \draw[fill](2.59808,3.5)circle(0.2);
    \draw(2.59808,3.5)--(3.4641,3);
    \draw(2.59808,3.5)--(3.4641,4);
    \draw(2.59808,3.5)--(2.59808,4.5);
    \draw[fill](6.06218,4.5)circle(0.2);
    \draw(6.06218,4.5)--(6.9282,4);
    \draw(6.06218,4.5)--(6.9282,5);
    \draw(6.06218,4.5)--(6.06218,5.5);
    \draw[fill](10.3923,3)circle(0.2);
    \draw(10.3923,3)--(11.2583,2.5);
    \draw(10.3923,3)--(11.2583,3.5);
    \draw(10.3923,3)--(10.3923,4);
    \draw[fill](1.73205,4)circle(0.2);
    \draw(1.73205,4)--(2.59808,3.5);
    \draw(1.73205,4)--(2.59808,4.5);
    \draw(1.73205,4)--(1.73205,5);
    \draw[fill](2.59808,5.5)circle(0.2);
    \draw(2.59808,5.5)--(3.4641,5);
    \draw(2.59808,5.5)--(3.4641,6);
    \draw[fill](5.19615,5)circle(0.2);
    \draw(5.19615,5)--(6.06218,4.5);
    \draw(5.19615,5)--(6.06218,5.5);
    \draw(5.19615,5)--(5.19615,6);
    \draw[fill](6.9282,10)circle(0.2);
    \draw(6.9282,10)--(7.79423,9.5);
    \draw(6.9282,10)--(7.79423,10.5);
    \draw(6.9282,10)--(6.9282,11);
    \draw[fill](4.33013,7.5)circle(0.2);
    \draw(4.33013,7.5)--(5.19615,7);
    \draw(4.33013,7.5)--(5.19615,8);
    \draw[fill](6.9282,8)circle(0.2);
    \draw(6.9282,8)--(7.79423,8.5);
    \draw(6.9282,8)--(6.9282,9);
    \draw[fill](4.33013,3.5)circle(0.2);
    \draw(4.33013,3.5)--(5.19615,3);
    \draw(4.33013,3.5)--(5.19615,4);
    \draw(4.33013,3.5)--(4.33013,4.5);
    \draw[fill](9.52628,3.5)circle(0.2);
    \draw(9.52628,3.5)--(10.3923,3);
    \draw(9.52628,3.5)--(10.3923,4);
    \draw(9.52628,3.5)--(9.52628,4.5);
    \draw[fill](10.3923,5)circle(0.2);
    \draw(10.3923,5)--(11.2583,4.5);
    \draw[fill](0.866025,1.5)circle(0.2);
    \draw(0.866025,1.5)--(1.73205,1);
    \draw(0.866025,1.5)--(0.866025,2.5);
    \draw[fill](9.52628,5.5)circle(0.2);
    \draw(9.52628,5.5)--(10.3923,5);
    \draw(9.52628,5.5)--(9.52628,6.5);
    \draw[fill](7.79423,5.5)circle(0.2);
    \draw(7.79423,5.5)--(8.66025,5);
    \draw[fill](1.73205,8)circle(0.2);
    \draw(1.73205,8)--(2.59808,7.5);
    \draw(1.73205,8)--(2.59808,8.5);
    \draw(1.73205,8)--(1.73205,9);
    \draw[fill](7.79423,10.5)circle(0.2);
    \draw(7.79423,10.5)--(8.66025,11);
    \draw(7.79423,10.5)--(7.79423,11.5);
    \draw[fill](3.4641,7)circle(0.2);
    \draw(3.4641,7)--(4.33013,6.5);
    \draw(3.4641,7)--(4.33013,7.5);
    \draw(3.4641,7)--(3.4641,8);
    \draw[fill](4.33013,2.5)circle(0.2);
    \draw(4.33013,2.5)--(5.19615,2);
    \draw(4.33013,2.5)--(5.19615,3);
    \draw(4.33013,2.5)--(4.33013,3.5);
    \draw[fill](3.4641,2)circle(0.2);
    \draw(3.4641,2)--(4.33013,1.5);
    \draw(3.4641,2)--(4.33013,2.5);
    \draw(3.4641,2)--(3.4641,3);
    \draw[fill](1.73205,1)circle(0.2);
    \draw[fill](6.06218,1.5)circle(0.2);
    \draw(6.06218,1.5)--(6.9282,1);
    \draw(6.06218,1.5)--(6.9282,2);
    \draw(6.06218,1.5)--(6.06218,2.5);
    \draw[fill](1.73205,0)circle(0.2);
    \draw(1.73205,0)--(1.73205,1);
    \draw[fill](8.66025,9)circle(0.2);
    \draw[fill](7.79423,3.5)circle(0.2);
    \draw(7.79423,3.5)--(8.66025,3);
    \draw(7.79423,3.5)--(8.66025,4);
    \draw(7.79423,3.5)--(7.79423,4.5);
    \draw[fill](6.9282,11)circle(0.2);
    \draw(6.9282,11)--(7.79423,10.5);
    \draw(6.9282,11)--(7.79423,11.5);
    \draw(6.9282,11)--(6.9282,12);
    \draw[fill](5.19615,3)circle(0.2);
    \draw(5.19615,3)--(6.06218,2.5);
    \draw(5.19615,3)--(6.06218,3.5);
    \draw(5.19615,3)--(5.19615,4);
    \draw[fill](6.06218,3.5)circle(0.2);
    \draw(6.06218,3.5)--(6.9282,3);
    \draw(6.06218,3.5)--(6.9282,4);
    \draw(6.06218,3.5)--(6.06218,4.5);
    \draw[fill](6.06218,8.5)circle(0.2);
    \draw(6.06218,8.5)--(6.9282,8);
    \draw(6.06218,8.5)--(6.9282,9);
    \draw(6.06218,8.5)--(6.06218,9.5);
    \draw[fill](3.4641,4)circle(0.2);
    \draw(3.4641,4)--(4.33013,3.5);
    \draw(3.4641,4)--(4.33013,4.5);
    \draw(3.4641,4)--(3.4641,5);
    \draw[fill](10.3923,4)circle(0.2);
    \draw(10.3923,4)--(11.2583,3.5);
    \draw(10.3923,4)--(11.2583,4.5);
    \draw(10.3923,4)--(10.3923,5);
    \draw[fill](0.866025,2.5)circle(0.2);
    \draw(0.866025,2.5)--(1.73205,3);
    \draw(0.866025,2.5)--(0.866025,3.5);
    \draw[fill](2.59808,7.5)circle(0.2);
    \draw(2.59808,7.5)--(3.4641,7);
    \draw(2.59808,7.5)--(3.4641,8);
    \draw(2.59808,7.5)--(2.59808,8.5);
    \draw[fill](6.06218,6.5)circle(0.2);
    \draw(6.06218,6.5)--(6.9282,6);
    \draw(6.06218,6.5)--(6.9282,7);
    \draw(6.06218,6.5)--(6.06218,7.5);
    \draw[fill](0.866025,5.5)circle(0.2);
    \draw(0.866025,5.5)--(1.73205,5);
    \draw[fill](11.2583,4.5)circle(0.2);
    \draw[fill](6.9282,12)circle(0.2);
    \draw(6.9282,12)--(7.79423,11.5);
    \draw[fill](8.66025,1)circle(0.2);
    \draw(8.66025,1)--(9.52628,0.5);
    \draw(8.66025,1)--(9.52628,1.5);
    \draw(8.66025,1)--(8.66025,2);
    \draw[fill](6.06218,7.5)circle(0.2);
    \draw(6.06218,7.5)--(6.9282,7);
    \draw(6.06218,7.5)--(6.9282,8);
    \draw(6.06218,7.5)--(6.06218,8.5);
    \draw[fill](6.06218,9.5)circle(0.2);
    \draw(6.06218,9.5)--(6.9282,9);
    \draw(6.06218,9.5)--(6.9282,10);
    \draw(6.06218,9.5)--(6.06218,10.5);
    \draw[fill](3.4641,3)circle(0.2);
    \draw(3.4641,3)--(4.33013,2.5);
    \draw(3.4641,3)--(4.33013,3.5);
    \draw(3.4641,3)--(3.4641,4);
    \draw[fill](2.59808,4.5)circle(0.2);
    \draw(2.59808,4.5)--(3.4641,4);
    \draw(2.59808,4.5)--(3.4641,5);
    \draw(2.59808,4.5)--(2.59808,5.5);
    \draw[fill](3.4641,8)circle(0.2);
    \draw(3.4641,8)--(4.33013,7.5);
    \draw[fill](6.9282,2)circle(0.2);
    \draw(6.9282,2)--(7.79423,1.5);
    \draw(6.9282,2)--(7.79423,2.5);
    \draw(6.9282,2)--(6.9282,3);
    \draw[fill](5.19615,1)circle(0.2);
    \draw(5.19615,1)--(6.06218,1.5);
    \draw(5.19615,1)--(5.19615,2);
    \draw[fill](8.66025,0)circle(0.2);
    \draw(8.66025,0)--(9.52628,0.5);
    \draw(8.66025,0)--(8.66025,1);
    \draw[fill](5.19615,0)circle(0.2);
    \draw(5.19615,0)--(5.19615,1);
    \draw[fill](3.4641,5)circle(0.2);
    \draw(3.4641,5)--(4.33013,4.5);
    \draw(3.4641,5)--(4.33013,5.5);
    \draw(3.4641,5)--(3.4641,6);
    \draw[fill](4.33013,4.5)circle(0.2);
    \draw(4.33013,4.5)--(5.19615,4);
    \draw(4.33013,4.5)--(5.19615,5);
    \draw(4.33013,4.5)--(4.33013,5.5);
    \draw[fill](6.06218,5.5)circle(0.2);
    \draw(6.06218,5.5)--(6.9282,5);
    \draw(6.06218,5.5)--(6.9282,6);
    \draw(6.06218,5.5)--(6.06218,6.5);
    \draw[fill](1.73205,3)circle(0.2);
    \draw(1.73205,3)--(2.59808,2.5);
    \draw(1.73205,3)--(2.59808,3.5);
    \draw(1.73205,3)--(1.73205,4);
    \draw[fill](6.06218,2.5)circle(0.2);
    \draw(6.06218,2.5)--(6.9282,2);
    \draw(6.06218,2.5)--(6.9282,3);
    \draw(6.06218,2.5)--(6.06218,3.5);
    \draw[fill](0,4)circle(0.2);
    \draw(0,4)--(0.866025,3.5);
    \draw(0,4)--(0.866025,4.5);
    \draw[fill](1.73205,9)circle(0.2);
    \draw(1.73205,9)--(2.59808,8.5);
    \draw[fill](0,3)circle(0.2);
    \draw(0,3)--(0.866025,2.5);
    \draw(0,3)--(0.866025,3.5);
    \draw(0,3)--(0,4);
    \draw[fill](8.66025,3)circle(0.2);
    \draw(8.66025,3)--(9.52628,2.5);
    \draw(8.66025,3)--(9.52628,3.5);
    \draw(8.66025,3)--(8.66025,4);
    \draw[fill](3.4641,6)circle(0.2);
    \draw(3.4641,6)--(4.33013,5.5);
    \draw(3.4641,6)--(4.33013,6.5);
    \draw(3.4641,6)--(3.4641,7);
    \draw[fill](1.73205,5)circle(0.2);
    \draw(1.73205,5)--(2.59808,4.5);
    \draw(1.73205,5)--(2.59808,5.5);
    \draw[fill](4.33013,1.5)circle(0.2);
    \draw(4.33013,1.5)--(5.19615,1);
    \draw(4.33013,1.5)--(5.19615,2);
    \draw(4.33013,1.5)--(4.33013,2.5);
    \draw[fill](6.9282,6)circle(0.2);
    \draw(6.9282,6)--(7.79423,5.5);
    \draw(6.9282,6)--(6.9282,7);
    \draw[fill](9.52628,0.5)circle(0.2);
    \draw(9.52628,0.5)--(9.52628,1.5);
    \draw[fill](8.66025,5)circle(0.2);
    \draw(8.66025,5)--(9.52628,4.5);
    \draw(8.66025,5)--(9.52628,5.5);
    \draw[fill](7.79423,0.5)circle(0.2);
    \draw(7.79423,0.5)--(8.66025,0);
    \draw(7.79423,0.5)--(8.66025,1);
    \draw(7.79423,0.5)--(7.79423,1.5);
    \draw[fill](5.19615,4)circle(0.2);
    \draw(5.19615,4)--(6.06218,3.5);
    \draw(5.19615,4)--(6.06218,4.5);
    \draw(5.19615,4)--(5.19615,5);
    \draw[fill](0.866025,4.5)circle(0.2);
    \draw(0.866025,4.5)--(1.73205,4);
    \draw(0.866025,4.5)--(1.73205,5);
    \draw(0.866025,4.5)--(0.866025,5.5);
    \draw[fill](2.59808,2.5)circle(0.2);
    \draw(2.59808,2.5)--(3.4641,2);
    \draw(2.59808,2.5)--(3.4641,3);
    \draw(2.59808,2.5)--(2.59808,3.5);
    \draw[fill](4.33013,6.5)circle(0.2);
    \draw(4.33013,6.5)--(5.19615,6);
    \draw(4.33013,6.5)--(5.19615,7);
    \draw(4.33013,6.5)--(4.33013,7.5);
    \draw[fill](4.33013,0.5)circle(0.2);
    \draw(4.33013,0.5)--(5.19615,0);
    \draw(4.33013,0.5)--(5.19615,1);
    \draw(4.33013,0.5)--(4.33013,1.5);
    \draw[fill](5.19615,2)circle(0.2);
    \draw(5.19615,2)--(6.06218,1.5);
    \draw(5.19615,2)--(6.06218,2.5);
    \draw(5.19615,2)--(5.19615,3);
    \draw[fill](7.79423,9.5)circle(0.2);
    \draw(7.79423,9.5)--(8.66025,9);
    \draw(7.79423,9.5)--(7.79423,10.5);
    \draw[fill](6.9282,5)circle(0.2);
    \draw(6.9282,5)--(7.79423,4.5);
    \draw(6.9282,5)--(7.79423,5.5);
    \draw(6.9282,5)--(6.9282,6);
    \draw[fill](6.9282,7)circle(0.2);
    \draw(6.9282,7)--(6.9282,8);
    \draw[fill](6.9282,4)circle(0.2);
    \draw(6.9282,4)--(7.79423,3.5);
    \draw(6.9282,4)--(7.79423,4.5);
    \draw(6.9282,4)--(6.9282,5);
    \draw[fill](8.66025,11)circle(0.2);
    \draw[fill](9.52628,6.5)circle(0.2);
    \draw[fill](7.79423,8.5)circle(0.2);
    \draw(7.79423,8.5)--(8.66025,9);
    \draw(7.79423,8.5)--(7.79423,9.5);
    \draw[fill](5.19615,8)circle(0.2);
    \draw(5.19615,8)--(6.06218,7.5);
    \draw(5.19615,8)--(6.06218,8.5);
    \draw(5.19615,8)--(5.19615,9);
    \draw[fill](6.9282,9)circle(0.2);
    \draw(6.9282,9)--(7.79423,8.5);
    \draw(6.9282,9)--(7.79423,9.5);
    \draw(6.9282,9)--(6.9282,10);
    \draw[fill](6.9282,1)circle(0.2);
    \draw(6.9282,1)--(7.79423,0.5);
    \draw(6.9282,1)--(7.79423,1.5);
    \draw(6.9282,1)--(6.9282,2);
    \draw[fill](5.19615,9)circle(0.2);
    \draw(5.19615,9)--(6.06218,8.5);
    \draw(5.19615,9)--(6.06218,9.5);
    \draw[fill](11.2583,2.5)circle(0.2);
    \draw(11.2583,2.5)--(11.2583,3.5);
    \draw[fill](8.66025,4)circle(0.2);
    \draw(8.66025,4)--(9.52628,3.5);
    \draw(8.66025,4)--(9.52628,4.5);
    \draw(8.66025,4)--(8.66025,5);
    \draw[fill](1.73205,7)circle(0.2);
    \draw(1.73205,7)--(2.59808,7.5);
    \draw(1.73205,7)--(1.73205,8);
    \draw[fill](9.52628,4.5)circle(0.2);
    \draw(9.52628,4.5)--(10.3923,4);
    \draw(9.52628,4.5)--(10.3923,5);
    \draw(9.52628,4.5)--(9.52628,5.5);
    \draw[fill](8.66025,2)circle(0.2);
    \draw(8.66025,2)--(9.52628,1.5);
    \draw(8.66025,2)--(9.52628,2.5);
    \draw(8.66025,2)--(8.66025,3);
    \draw[fill](9.52628,1.5)circle(0.2);
    \draw(9.52628,1.5)--(10.3923,2);
    \draw(9.52628,1.5)--(9.52628,2.5);
    \draw[fill](5.19615,7)circle(0.2);
    \draw(5.19615,7)--(6.06218,6.5);
    \draw(5.19615,7)--(6.06218,7.5);
    \draw(5.19615,7)--(5.19615,8);
    \draw[fill](9.52628,2.5)circle(0.2);
    \draw(9.52628,2.5)--(10.3923,2);
    \draw(9.52628,2.5)--(10.3923,3);
    \draw(9.52628,2.5)--(9.52628,3.5);
    \draw[fill](7.79423,11.5)circle(0.2);
    \draw(7.79423,11.5)--(8.66025,11);
\end{tikzpicture}

%% file: line100_bias4_5mil_rotated.txt
\begin{tikzpicture}[scale=0.17,rotate=30]
    \draw[fill](0.866025,8.5)circle(0.2);
    \draw[fill](7.79423,6.5)circle(0.2);
    \draw(7.79423,6.5)--(8.66025,6);
    \draw(7.79423,6.5)--(8.66025,7);
    \draw(7.79423,6.5)--(7.79423,7.5);
    \draw[fill](3.4641,1)circle(0.2);
    \draw(3.4641,1)--(4.33013,0.5);
    \draw(3.4641,1)--(4.33013,1.5);
    \draw(3.4641,1)--(3.4641,2);
    \draw[fill](4.33013,9.5)circle(0.2);
    \draw(4.33013,9.5)--(5.19615,9);
    \draw(4.33013,9.5)--(5.19615,10);
    \draw[fill](10.3923,7)circle(0.2);
    \draw[fill](0.866025,3.5)circle(0.2);
    \draw(0.866025,3.5)--(1.73205,3);
    \draw(0.866025,3.5)--(1.73205,4);
    \draw(0.866025,3.5)--(0.866025,4.5);
    \draw[fill](1.73205,7)circle(0.2);
    \draw(1.73205,7)--(2.59808,6.5);
    \draw(1.73205,7)--(2.59808,7.5);
    \draw[fill](6.9282,10)circle(0.2);
    \draw(6.9282,10)--(7.79423,9.5);
    \draw(6.9282,10)--(7.79423,10.5);
    \draw(6.9282,10)--(6.9282,11);
    \draw[fill](3.4641,5)circle(0.2);
    \draw(3.4641,5)--(4.33013,4.5);
    \draw(3.4641,5)--(4.33013,5.5);
    \draw(3.4641,5)--(3.4641,6);
    \draw[fill](7.79423,2.5)circle(0.2);
    \draw[fill](4.33013,1.5)circle(0.2);
    \draw(4.33013,1.5)--(5.19615,2);
    \draw(4.33013,1.5)--(4.33013,2.5);
    \draw[fill](6.9282,2)circle(0.2);
    \draw(6.9282,2)--(7.79423,1.5);
    \draw(6.9282,2)--(7.79423,2.5);
    \draw(6.9282,2)--(6.9282,3);
    \draw[fill](5.19615,0)circle(0.2);
    \draw[fill](6.06218,9.5)circle(0.2);
    \draw(6.06218,9.5)--(6.9282,9);
    \draw(6.06218,9.5)--(6.9282,10);
    \draw(6.06218,9.5)--(6.06218,10.5);
    \draw[fill](4.33013,3.5)circle(0.2);
    \draw(4.33013,3.5)--(5.19615,3);
    \draw(4.33013,3.5)--(5.19615,4);
    \draw(4.33013,3.5)--(4.33013,4.5);
    \draw[fill](0,7)circle(0.2);
    \draw(0,7)--(0.866025,6.5);
    \draw(0,7)--(0.866025,7.5);
    \draw(0,7)--(0,8);
    \draw[fill](0,6)circle(0.2);
    \draw(0,6)--(0.866025,5.5);
    \draw(0,6)--(0.866025,6.5);
    \draw(0,6)--(0,7);
    \draw[fill](6.9282,8)circle(0.2);
    \draw(6.9282,8)--(7.79423,7.5);
    \draw(6.9282,8)--(7.79423,8.5);
    \draw(6.9282,8)--(6.9282,9);
    \draw[fill](10.3923,6)circle(0.2);
    \draw(10.3923,6)--(11.2583,5.5);
    \draw(10.3923,6)--(10.3923,7);
    \draw[fill](2.59808,5.5)circle(0.2);
    \draw(2.59808,5.5)--(3.4641,5);
    \draw(2.59808,5.5)--(3.4641,6);
    \draw(2.59808,5.5)--(2.59808,6.5);
    \draw[fill](10.3923,5)circle(0.2);
    \draw(10.3923,5)--(11.2583,5.5);
    \draw(10.3923,5)--(10.3923,6);
    \draw[fill](4.33013,8.5)circle(0.2);
    \draw(4.33013,8.5)--(5.19615,8);
    \draw(4.33013,8.5)--(5.19615,9);
    \draw(4.33013,8.5)--(4.33013,9.5);
    \draw[fill](7.79423,5.5)circle(0.2);
    \draw(7.79423,5.5)--(8.66025,5);
    \draw(7.79423,5.5)--(8.66025,6);
    \draw(7.79423,5.5)--(7.79423,6.5);
    \draw[fill](6.06218,3.5)circle(0.2);
    \draw(6.06218,3.5)--(6.9282,3);
    \draw(6.06218,3.5)--(6.9282,4);
    \draw(6.06218,3.5)--(6.06218,4.5);
    \draw[fill](8.66025,9)circle(0.2);
    \draw(8.66025,9)--(8.66025,10);
    \draw[fill](8.66025,11)circle(0.2);
    \draw[fill](5.19615,9)circle(0.2);
    \draw(5.19615,9)--(6.06218,8.5);
    \draw(5.19615,9)--(6.06218,9.5);
    \draw(5.19615,9)--(5.19615,10);
    \draw[fill](5.19615,7)circle(0.2);
    \draw(5.19615,7)--(6.06218,6.5);
    \draw(5.19615,7)--(6.06218,7.5);
    \draw(5.19615,7)--(5.19615,8);
    \draw[fill](4.33013,7.5)circle(0.2);
    \draw(4.33013,7.5)--(5.19615,7);
    \draw(4.33013,7.5)--(5.19615,8);
    \draw(4.33013,7.5)--(4.33013,8.5);
    \draw[fill](1.73205,5)circle(0.2);
    \draw(1.73205,5)--(2.59808,4.5);
    \draw(1.73205,5)--(2.59808,5.5);
    \draw(1.73205,5)--(1.73205,6);
    \draw[fill](6.9282,6)circle(0.2);
    \draw(6.9282,6)--(7.79423,5.5);
    \draw(6.9282,6)--(7.79423,6.5);
    \draw(6.9282,6)--(6.9282,7);
    \draw[fill](4.33013,6.5)circle(0.2);
    \draw(4.33013,6.5)--(5.19615,6);
    \draw(4.33013,6.5)--(5.19615,7);
    \draw(4.33013,6.5)--(4.33013,7.5);
    \draw[fill](3.4641,7)circle(0.2);
    \draw(3.4641,7)--(4.33013,6.5);
    \draw(3.4641,7)--(4.33013,7.5);
    \draw(3.4641,7)--(3.4641,8);
    \draw[fill](6.06218,10.5)circle(0.2);
    \draw(6.06218,10.5)--(6.9282,10);
    \draw(6.06218,10.5)--(6.9282,11);
    \draw[fill](5.19615,6)circle(0.2);
    \draw(5.19615,6)--(6.06218,5.5);
    \draw(5.19615,6)--(6.06218,6.5);
    \draw(5.19615,6)--(5.19615,7);
    \draw[fill](6.9282,9)circle(0.2);
    \draw(6.9282,9)--(7.79423,8.5);
    \draw(6.9282,9)--(7.79423,9.5);
    \draw(6.9282,9)--(6.9282,10);
    \draw[fill](9.52628,6.5)circle(0.2);
    \draw(9.52628,6.5)--(10.3923,6);
    \draw(9.52628,6.5)--(10.3923,7);
    \draw[fill](9.52628,4.5)circle(0.2);
    \draw(9.52628,4.5)--(10.3923,4);
    \draw(9.52628,4.5)--(10.3923,5);
    \draw(9.52628,4.5)--(9.52628,5.5);
    \draw[fill](1.73205,4)circle(0.2);
    \draw(1.73205,4)--(2.59808,3.5);
    \draw(1.73205,4)--(2.59808,4.5);
    \draw(1.73205,4)--(1.73205,5);
    \draw[fill](8.66025,6)circle(0.2);
    \draw(8.66025,6)--(9.52628,5.5);
    \draw(8.66025,6)--(9.52628,6.5);
    \draw(8.66025,6)--(8.66025,7);
    \draw[fill](2.59808,4.5)circle(0.2);
    \draw(2.59808,4.5)--(3.4641,4);
    \draw(2.59808,4.5)--(3.4641,5);
    \draw(2.59808,4.5)--(2.59808,5.5);
    \draw[fill](0,8)circle(0.2);
    \draw(0,8)--(0.866025,7.5);
    \draw(0,8)--(0.866025,8.5);
    \draw[fill](8.66025,5)circle(0.2);
    \draw(8.66025,5)--(9.52628,4.5);
    \draw(8.66025,5)--(9.52628,5.5);
    \draw(8.66025,5)--(8.66025,6);
    \draw[fill](7.79423,8.5)circle(0.2);
    \draw(7.79423,8.5)--(8.66025,8);
    \draw(7.79423,8.5)--(8.66025,9);
    \draw(7.79423,8.5)--(7.79423,9.5);
    \draw[fill](3.4641,2)circle(0.2);
    \draw(3.4641,2)--(4.33013,1.5);
    \draw(3.4641,2)--(4.33013,2.5);
    \draw(3.4641,2)--(3.4641,3);
    \draw[fill](3.4641,9)circle(0.2);
    \draw(3.4641,9)--(4.33013,8.5);
    \draw(3.4641,9)--(4.33013,9.5);
    \draw(3.4641,9)--(3.4641,10);
    \draw[fill](0.866025,5.5)circle(0.2);
    \draw(0.866025,5.5)--(1.73205,5);
    \draw(0.866025,5.5)--(1.73205,6);
    \draw(0.866025,5.5)--(0.866025,6.5);
    \draw[fill](5.19615,8)circle(0.2);
    \draw(5.19615,8)--(6.06218,7.5);
    \draw(5.19615,8)--(6.06218,8.5);
    \draw(5.19615,8)--(5.19615,9);
    \draw[fill](5.19615,3)circle(0.2);
    \draw(5.19615,3)--(6.06218,2.5);
    \draw(5.19615,3)--(6.06218,3.5);
    \draw(5.19615,3)--(5.19615,4);
    \draw[fill](5.19615,11)circle(0.2);
    \draw(5.19615,11)--(6.06218,10.5);
    \draw[fill](4.33013,4.5)circle(0.2);
    \draw(4.33013,4.5)--(5.19615,4);
    \draw(4.33013,4.5)--(5.19615,5);
    \draw(4.33013,4.5)--(4.33013,5.5);
    \draw[fill](4.33013,0.5)circle(0.2);
    \draw(4.33013,0.5)--(5.19615,0);
    \draw(4.33013,0.5)--(4.33013,1.5);
    \draw[fill](11.2583,5.5)circle(0.2);
    \draw[fill](6.06218,7.5)circle(0.2);
    \draw(6.06218,7.5)--(6.9282,7);
    \draw(6.06218,7.5)--(6.9282,8);
    \draw(6.06218,7.5)--(6.06218,8.5);
    \draw[fill](3.4641,4)circle(0.2);
    \draw(3.4641,4)--(4.33013,3.5);
    \draw(3.4641,4)--(4.33013,4.5);
    \draw(3.4641,4)--(3.4641,5);
    \draw[fill](0.866025,7.5)circle(0.2);
    \draw(0.866025,7.5)--(1.73205,7);
    \draw(0.866025,7.5)--(0.866025,8.5);
    \draw[fill](2.59808,7.5)circle(0.2);
    \draw(2.59808,7.5)--(3.4641,7);
    \draw(2.59808,7.5)--(3.4641,8);
    \draw(2.59808,7.5)--(2.59808,8.5);
    \draw[fill](6.9282,3)circle(0.2);
    \draw(6.9282,3)--(7.79423,2.5);
    \draw(6.9282,3)--(6.9282,4);
    \draw[fill](2.59808,8.5)circle(0.2);
    \draw(2.59808,8.5)--(3.4641,8);
    \draw(2.59808,8.5)--(3.4641,9);
    \draw[fill](1.73205,6)circle(0.2);
    \draw(1.73205,6)--(2.59808,5.5);
    \draw(1.73205,6)--(2.59808,6.5);
    \draw(1.73205,6)--(1.73205,7);
    \draw[fill](6.06218,4.5)circle(0.2);
    \draw(6.06218,4.5)--(6.9282,4);
    \draw(6.06218,4.5)--(6.9282,5);
    \draw(6.06218,4.5)--(6.06218,5.5);
    \draw[fill](5.19615,4)circle(0.2);
    \draw(5.19615,4)--(6.06218,3.5);
    \draw(5.19615,4)--(6.06218,4.5);
    \draw(5.19615,4)--(5.19615,5);
    \draw[fill](8.66025,10)circle(0.2);
    \draw(8.66025,10)--(8.66025,11);
    \draw[fill](3.4641,6)circle(0.2);
    \draw(3.4641,6)--(4.33013,5.5);
    \draw(3.4641,6)--(4.33013,6.5);
    \draw(3.4641,6)--(3.4641,7);
    \draw[fill](6.9282,11)circle(0.2);
    \draw(6.9282,11)--(7.79423,10.5);
    \draw(6.9282,11)--(7.79423,11.5);
    \draw[fill](6.06218,5.5)circle(0.2);
    \draw(6.06218,5.5)--(6.9282,5);
    \draw(6.06218,5.5)--(6.9282,6);
    \draw(6.06218,5.5)--(6.06218,6.5);
    \draw[fill](4.33013,2.5)circle(0.2);
    \draw(4.33013,2.5)--(5.19615,2);
    \draw(4.33013,2.5)--(5.19615,3);
    \draw(4.33013,2.5)--(4.33013,3.5);
    \draw[fill](6.06218,8.5)circle(0.2);
    \draw(6.06218,8.5)--(6.9282,8);
    \draw(6.06218,8.5)--(6.9282,9);
    \draw(6.06218,8.5)--(6.06218,9.5);
    \draw[fill](6.06218,6.5)circle(0.2);
    \draw(6.06218,6.5)--(6.9282,6);
    \draw(6.06218,6.5)--(6.9282,7);
    \draw(6.06218,6.5)--(6.06218,7.5);
    \draw[fill](0,4)circle(0.2);
    \draw(0,4)--(0.866025,3.5);
    \draw(0,4)--(0.866025,4.5);
    \draw(0,4)--(0,5);
    \draw[fill](9.52628,5.5)circle(0.2);
    \draw(9.52628,5.5)--(10.3923,5);
    \draw(9.52628,5.5)--(10.3923,6);
    \draw(9.52628,5.5)--(9.52628,6.5);
    \draw[fill](3.4641,3)circle(0.2);
    \draw(3.4641,3)--(4.33013,2.5);
    \draw(3.4641,3)--(4.33013,3.5);
    \draw(3.4641,3)--(3.4641,4);
    \draw[fill](6.9282,4)circle(0.2);
    \draw(6.9282,4)--(7.79423,4.5);
    \draw(6.9282,4)--(6.9282,5);
    \draw[fill](7.79423,11.5)circle(0.2);
    \draw(7.79423,11.5)--(8.66025,11);
    \draw[fill](6.06218,2.5)circle(0.2);
    \draw(6.06218,2.5)--(6.9282,2);
    \draw(6.06218,2.5)--(6.9282,3);
    \draw(6.06218,2.5)--(6.06218,3.5);
    \draw[fill](0,5)circle(0.2);
    \draw(0,5)--(0.866025,4.5);
    \draw(0,5)--(0.866025,5.5);
    \draw(0,5)--(0,6);
    \draw[fill](3.4641,8)circle(0.2);
    \draw(3.4641,8)--(4.33013,7.5);
    \draw(3.4641,8)--(4.33013,8.5);
    \draw(3.4641,8)--(3.4641,9);
    \draw[fill](5.19615,5)circle(0.2);
    \draw(5.19615,5)--(6.06218,4.5);
    \draw(5.19615,5)--(6.06218,5.5);
    \draw(5.19615,5)--(5.19615,6);
    \draw[fill](1.73205,3)circle(0.2);
    \draw(1.73205,3)--(2.59808,2.5);
    \draw(1.73205,3)--(2.59808,3.5);
    \draw(1.73205,3)--(1.73205,4);
    \draw[fill](2.59808,6.5)circle(0.2);
    \draw(2.59808,6.5)--(3.4641,6);
    \draw(2.59808,6.5)--(3.4641,7);
    \draw(2.59808,6.5)--(2.59808,7.5);
    \draw[fill](0,3)circle(0.2);
    \draw(0,3)--(0.866025,3.5);
    \draw(0,3)--(0,4);
    \draw[fill](7.79423,1.5)circle(0.2);
    \draw(7.79423,1.5)--(7.79423,2.5);
    \draw[fill](1.73205,2)circle(0.2);
    \draw(1.73205,2)--(2.59808,2.5);
    \draw(1.73205,2)--(1.73205,3);
    \draw[fill](10.3923,4)circle(0.2);
    \draw(10.3923,4)--(10.3923,5);
    \draw[fill](0.866025,4.5)circle(0.2);
    \draw(0.866025,4.5)--(1.73205,4);
    \draw(0.866025,4.5)--(1.73205,5);
    \draw(0.866025,4.5)--(0.866025,5.5);
    \draw[fill](8.66025,8)circle(0.2);
    \draw(8.66025,8)--(8.66025,9);
    \draw[fill](6.9282,7)circle(0.2);
    \draw(6.9282,7)--(7.79423,6.5);
    \draw(6.9282,7)--(7.79423,7.5);
    \draw(6.9282,7)--(6.9282,8);
    \draw[fill](7.79423,9.5)circle(0.2);
    \draw(7.79423,9.5)--(8.66025,9);
    \draw(7.79423,9.5)--(8.66025,10);
    \draw(7.79423,9.5)--(7.79423,10.5);
    \draw[fill](2.59808,3.5)circle(0.2);
    \draw(2.59808,3.5)--(3.4641,3);
    \draw(2.59808,3.5)--(3.4641,4);
    \draw(2.59808,3.5)--(2.59808,4.5);
    \draw[fill](5.19615,2)circle(0.2);
    \draw(5.19615,2)--(6.06218,2.5);
    \draw(5.19615,2)--(5.19615,3);
    \draw[fill](3.4641,10)circle(0.2);
    \draw(3.4641,10)--(4.33013,9.5);
    \draw[fill](8.66025,7)circle(0.2);
    \draw(8.66025,7)--(9.52628,6.5);
    \draw(8.66025,7)--(8.66025,8);
    \draw[fill](7.79423,7.5)circle(0.2);
    \draw(7.79423,7.5)--(8.66025,7);
    \draw(7.79423,7.5)--(8.66025,8);
    \draw(7.79423,7.5)--(7.79423,8.5);
    \draw[fill](6.9282,5)circle(0.2);
    \draw(6.9282,5)--(7.79423,4.5);
    \draw(6.9282,5)--(7.79423,5.5);
    \draw(6.9282,5)--(6.9282,6);
    \draw[fill](0.866025,6.5)circle(0.2);
    \draw(0.866025,6.5)--(1.73205,6);
    \draw(0.866025,6.5)--(1.73205,7);
    \draw(0.866025,6.5)--(0.866025,7.5);
    \draw[fill](7.79423,4.5)circle(0.2);
    \draw(7.79423,4.5)--(8.66025,5);
    \draw(7.79423,4.5)--(7.79423,5.5);
    \draw[fill](7.79423,10.5)circle(0.2);
    \draw(7.79423,10.5)--(8.66025,10);
    \draw(7.79423,10.5)--(8.66025,11);
    \draw(7.79423,10.5)--(7.79423,11.5);
    \draw[fill](2.59808,2.5)circle(0.2);
    \draw(2.59808,2.5)--(3.4641,2);
    \draw(2.59808,2.5)--(3.4641,3);
    \draw(2.59808,2.5)--(2.59808,3.5);
    \draw[fill](5.19615,10)circle(0.2);
    \draw(5.19615,10)--(6.06218,9.5);
    \draw(5.19615,10)--(6.06218,10.5);
    \draw(5.19615,10)--(5.19615,11);
    \draw[fill](4.33013,5.5)circle(0.2);
    \draw(4.33013,5.5)--(5.19615,5);
    \draw(4.33013,5.5)--(5.19615,6);
    \draw(4.33013,5.5)--(4.33013,6.5);
\end{tikzpicture}

%% file: hexagonlight_0_modified_cropped.txt
\begin{tikzpicture}[scale = 0.14,rotate = -60]
\draw [gray] (11.2583,15.5) -- (12.1244,15);
\draw [gray] (11.2583,15.5) -- (12.1244,16);
\draw [gray] (11.2583,15.5) -- (11.2583,16.5);
\draw [gray] (11.2583,16.5) -- (12.1244,16);
\draw [gray] (11.2583,16.5) -- (12.1244,17);
\draw [gray] (11.2583,16.5) -- (11.2583,17.5);
\draw [gray] (11.2583,17.5) -- (12.1244,17);
\draw [gray] (11.2583,17.5) -- (12.1244,18);
\draw [gray] (11.2583,17.5) -- (11.2583,18.5);
\draw [gray] (11.2583,18.5) -- (12.1244,18);
\draw [gray] (11.2583,18.5) -- (12.1244,19);
\draw [gray] (11.2583,18.5) -- (11.2583,19.5);
\draw [gray] (11.2583,19.5) -- (12.1244,19);
\draw [gray] (11.2583,19.5) -- (12.1244,20);
\draw [gray] (11.2583,19.5) -- (11.2583,20.5);
\draw [gray] (11.2583,20.5) -- (12.1244,20);
\draw [gray] (11.2583,20.5) -- (12.1244,21);
\draw [gray] (12.1244,15) -- (12.9904,14.5);
\draw [gray] (12.1244,15) -- (12.9904,15.5);
\draw [gray] (12.1244,15) -- (12.1244,16);
\draw [gray] (12.1244,16) -- (12.9904,15.5);
\draw [gray] (12.1244,16) -- (12.9904,16.5);
\draw [gray] (12.1244,16) -- (12.1244,17);
\draw [gray] (12.1244,17) -- (12.9904,16.5);
\draw [gray] (12.1244,17) -- (12.9904,17.5);
\draw [gray] (12.1244,17) -- (12.1244,18);
\draw [gray] (12.1244,18) -- (12.9904,17.5);
\draw [gray] (12.1244,18) -- (12.9904,18.5);
\draw [gray] (12.1244,18) -- (12.1244,19);
\draw [gray] (12.1244,19) -- (12.9904,18.5);
\draw [gray] (12.1244,19) -- (12.9904,19.5);
\draw [gray] (12.1244,19) -- (12.1244,20);
\draw [gray] (12.1244,20) -- (12.9904,19.5);
\draw [gray] (12.1244,20) -- (12.9904,20.5);
\draw [gray] (12.1244,20) -- (12.1244,21);
\draw [gray] (12.1244,21) -- (12.9904,20.5);
\draw [gray] (12.1244,21) -- (12.9904,21.5);
\draw [gray] (12.9904,14.5) -- (13.8564,14);
\draw [gray] (12.9904,14.5) -- (13.8564,15);
\draw [gray] (12.9904,14.5) -- (12.9904,15.5);
\draw [gray] (12.9904,15.5) -- (13.8564,15);
\draw [gray] (12.9904,15.5) -- (13.8564,16);
\draw [gray] (12.9904,15.5) -- (12.9904,16.5);
\draw [gray] (12.9904,16.5) -- (13.8564,16);
\draw [gray] (12.9904,16.5) -- (13.8564,17);
\draw [gray] (12.9904,16.5) -- (12.9904,17.5);
\draw [gray] (12.9904,17.5) -- (13.8564,17);
\draw [gray] (12.9904,17.5) -- (13.8564,18);
\draw [gray] (12.9904,17.5) -- (12.9904,18.5);
\draw [gray] (12.9904,18.5) -- (13.8564,18);
\draw [gray] (12.9904,18.5) -- (13.8564,19);
\draw [gray] (12.9904,18.5) -- (12.9904,19.5);
\draw [gray] (12.9904,19.5) -- (13.8564,19);
\draw [gray] (12.9904,19.5) -- (13.8564,20);
\draw [gray] (12.9904,19.5) -- (12.9904,20.5);
\draw [gray] (12.9904,20.5) -- (13.8564,20);
\draw [gray] (12.9904,20.5) -- (13.8564,21);
\draw [gray] (12.9904,20.5) -- (12.9904,21.5);
\draw [gray] (12.9904,21.5) -- (13.8564,21);
\draw [gray] (12.9904,21.5) -- (13.8564,22);
\draw [gray] (13.8564,14) -- (14.7224,13.5);
\draw [gray] (13.8564,14) -- (14.7224,14.5);
\draw [gray] (13.8564,14) -- (13.8564,15);
\draw [gray] (13.8564,15) -- (14.7224,14.5);
\draw [gray] (13.8564,15) -- (14.7224,15.5);
\draw [gray] (13.8564,15) -- (13.8564,16);
\draw [gray] (13.8564,16) -- (14.7224,15.5);
\draw [gray] (13.8564,16) -- (14.7224,16.5);
\draw [gray] (13.8564,16) -- (13.8564,17);
\draw [gray] (13.8564,17) -- (14.7224,16.5);
\draw [gray] (13.8564,17) -- (14.7224,17.5);
\draw [gray] (13.8564,17) -- (13.8564,18);
\draw [gray] (13.8564,18) -- (14.7224,17.5);
\draw [gray] (13.8564,18) -- (14.7224,18.5);
\draw [gray] (13.8564,18) -- (13.8564,19);
\draw [gray] (13.8564,19) -- (14.7224,18.5);
\draw [gray] (13.8564,19) -- (14.7224,19.5);
\draw [gray] (13.8564,19) -- (13.8564,20);
\draw [gray] (13.8564,20) -- (14.7224,19.5);
\draw [gray] (13.8564,20) -- (14.7224,20.5);
\draw [gray] (13.8564,20) -- (13.8564,21);
\draw [gray] (13.8564,21) -- (14.7224,20.5);
\draw [gray] (13.8564,21) -- (14.7224,21.5);
\draw [gray] (13.8564,21) -- (13.8564,22);
\draw [gray] (13.8564,22) -- (14.7224,21.5);
\draw [gray] (13.8564,22) -- (14.7224,22.5);
\draw [gray] (14.7224,13.5) -- (15.5885,13);
\draw [gray] (14.7224,13.5) -- (15.5885,14);
\draw [gray] (14.7224,13.5) -- (14.7224,14.5);
\draw [gray] (14.7224,14.5) -- (15.5885,14);
\draw [gray] (14.7224,14.5) -- (15.5885,15);
\draw [gray] (14.7224,14.5) -- (14.7224,15.5);
\draw [gray] (14.7224,15.5) -- (15.5885,15);
\draw [gray] (14.7224,15.5) -- (15.5885,16);
\draw [gray] (14.7224,15.5) -- (14.7224,16.5);
\draw [gray] (14.7224,16.5) -- (15.5885,16);
\draw [gray] (14.7224,16.5) -- (15.5885,17);
\draw [gray] (14.7224,16.5) -- (14.7224,17.5);
\draw [gray] (14.7224,17.5) -- (15.5885,17);
\draw [gray] (14.7224,17.5) -- (15.5885,18);
\draw [gray] (14.7224,17.5) -- (14.7224,18.5);
\draw [gray] (14.7224,18.5) -- (15.5885,18);
\draw [gray] (14.7224,18.5) -- (15.5885,19);
\draw [gray] (14.7224,18.5) -- (14.7224,19.5);
\draw [gray] (14.7224,19.5) -- (15.5885,19);
\draw [gray] (14.7224,19.5) -- (15.5885,20);
\draw [gray] (14.7224,19.5) -- (14.7224,20.5);
\draw [gray] (14.7224,20.5) -- (15.5885,20);
\draw [gray] (14.7224,20.5) -- (15.5885,21);
\draw [gray] (14.7224,20.5) -- (14.7224,21.5);
\draw [gray] (14.7224,21.5) -- (15.5885,21);
\draw [gray] (14.7224,21.5) -- (15.5885,22);
\draw [gray] (14.7224,21.5) -- (14.7224,22.5);
\draw [gray] (14.7224,22.5) -- (15.5885,22);
\draw [gray] (14.7224,22.5) -- (15.5885,23);
\draw [gray] (15.5885,13) -- (16.4545,13.5);
\draw [gray] (15.5885,13) -- (15.5885,14);
\draw [gray] (15.5885,14) -- (16.4545,13.5);
\draw [gray] (15.5885,14) -- (16.4545,14.5);
\draw [gray] (15.5885,14) -- (15.5885,15);
\draw [gray] (15.5885,15) -- (16.4545,14.5);
\draw [gray] (15.5885,15) -- (16.4545,15.5);
\draw [gray] (15.5885,15) -- (15.5885,16);
\draw [gray] (15.5885,16) -- (16.4545,15.5);
\draw [gray] (15.5885,16) -- (16.4545,16.5);
\draw [gray] (15.5885,16) -- (15.5885,17);
\draw [gray] (15.5885,17) -- (16.4545,16.5);
\draw [gray] (15.5885,17) -- (16.4545,17.5);
\draw [gray] (15.5885,17) -- (15.5885,18);
\draw [gray] (15.5885,18) -- (16.4545,17.5);
\draw [gray] (15.5885,18) -- (16.4545,18.5);
\draw [gray] (15.5885,18) -- (15.5885,19);
\draw [gray] (15.5885,19) -- (16.4545,18.5);
\draw [gray] (15.5885,19) -- (16.4545,19.5);
\draw [gray] (15.5885,19) -- (15.5885,20);
\draw [gray] (15.5885,20) -- (16.4545,19.5);
\draw [gray] (15.5885,20) -- (16.4545,20.5);
\draw [gray] (15.5885,20) -- (15.5885,21);
\draw [gray] (15.5885,21) -- (16.4545,20.5);
\draw [gray] (15.5885,21) -- (16.4545,21.5);
\draw [gray] (15.5885,21) -- (15.5885,22);
\draw [gray] (15.5885,22) -- (16.4545,21.5);
\draw [gray] (15.5885,22) -- (16.4545,22.5);
\draw [gray] (15.5885,22) -- (15.5885,23);
\draw [gray] (15.5885,23) -- (16.4545,22.5);
\draw [gray] (16.4545,13.5) -- (17.3205,14);
\draw [gray] (16.4545,13.5) -- (16.4545,14.5);
\draw [gray] (16.4545,14.5) -- (17.3205,14);
\draw [gray] (16.4545,14.5) -- (17.3205,15);
\draw [gray] (16.4545,14.5) -- (16.4545,15.5);
\draw [gray] (16.4545,15.5) -- (17.3205,15);
\draw [gray] (16.4545,15.5) -- (17.3205,16);
\draw [gray] (16.4545,15.5) -- (16.4545,16.5);
\draw [gray] (16.4545,16.5) -- (17.3205,16);
\draw [gray] (16.4545,16.5) -- (17.3205,17);
\draw [gray] (16.4545,16.5) -- (16.4545,17.5);
\draw [gray] (16.4545,17.5) -- (17.3205,17);
\draw [gray] (16.4545,17.5) -- (17.3205,18);
\draw [gray] (16.4545,17.5) -- (16.4545,18.5);
\draw [gray] (16.4545,18.5) -- (17.3205,18);
\draw [gray] (16.4545,18.5) -- (17.3205,19);
\draw [gray] (16.4545,18.5) -- (16.4545,19.5);
\draw [gray] (16.4545,19.5) -- (17.3205,19);
\draw [gray] (16.4545,19.5) -- (17.3205,20);
\draw [gray] (16.4545,19.5) -- (16.4545,20.5);
\draw [gray] (16.4545,20.5) -- (17.3205,20);
\draw [gray] (16.4545,20.5) -- (17.3205,21);
\draw [gray] (16.4545,20.5) -- (16.4545,21.5);
\draw [gray] (16.4545,21.5) -- (17.3205,21);
\draw [gray] (16.4545,21.5) -- (17.3205,22);
\draw [gray] (16.4545,21.5) -- (16.4545,22.5);
\draw [gray] (16.4545,22.5) -- (17.3205,22);
\draw [gray] (17.3205,14) -- (18.1865,14.5);
\draw [gray] (17.3205,14) -- (17.3205,15);
\draw [gray] (17.3205,15) -- (18.1865,14.5);
\draw [gray] (17.3205,15) -- (18.1865,15.5);
\draw [gray] (17.3205,15) -- (17.3205,16);
\draw [gray] (17.3205,16) -- (18.1865,15.5);
\draw [gray] (17.3205,16) -- (18.1865,16.5);
\draw [gray] (17.3205,16) -- (17.3205,17);
\draw [gray] (17.3205,17) -- (18.1865,16.5);
\draw [gray] (17.3205,17) -- (18.1865,17.5);
\draw [gray] (17.3205,17) -- (17.3205,18);
\draw [gray] (17.3205,18) -- (18.1865,17.5);
\draw [gray] (17.3205,18) -- (18.1865,18.5);
\draw [gray] (17.3205,18) -- (17.3205,19);
\draw [gray] (17.3205,19) -- (18.1865,18.5);
\draw [gray] (17.3205,19) -- (18.1865,19.5);
\draw [gray] (17.3205,19) -- (17.3205,20);
\draw [gray] (17.3205,20) -- (18.1865,19.5);
\draw [gray] (17.3205,20) -- (18.1865,20.5);
\draw [gray] (17.3205,20) -- (17.3205,21);
\draw [gray] (17.3205,21) -- (18.1865,20.5);
\draw [gray] (17.3205,21) -- (18.1865,21.5);
\draw [gray] (17.3205,21) -- (17.3205,22);
\draw [gray] (17.3205,22) -- (18.1865,21.5);
\draw [gray] (18.1865,14.5) -- (19.0526,15);
\draw [gray] (18.1865,14.5) -- (18.1865,15.5);
\draw [gray] (18.1865,15.5) -- (19.0526,15);
\draw [gray] (18.1865,15.5) -- (19.0526,16);
\draw [gray] (18.1865,15.5) -- (18.1865,16.5);
\draw [gray] (18.1865,16.5) -- (19.0526,16);
\draw [gray] (18.1865,16.5) -- (19.0526,17);
\draw [gray] (18.1865,16.5) -- (18.1865,17.5);
\draw [gray] (18.1865,17.5) -- (19.0526,17);
\draw [gray] (18.1865,17.5) -- (19.0526,18);
\draw [gray] (18.1865,17.5) -- (18.1865,18.5);
\draw [gray] (18.1865,18.5) -- (19.0526,18);
\draw [gray] (18.1865,18.5) -- (19.0526,19);
\draw [gray] (18.1865,18.5) -- (18.1865,19.5);
\draw [gray] (18.1865,19.5) -- (19.0526,19);
\draw [gray] (18.1865,19.5) -- (19.0526,20);
\draw [gray] (18.1865,19.5) -- (18.1865,20.5);
\draw [gray] (18.1865,20.5) -- (19.0526,20);
\draw [gray] (18.1865,20.5) -- (19.0526,21);
\draw [gray] (18.1865,20.5) -- (18.1865,21.5);
\draw [gray] (18.1865,21.5) -- (19.0526,21);
\draw [gray] (19.0526,15) -- (19.9186,15.5);
\draw [gray] (19.0526,15) -- (19.0526,16);
\draw [gray] (19.0526,16) -- (19.9186,15.5);
\draw [gray] (19.0526,16) -- (19.9186,16.5);
\draw [gray] (19.0526,16) -- (19.0526,17);
\draw [gray] (19.0526,17) -- (19.9186,16.5);
\draw [gray] (19.0526,17) -- (19.9186,17.5);
\draw [gray] (19.0526,17) -- (19.0526,18);
\draw [gray] (19.0526,18) -- (19.9186,17.5);
\draw [gray] (19.0526,18) -- (19.9186,18.5);
\draw [gray] (19.0526,18) -- (19.0526,19);
\draw [gray] (19.0526,19) -- (19.9186,18.5);
\draw [gray] (19.0526,19) -- (19.9186,19.5);
\draw [gray] (19.0526,19) -- (19.0526,20);
\draw [gray] (19.0526,20) -- (19.9186,19.5);
\draw [gray] (19.0526,20) -- (19.9186,20.5);
\draw [gray] (19.0526,20) -- (19.0526,21);
\draw [gray] (19.0526,21) -- (19.9186,20.5);
\draw [gray] (19.9186,15.5) -- (19.9186,16.5);
\draw [gray] (19.9186,16.5) -- (19.9186,17.5);
\draw [gray] (19.9186,17.5) -- (19.9186,18.5);
\draw [gray] (19.9186,18.5) -- (19.9186,19.5);
\draw [gray] (19.9186,19.5) -- (19.9186,20.5);
\draw [black,fill=] (11.2583,15.5) circle [radius = 0.2];
\draw [black,fill=] (11.2583,16.5) circle [radius = 0.2];
\draw [black,fill=] (11.2583,17.5) circle [radius = 0.2];
\draw [black,fill=] (11.2583,18.5) circle [radius = 0.2];
\draw [black,fill=] (11.2583,19.5) circle [radius = 0.2];
\draw [black,fill=] (11.2583,20.5) circle [radius = 0.2];
\draw [black,fill=] (12.1244,15) circle [radius = 0.2];
\draw [black,fill=] (12.1244,16) circle [radius = 0.2];
\draw [black,fill=] (12.1244,17) circle [radius = 0.2];
\draw [black,fill=] (12.1244,18) circle [radius = 0.2];
\draw [black,fill=] (12.1244,19) circle [radius = 0.2];
\draw [black,fill=] (12.1244,20) circle [radius = 0.2];
\draw [black,fill=] (12.1244,21) circle [radius = 0.2];
\draw [black,fill=] (12.9904,14.5) circle [radius = 0.2];
\draw [black,fill=] (12.9904,15.5) circle [radius = 0.2];
\draw [black,fill=] (12.9904,16.5) circle [radius = 0.2];
\draw [black,fill=] (12.9904,17.5) circle [radius = 0.2];
\draw [black,fill=] (12.9904,18.5) circle [radius = 0.2];
\draw [black,fill=] (12.9904,19.5) circle [radius = 0.2];
\draw [black,fill=] (12.9904,20.5) circle [radius = 0.2];
\draw [black,fill=] (12.9904,21.5) circle [radius = 0.2];
\draw [black,fill=] (13.8564,14) circle [radius = 0.2];
\draw [black,fill=] (13.8564,15) circle [radius = 0.2];
\draw [black,fill=] (13.8564,16) circle [radius = 0.2];
\draw [black,fill=] (13.8564,17) circle [radius = 0.2];
\draw [black,fill=] (13.8564,18) circle [radius = 0.2];
\draw [black,fill=] (13.8564,19) circle [radius = 0.2];
\draw [black,fill=] (13.8564,20) circle [radius = 0.2];
\draw [black,fill=] (13.8564,21) circle [radius = 0.2];
\draw [black,fill=] (13.8564,22) circle [radius = 0.2];
\draw [black,fill=] (14.7224,13.5) circle [radius = 0.2];
\draw [black,fill=] (14.7224,14.5) circle [radius = 0.2];
\draw [black,fill=] (14.7224,15.5) circle [radius = 0.2];
\draw [black,fill=] (14.7224,16.5) circle [radius = 0.2];
\draw [black,fill=] (14.7224,17.5) circle [radius = 0.2];
\draw [black,fill=] (14.7224,18.5) circle [radius = 0.2];
\draw [black,fill=] (14.7224,19.5) circle [radius = 0.2];
\draw [black,fill=] (14.7224,20.5) circle [radius = 0.2];
\draw [black,fill=] (14.7224,21.5) circle [radius = 0.2];
\draw [black,fill=] (14.7224,22.5) circle [radius = 0.2];
\draw [black,fill=] (15.5885,13) circle [radius = 0.2];
\draw [black,fill=] (15.5885,14) circle [radius = 0.2];
\draw [black,fill=] (15.5885,15) circle [radius = 0.2];
\draw [black,fill=] (15.5885,16) circle [radius = 0.2];
\draw [black,fill=] (15.5885,17) circle [radius = 0.2];
\draw [black,fill=] (15.5885,18) circle [radius = 0.2];
\draw [black,fill=] (15.5885,19) circle [radius = 0.2];
\draw [black,fill=] (15.5885,20) circle [radius = 0.2];
\draw [black,fill=] (15.5885,21) circle [radius = 0.2];
\draw [black,fill=] (15.5885,22) circle [radius = 0.2];
\draw [black,fill=] (15.5885,23) circle [radius = 0.2];
\draw [black,fill=] (16.4545,13.5) circle [radius = 0.2];
\draw [black,fill=] (16.4545,14.5) circle [radius = 0.2];
\draw [black,fill=] (16.4545,15.5) circle [radius = 0.2];
\draw [black,fill=] (16.4545,16.5) circle [radius = 0.2];
\draw [black,fill=] (16.4545,17.5) circle [radius = 0.2];
\draw [black,fill=] (16.4545,18.5) circle [radius = 0.2];
\draw [black,fill=] (16.4545,19.5) circle [radius = 0.2];
\draw [black,fill=] (16.4545,20.5) circle [radius = 0.2];
\draw [black,fill=] (16.4545,21.5) circle [radius = 0.2];
\draw [black,fill=] (16.4545,22.5) circle [radius = 0.2];
\draw [black,fill=] (17.3205,14) circle [radius = 0.2];
\draw [black,fill=] (17.3205,15) circle [radius = 0.2];
\draw [black,fill=] (17.3205,16) circle [radius = 0.2];
\draw [black,fill=] (17.3205,17) circle [radius = 0.2];
\draw [black,fill=] (17.3205,18) circle [radius = 0.2];
\draw [black,fill=] (17.3205,19) circle [radius = 0.2];
\draw [black,fill=] (17.3205,20) circle [radius = 0.2];
\draw [black,fill=] (17.3205,21) circle [radius = 0.2];
\draw [black,fill=] (17.3205,22) circle [radius = 0.2];
\draw [black,fill=] (18.1865,14.5) circle [radius = 0.2];
\draw [black,fill=] (18.1865,15.5) circle [radius = 0.2];
\draw [black,fill=] (18.1865,16.5) circle [radius = 0.2];
\draw [black,fill=] (18.1865,17.5) circle [radius = 0.2];
\draw [black,fill=] (18.1865,18.5) circle [radius = 0.2];
\draw [black,fill=] (18.1865,19.5) circle [radius = 0.2];
\draw [black,fill=] (18.1865,20.5) circle [radius = 0.2];
\draw [black,fill=] (18.1865,21.5) circle [radius = 0.2];
\draw [black,fill=] (19.0526,15) circle [radius = 0.2];
\draw [black,fill=] (19.0526,16) circle [radius = 0.2];
\draw [black,fill=] (19.0526,17) circle [radius = 0.2];
\draw [black,fill=] (19.0526,18) circle [radius = 0.2];
\draw [black,fill=] (19.0526,19) circle [radius = 0.2];
\draw [black,fill=] (19.0526,20) circle [radius = 0.2];
\draw [black,fill=] (19.0526,21) circle [radius = 0.2];
\draw [black,fill=] (19.9186,15.5) circle [radius = 0.2];
\draw [black,fill=] (19.9186,16.5) circle [radius = 0.2];
\draw [black,fill=] (19.9186,17.5) circle [radius = 0.2];
\draw [black,fill=] (19.9186,18.5) circle [radius = 0.2];
\draw [black,fill=] (19.9186,19.5) circle [radius = 0.2];
\draw [black,fill=] (19.9186,20.5) circle [radius = 0.2];
\draw [gray] (20.7846,9) -- (20.7846,8);
\draw [gray] (21.6506,10.5) -- (21.6506,9.5);
\draw [gray] (22.5167,12) -- (22.5167,11);
\draw [gray] (23.3827,13.5) -- (23.3827,12.5);
\draw [gray] (24.2487,15) -- (24.2487,14);
\draw [gray] (25.1147,16.5) -- (25.1147,15.5);
\draw [gray] (25.9808,18) -- (25.9808,17);
\draw [gray] (20.7846,9) -- (21.6506,9.5);
\draw [gray] (21.6506,10.5) -- (22.5167,11);
\draw [gray] (22.5167,12) -- (23.3827,12.5);
\draw [gray] (23.3827,13.5) -- (24.2487,14);
\draw [gray] (24.2487,15) -- (25.1147,15.5);
\draw [gray] (25.1147,16.5) -- (25.9808,17);
\draw [gray] (25.9808,18) -- (26.8468,18.5);
\draw [red,fill=] (20.7846,8) circle [radius = 0.2];
\draw [red,fill=] (20.7846,9) circle [radius = 0.2];
\draw [red,fill=] (21.6506,9.5) circle [radius = 0.2];
\draw [red,fill=] (21.6506,10.5) circle [radius = 0.2];
\draw [red,fill=] (22.5167,11) circle [radius = 0.2];
\draw [red,fill=] (22.5167,12) circle [radius = 0.2];
\draw [red,fill=] (23.3827,12.5) circle [radius = 0.2];
\draw [red,fill=] (23.3827,13.5) circle [radius = 0.2];
\draw [red,fill=] (24.2487,14) circle [radius = 0.2];
\draw [red,fill=] (24.2487,15) circle [radius = 0.2];
\draw [red,fill=] (25.1147,15.5) circle [radius = 0.2];
\draw [red,fill=] (25.1147,16.5) circle [radius = 0.2];
\draw [red,fill=] (25.9808,17) circle [radius = 0.2];
\draw [red,fill=] (25.9808,18) circle [radius = 0.2];
\draw [red,fill=] (26.8468,18.5) circle [radius = 0.2];
\end{tikzpicture}

%% file: hexagonlight_10000000_modified_cropped.txt
\begin{tikzpicture}[scale = 0.14, rotate = -60]
\draw [gray] (12.9904,22.5) -- (13.8564,22);
\draw [gray] (12.9904,22.5) -- (13.8564,23);
\draw [gray] (12.9904,22.5) -- (12.9904,23.5);
\draw [gray] (9.52628,25.5) -- (10.3923,25);
\draw [gray] (9.52628,25.5) -- (10.3923,26);
\draw [gray] (7.79423,25.5) -- (8.66025,25);
\draw [gray] (7.79423,25.5) -- (8.66025,26);
\draw [gray] (13.8564,19) -- (14.7224,18.5);
\draw [gray] (13.8564,19) -- (14.7224,19.5);
\draw [gray] (13.8564,19) -- (13.8564,20);
\draw [gray] (17.3205,19) -- (18.1865,18.5);
\draw [gray] (17.3205,19) -- (18.1865,19.5);
\draw [gray] (17.3205,19) -- (17.3205,20);
\draw [gray] (8.66025,25) -- (9.52628,24.5);
\draw [gray] (8.66025,25) -- (9.52628,25.5);
\draw [gray] (8.66025,25) -- (8.66025,26);
\draw [gray] (12.9904,20.5) -- (13.8564,20);
\draw [gray] (12.9904,20.5) -- (13.8564,21);
\draw [gray] (12.9904,20.5) -- (12.9904,21.5);
\draw [gray] (13.8564,23) -- (14.7224,22.5);
\draw [gray] (13.8564,23) -- (14.7224,23.5);
\draw [gray] (13.8564,23) -- (13.8564,24);
\draw [gray] (11.2583,22.5) -- (12.1244,22);
\draw [gray] (11.2583,22.5) -- (12.1244,23);
\draw [gray] (11.2583,22.5) -- (11.2583,23.5);
\draw [gray] (11.2583,21.5) -- (12.1244,21);
\draw [gray] (11.2583,21.5) -- (12.1244,22);
\draw [gray] (11.2583,21.5) -- (11.2583,22.5);
\draw [gray] (10.3923,21) -- (11.2583,20.5);
\draw [gray] (10.3923,21) -- (11.2583,21.5);
\draw [gray] (10.3923,21) -- (10.3923,22);
\draw [gray] (13.8564,17) -- (14.7224,16.5);
\draw [gray] (13.8564,17) -- (14.7224,17.5);
\draw [gray] (13.8564,17) -- (13.8564,18);
\draw [gray] (14.7224,21.5) -- (15.5885,21);
\draw [gray] (14.7224,21.5) -- (15.5885,22);
\draw [gray] (14.7224,21.5) -- (14.7224,22.5);
\draw [gray] (12.9904,18.5) -- (13.8564,18);
\draw [gray] (12.9904,18.5) -- (13.8564,19);
\draw [gray] (12.9904,18.5) -- (12.9904,19.5);
\draw [gray] (10.3923,22) -- (11.2583,21.5);
\draw [gray] (10.3923,22) -- (11.2583,22.5);
\draw [gray] (10.3923,22) -- (10.3923,23);
\draw [gray] (14.7224,16.5) -- (15.5885,16);
\draw [gray] (14.7224,16.5) -- (15.5885,17);
\draw [gray] (14.7224,16.5) -- (14.7224,17.5);
\draw [gray] (9.52628,23.5) -- (10.3923,23);
\draw [gray] (9.52628,23.5) -- (10.3923,24);
\draw [gray] (9.52628,23.5) -- (9.52628,24.5);
\draw [gray] (18.1865,20.5) -- (19.0526,20);
\draw [gray] (18.1865,20.5) -- (18.1865,21.5);
\draw [gray] (14.7224,20.5) -- (15.5885,20);
\draw [gray] (14.7224,20.5) -- (15.5885,21);
\draw [gray] (14.7224,20.5) -- (14.7224,21.5);
\draw [gray] (9.52628,24.5) -- (10.3923,24);
\draw [gray] (9.52628,24.5) -- (10.3923,25);
\draw [gray] (9.52628,24.5) -- (9.52628,25.5);
\draw [gray] (12.9904,19.5) -- (13.8564,19);
\draw [gray] (12.9904,19.5) -- (13.8564,20);
\draw [gray] (12.9904,19.5) -- (12.9904,20.5);
\draw [gray] (17.3205,21) -- (18.1865,20.5);
\draw [gray] (17.3205,21) -- (18.1865,21.5);
\draw [gray] (17.3205,21) -- (17.3205,22);
\draw [gray] (15.5885,16) -- (16.4545,16.5);
\draw [gray] (15.5885,16) -- (15.5885,17);
\draw [gray] (12.1244,23) -- (12.9904,22.5);
\draw [gray] (12.1244,23) -- (12.9904,23.5);
\draw [gray] (12.1244,23) -- (12.1244,24);
\draw [gray] (12.1244,25) -- (12.9904,24.5);
\draw [gray] (18.1865,17.5) -- (19.0526,18);
\draw [gray] (18.1865,17.5) -- (18.1865,18.5);
\draw [gray] (12.9904,17.5) -- (13.8564,17);
\draw [gray] (12.9904,17.5) -- (13.8564,18);
\draw [gray] (12.9904,17.5) -- (12.9904,18.5);
\draw [gray] (11.2583,18.5) -- (12.1244,18);
\draw [gray] (11.2583,18.5) -- (12.1244,19);
\draw [gray] (11.2583,18.5) -- (11.2583,19.5);
\draw [gray] (11.2583,24.5) -- (12.1244,24);
\draw [gray] (11.2583,24.5) -- (12.1244,25);
\draw [gray] (11.2583,24.5) -- (11.2583,25.5);
\draw [gray] (17.3205,17) -- (18.1865,17.5);
\draw [gray] (17.3205,17) -- (17.3205,18);
\draw [gray] (10.3923,25) -- (11.2583,24.5);
\draw [gray] (10.3923,25) -- (11.2583,25.5);
\draw [gray] (10.3923,25) -- (10.3923,26);
\draw [gray] (15.5885,18) -- (16.4545,17.5);
\draw [gray] (15.5885,18) -- (16.4545,18.5);
\draw [gray] (15.5885,18) -- (15.5885,19);
\draw [gray] (12.1244,20) -- (12.9904,19.5);
\draw [gray] (12.1244,20) -- (12.9904,20.5);
\draw [gray] (12.1244,20) -- (12.1244,21);
\draw [gray] (18.1865,19.5) -- (19.0526,19);
\draw [gray] (18.1865,19.5) -- (19.0526,20);
\draw [gray] (18.1865,19.5) -- (18.1865,20.5);
\draw [gray] (12.1244,22) -- (12.9904,21.5);
\draw [gray] (12.1244,22) -- (12.9904,22.5);
\draw [gray] (12.1244,22) -- (12.1244,23);
\draw [gray] (12.9904,16.5) -- (13.8564,16);
\draw [gray] (12.9904,16.5) -- (13.8564,17);
\draw [gray] (12.9904,16.5) -- (12.9904,17.5);
\draw [gray] (10.3923,26) -- (11.2583,25.5);
\draw [gray] (10.3923,26) -- (10.3923,27);
\draw [gray] (14.7224,22.5) -- (15.5885,22);
\draw [gray] (14.7224,22.5) -- (15.5885,23);
\draw [gray] (14.7224,22.5) -- (14.7224,23.5);
\draw [gray] (15.5885,21) -- (16.4545,20.5);
\draw [gray] (15.5885,21) -- (16.4545,21.5);
\draw [gray] (15.5885,21) -- (15.5885,22);
\draw [gray] (16.4545,16.5) -- (17.3205,17);
\draw [gray] (16.4545,16.5) -- (16.4545,17.5);
\draw [gray] (16.4545,21.5) -- (17.3205,21);
\draw [gray] (16.4545,21.5) -- (17.3205,22);
\draw [gray] (16.4545,21.5) -- (16.4545,22.5);
\draw [gray] (11.2583,20.5) -- (12.1244,20);
\draw [gray] (11.2583,20.5) -- (12.1244,21);
\draw [gray] (11.2583,20.5) -- (11.2583,21.5);
\draw [gray] (12.9904,21.5) -- (13.8564,21);
\draw [gray] (12.9904,21.5) -- (13.8564,22);
\draw [gray] (12.9904,21.5) -- (12.9904,22.5);
\draw [gray] (17.3205,20) -- (18.1865,19.5);
\draw [gray] (17.3205,20) -- (18.1865,20.5);
\draw [gray] (17.3205,20) -- (17.3205,21);
\draw [gray] (14.7224,18.5) -- (15.5885,18);
\draw [gray] (14.7224,18.5) -- (15.5885,19);
\draw [gray] (14.7224,18.5) -- (14.7224,19.5);
\draw [gray] (16.4545,18.5) -- (17.3205,18);
\draw [gray] (16.4545,18.5) -- (17.3205,19);
\draw [gray] (16.4545,18.5) -- (16.4545,19.5);
\draw [gray] (14.7224,15.5) -- (15.5885,16);
\draw [gray] (14.7224,15.5) -- (14.7224,16.5);
\draw [gray] (10.3923,24) -- (11.2583,23.5);
\draw [gray] (10.3923,24) -- (11.2583,24.5);
\draw [gray] (10.3923,24) -- (10.3923,25);
\draw [gray] (13.8564,16) -- (14.7224,15.5);
\draw [gray] (13.8564,16) -- (14.7224,16.5);
\draw [gray] (13.8564,16) -- (13.8564,17);
\draw [gray] (16.4545,19.5) -- (17.3205,19);
\draw [gray] (16.4545,19.5) -- (17.3205,20);
\draw [gray] (16.4545,19.5) -- (16.4545,20.5);
\draw [gray] (13.8564,18) -- (14.7224,17.5);
\draw [gray] (13.8564,18) -- (14.7224,18.5);
\draw [gray] (13.8564,18) -- (13.8564,19);
\draw [gray] (13.8564,25) -- (14.7224,24.5);
\draw [gray] (15.5885,22) -- (16.4545,21.5);
\draw [gray] (15.5885,22) -- (16.4545,22.5);
\draw [gray] (15.5885,22) -- (15.5885,23);
\draw [gray] (12.1244,19) -- (12.9904,18.5);
\draw [gray] (12.1244,19) -- (12.9904,19.5);
\draw [gray] (12.1244,19) -- (12.1244,20);
\draw [gray] (13.8564,20) -- (14.7224,19.5);
\draw [gray] (13.8564,20) -- (14.7224,20.5);
\draw [gray] (13.8564,20) -- (13.8564,21);
\draw [gray] (18.1865,18.5) -- (19.0526,18);
\draw [gray] (18.1865,18.5) -- (19.0526,19);
\draw [gray] (18.1865,18.5) -- (18.1865,19.5);
\draw [gray] (19.0526,19) -- (19.0526,20);
\draw [gray] (10.3923,19) -- (11.2583,18.5);
\draw [gray] (10.3923,19) -- (11.2583,19.5);
\draw [gray] (10.3923,19) -- (10.3923,20);
\draw [gray] (14.7224,17.5) -- (15.5885,17);
\draw [gray] (14.7224,17.5) -- (15.5885,18);
\draw [gray] (14.7224,17.5) -- (14.7224,18.5);
\draw [gray] (13.8564,21) -- (14.7224,20.5);
\draw [gray] (13.8564,21) -- (14.7224,21.5);
\draw [gray] (13.8564,21) -- (13.8564,22);
\draw [gray] (19.0526,18) -- (19.0526,19);
\draw [gray] (9.52628,19.5) -- (10.3923,19);
\draw [gray] (9.52628,19.5) -- (10.3923,20);
\draw [gray] (12.1244,18) -- (12.9904,17.5);
\draw [gray] (12.1244,18) -- (12.9904,18.5);
\draw [gray] (12.1244,18) -- (12.1244,19);
\draw [gray] (16.4545,20.5) -- (17.3205,20);
\draw [gray] (16.4545,20.5) -- (17.3205,21);
\draw [gray] (16.4545,20.5) -- (16.4545,21.5);
\draw [gray] (17.3205,22) -- (18.1865,21.5);
\draw [gray] (15.5885,19) -- (16.4545,18.5);
\draw [gray] (15.5885,19) -- (16.4545,19.5);
\draw [gray] (15.5885,19) -- (15.5885,20);
\draw [gray] (11.2583,23.5) -- (12.1244,23);
\draw [gray] (11.2583,23.5) -- (12.1244,24);
\draw [gray] (11.2583,23.5) -- (11.2583,24.5);
\draw [gray] (17.3205,18) -- (18.1865,17.5);
\draw [gray] (17.3205,18) -- (18.1865,18.5);
\draw [gray] (17.3205,18) -- (17.3205,19);
\draw [gray] (12.9904,24.5) -- (13.8564,24);
\draw [gray] (12.9904,24.5) -- (13.8564,25);
\draw [gray] (14.7224,23.5) -- (15.5885,23);
\draw [gray] (14.7224,23.5) -- (14.7224,24.5);
\draw [gray] (15.5885,20) -- (16.4545,19.5);
\draw [gray] (15.5885,20) -- (16.4545,20.5);
\draw [gray] (15.5885,20) -- (15.5885,21);
\draw [gray] (13.8564,24) -- (14.7224,23.5);
\draw [gray] (13.8564,24) -- (14.7224,24.5);
\draw [gray] (13.8564,24) -- (13.8564,25);
\draw [gray] (16.4545,22.5) -- (17.3205,22);
\draw [gray] (16.4545,22.5) -- (16.4545,23.5);
\draw [gray] (11.2583,25.5) -- (12.1244,25);
\draw [gray] (13.8564,22) -- (14.7224,21.5);
\draw [gray] (13.8564,22) -- (14.7224,22.5);
\draw [gray] (13.8564,22) -- (13.8564,23);
\draw [gray] (15.5885,23) -- (16.4545,22.5);
\draw [gray] (15.5885,23) -- (16.4545,23.5);
\draw [gray] (12.1244,21) -- (12.9904,20.5);
\draw [gray] (12.1244,21) -- (12.9904,21.5);
\draw [gray] (12.1244,21) -- (12.1244,22);
\draw [gray] (12.1244,24) -- (12.9904,23.5);
\draw [gray] (12.1244,24) -- (12.9904,24.5);
\draw [gray] (12.1244,24) -- (12.1244,25);
\draw [gray] (10.3923,20) -- (11.2583,19.5);
\draw [gray] (10.3923,20) -- (11.2583,20.5);
\draw [gray] (10.3923,20) -- (10.3923,21);
\draw [gray] (12.9904,23.5) -- (13.8564,23);
\draw [gray] (12.9904,23.5) -- (13.8564,24);
\draw [gray] (12.9904,23.5) -- (12.9904,24.5);
\draw [gray] (15.5885,17) -- (16.4545,16.5);
\draw [gray] (15.5885,17) -- (16.4545,17.5);
\draw [gray] (15.5885,17) -- (15.5885,18);
\draw [gray] (14.7224,19.5) -- (15.5885,19);
\draw [gray] (14.7224,19.5) -- (15.5885,20);
\draw [gray] (14.7224,19.5) -- (14.7224,20.5);
\draw [gray] (16.4545,17.5) -- (17.3205,17);
\draw [gray] (16.4545,17.5) -- (17.3205,18);
\draw [gray] (16.4545,17.5) -- (16.4545,18.5);
\draw [gray] (11.2583,19.5) -- (12.1244,19);
\draw [gray] (11.2583,19.5) -- (12.1244,20);
\draw [gray] (11.2583,19.5) -- (11.2583,20.5);
\draw [gray] (10.3923,23) -- (11.2583,22.5);
\draw [gray] (10.3923,23) -- (11.2583,23.5);
\draw [gray] (10.3923,23) -- (10.3923,24);
\draw [gray] (8.66025,26) -- (9.52628,25.5);
\draw [black,fill=] (12.9904,22.5) circle [radius = 0.2];
\draw [black,fill=] (9.52628,25.5) circle [radius = 0.2];
\draw [black,fill=] (7.79423,25.5) circle [radius = 0.2];
\draw [black,fill=] (13.8564,19) circle [radius = 0.2];
\draw [black,fill=] (17.3205,19) circle [radius = 0.2];
\draw [black,fill=] (8.66025,25) circle [radius = 0.2];
\draw [black,fill=] (12.9904,20.5) circle [radius = 0.2];
\draw [black,fill=] (13.8564,23) circle [radius = 0.2];
\draw [black,fill=] (11.2583,22.5) circle [radius = 0.2];
\draw [black,fill=] (11.2583,21.5) circle [radius = 0.2];
\draw [black,fill=] (10.3923,21) circle [radius = 0.2];
\draw [black,fill=] (13.8564,17) circle [radius = 0.2];
\draw [black,fill=] (14.7224,21.5) circle [radius = 0.2];
\draw [black,fill=] (12.9904,18.5) circle [radius = 0.2];
\draw [black,fill=] (10.3923,22) circle [radius = 0.2];
\draw [black,fill=] (10.3923,27) circle [radius = 0.2];
\draw [black,fill=] (14.7224,16.5) circle [radius = 0.2];
\draw [black,fill=] (9.52628,23.5) circle [radius = 0.2];
\draw [black,fill=] (18.1865,20.5) circle [radius = 0.2];
\draw [black,fill=] (14.7224,20.5) circle [radius = 0.2];
\draw [black,fill=] (9.52628,24.5) circle [radius = 0.2];
\draw [black,fill=] (12.9904,19.5) circle [radius = 0.2];
\draw [black,fill=] (17.3205,21) circle [radius = 0.2];
\draw [black,fill=] (15.5885,16) circle [radius = 0.2];
\draw [black,fill=] (12.1244,23) circle [radius = 0.2];
\draw [black,fill=] (12.1244,25) circle [radius = 0.2];
\draw [black,fill=] (18.1865,17.5) circle [radius = 0.2];
\draw [black,fill=] (12.9904,17.5) circle [radius = 0.2];
\draw [black,fill=] (11.2583,18.5) circle [radius = 0.2];
\draw [black,fill=] (11.2583,24.5) circle [radius = 0.2];
\draw [black,fill=] (17.3205,17) circle [radius = 0.2];
\draw [black,fill=] (10.3923,25) circle [radius = 0.2];
\draw [black,fill=] (15.5885,18) circle [radius = 0.2];
\draw [black,fill=] (12.1244,20) circle [radius = 0.2];
\draw [black,fill=] (18.1865,19.5) circle [radius = 0.2];
\draw [black,fill=] (12.1244,22) circle [radius = 0.2];
\draw [black,fill=] (12.9904,16.5) circle [radius = 0.2];
\draw [black,fill=] (10.3923,26) circle [radius = 0.2];
\draw [black,fill=] (14.7224,22.5) circle [radius = 0.2];
\draw [black,fill=] (15.5885,21) circle [radius = 0.2];
\draw [black,fill=] (16.4545,16.5) circle [radius = 0.2];
\draw [black,fill=] (16.4545,21.5) circle [radius = 0.2];
\draw [black,fill=] (11.2583,20.5) circle [radius = 0.2];
\draw [black,fill=] (12.9904,21.5) circle [radius = 0.2];
\draw [black,fill=] (17.3205,20) circle [radius = 0.2];
\draw [black,fill=] (14.7224,18.5) circle [radius = 0.2];
\draw [black,fill=] (16.4545,18.5) circle [radius = 0.2];
\draw [black,fill=] (14.7224,15.5) circle [radius = 0.2];
\draw [black,fill=] (10.3923,24) circle [radius = 0.2];
\draw [black,fill=] (13.8564,16) circle [radius = 0.2];
\draw [black,fill=] (16.4545,19.5) circle [radius = 0.2];
\draw [black,fill=] (13.8564,18) circle [radius = 0.2];
\draw [black,fill=] (13.8564,25) circle [radius = 0.2];
\draw [black,fill=] (18.1865,21.5) circle [radius = 0.2];
\draw [black,fill=] (15.5885,22) circle [radius = 0.2];
\draw [black,fill=] (12.1244,19) circle [radius = 0.2];
\draw [black,fill=] (13.8564,20) circle [radius = 0.2];
\draw [black,fill=] (18.1865,18.5) circle [radius = 0.2];
\draw [black,fill=] (19.0526,19) circle [radius = 0.2];
\draw [black,fill=] (10.3923,19) circle [radius = 0.2];
\draw [black,fill=] (14.7224,17.5) circle [radius = 0.2];
\draw [black,fill=] (13.8564,21) circle [radius = 0.2];
\draw [black,fill=] (19.0526,18) circle [radius = 0.2];
\draw [black,fill=] (9.52628,19.5) circle [radius = 0.2];
\draw [black,fill=] (12.1244,18) circle [radius = 0.2];
\draw [black,fill=] (16.4545,20.5) circle [radius = 0.2];
\draw [black,fill=] (16.4545,23.5) circle [radius = 0.2];
\draw [black,fill=] (17.3205,22) circle [radius = 0.2];
\draw [black,fill=] (15.5885,19) circle [radius = 0.2];
\draw [black,fill=] (11.2583,23.5) circle [radius = 0.2];
\draw [black,fill=] (17.3205,18) circle [radius = 0.2];
\draw [black,fill=] (12.9904,24.5) circle [radius = 0.2];
\draw [black,fill=] (14.7224,23.5) circle [radius = 0.2];
\draw [black,fill=] (15.5885,20) circle [radius = 0.2];
\draw [black,fill=] (13.8564,24) circle [radius = 0.2];
\draw [black,fill=] (16.4545,22.5) circle [radius = 0.2];
\draw [black,fill=] (11.2583,25.5) circle [radius = 0.2];
\draw [black,fill=] (14.7224,24.5) circle [radius = 0.2];
\draw [black,fill=] (13.8564,22) circle [radius = 0.2];
\draw [black,fill=] (15.5885,23) circle [radius = 0.2];
\draw [black,fill=] (12.1244,21) circle [radius = 0.2];
\draw [black,fill=] (12.1244,24) circle [radius = 0.2];
\draw [black,fill=] (10.3923,20) circle [radius = 0.2];
\draw [black,fill=] (12.9904,23.5) circle [radius = 0.2];
\draw [black,fill=] (15.5885,17) circle [radius = 0.2];
\draw [black,fill=] (19.0526,20) circle [radius = 0.2];
\draw [black,fill=] (14.7224,19.5) circle [radius = 0.2];
\draw [black,fill=] (16.4545,17.5) circle [radius = 0.2];
\draw [black,fill=] (11.2583,19.5) circle [radius = 0.2];
\draw [black,fill=] (10.3923,23) circle [radius = 0.2];
\draw [black,fill=] (8.66025,26) circle [radius = 0.2];
\draw [gray] (21.6506,10.5) -- (21.6506,9.5);
\draw [gray] (22.5167,12) -- (22.5167,11);
\draw [gray] (23.3827,13.5) -- (23.3827,12.5);
\draw [gray] (24.2487,15) -- (24.2487,14);
\draw [gray] (25.1147,16.5) -- (25.1147,15.5);
\draw [gray] (25.9808,18) -- (25.9808,17);
\draw [gray] (26.8468,19.5) -- (26.8468,18.5);
\draw [gray] (20.7846,9) -- (21.6506,9.5);
\draw [gray] (21.6506,10.5) -- (22.5167,11);
\draw [gray] (22.5167,12) -- (23.3827,12.5);
\draw [gray] (23.3827,13.5) -- (24.2487,14);
\draw [gray] (24.2487,15) -- (25.1147,15.5);
\draw [gray] (25.1147,16.5) -- (25.9808,17);
\draw [gray] (25.9808,18) -- (26.8468,18.5);
\draw [red,fill=] (20.7846,9) circle [radius = 0.2];
\draw [red,fill=] (21.6506,9.5) circle [radius = 0.2];
\draw [red,fill=] (21.6506,10.5) circle [radius = 0.2];
\draw [red,fill=] (22.5167,11) circle [radius = 0.2];
\draw [red,fill=] (22.5167,12) circle [radius = 0.2];
\draw [red,fill=] (23.3827,12.5) circle [radius = 0.2];
\draw [red,fill=] (23.3827,13.5) circle [radius = 0.2];
\draw [red,fill=] (24.2487,14) circle [radius = 0.2];
\draw [red,fill=] (24.2487,15) circle [radius = 0.2];
\draw [red,fill=] (25.1147,15.5) circle [radius = 0.2];
\draw [red,fill=] (25.1147,16.5) circle [radius = 0.2];
\draw [red,fill=] (25.9808,17) circle [radius = 0.2];
\draw [red,fill=] (25.9808,18) circle [radius = 0.2];
\draw [red,fill=] (26.8468,18.5) circle [radius = 0.2];
\draw [red,fill=] (26.8468,19.5) circle [radius = 0.2];
\end{tikzpicture}

%% file: hexagonlight_20000000_modified_cropped.txt
\begin{tikzpicture}[scale = 0.14, rotate = -60]
\draw [gray] (16.4545,20.5) -- (17.3205,20);
\draw [gray] (16.4545,20.5) -- (17.3205,21);
\draw [gray] (16.4545,20.5) -- (16.4545,21.5);
\draw [gray] (13.8564,25) -- (14.7224,24.5);
\draw [gray] (13.8564,25) -- (14.7224,25.5);
\draw [gray] (13.8564,25) -- (13.8564,26);
\draw [gray] (10.3923,23) -- (11.2583,22.5);
\draw [gray] (10.3923,23) -- (11.2583,23.5);
\draw [gray] (10.3923,23) -- (10.3923,24);
\draw [gray] (11.2583,23.5) -- (12.1244,23);
\draw [gray] (11.2583,23.5) -- (12.1244,24);
\draw [gray] (11.2583,23.5) -- (11.2583,24.5);
\draw [gray] (16.4545,24.5) -- (17.3205,24);
\draw [gray] (16.4545,24.5) -- (17.3205,25);
\draw [gray] (16.4545,24.5) -- (16.4545,25.5);
\draw [gray] (9.52628,26.5) -- (10.3923,26);
\draw [gray] (9.52628,26.5) -- (10.3923,27);
\draw [gray] (17.3205,26) -- (18.1865,25.5);
\draw [gray] (11.2583,27.5) -- (12.1244,27);
\draw [gray] (11.2583,27.5) -- (12.1244,28);
\draw [gray] (13.8564,24) -- (14.7224,23.5);
\draw [gray] (13.8564,24) -- (14.7224,24.5);
\draw [gray] (13.8564,24) -- (13.8564,25);
\draw [gray] (16.4545,23.5) -- (17.3205,23);
\draw [gray] (16.4545,23.5) -- (17.3205,24);
\draw [gray] (16.4545,23.5) -- (16.4545,24.5);
\draw [gray] (17.3205,20) -- (18.1865,20.5);
\draw [gray] (17.3205,20) -- (17.3205,21);
\draw [gray] (20.7846,21) -- (20.7846,22);
\draw [gray] (18.1865,20.5) -- (18.1865,21.5);
\draw [gray] (17.3205,19) -- (17.3205,20);
\draw [gray] (13.8564,20) -- (14.7224,20.5);
\draw [gray] (13.8564,20) -- (13.8564,21);
\draw [gray] (11.2583,21.5) -- (12.1244,21);
\draw [gray] (11.2583,21.5) -- (12.1244,22);
\draw [gray] (11.2583,21.5) -- (11.2583,22.5);
\draw [gray] (15.5885,23) -- (16.4545,22.5);
\draw [gray] (15.5885,23) -- (16.4545,23.5);
\draw [gray] (15.5885,23) -- (15.5885,24);
\draw [gray] (12.9904,24.5) -- (13.8564,24);
\draw [gray] (12.9904,24.5) -- (13.8564,25);
\draw [gray] (12.9904,24.5) -- (12.9904,25.5);
\draw [gray] (14.7224,20.5) -- (15.5885,20);
\draw [gray] (14.7224,20.5) -- (15.5885,21);
\draw [gray] (14.7224,20.5) -- (14.7224,21.5);
\draw [gray] (11.2583,22.5) -- (12.1244,22);
\draw [gray] (11.2583,22.5) -- (12.1244,23);
\draw [gray] (11.2583,22.5) -- (11.2583,23.5);
\draw [gray] (12.1244,24) -- (12.9904,23.5);
\draw [gray] (12.1244,24) -- (12.9904,24.5);
\draw [gray] (12.1244,24) -- (12.1244,25);
\draw [gray] (9.52628,25.5) -- (10.3923,25);
\draw [gray] (9.52628,25.5) -- (10.3923,26);
\draw [gray] (9.52628,25.5) -- (9.52628,26.5);
\draw [gray] (18.1865,23.5) -- (19.0526,23);
\draw [gray] (18.1865,23.5) -- (19.0526,24);
\draw [gray] (18.1865,23.5) -- (18.1865,24.5);
\draw [gray] (13.8564,23) -- (14.7224,22.5);
\draw [gray] (13.8564,23) -- (14.7224,23.5);
\draw [gray] (13.8564,23) -- (13.8564,24);
\draw [gray] (15.5885,20) -- (16.4545,19.5);
\draw [gray] (15.5885,20) -- (16.4545,20.5);
\draw [gray] (15.5885,20) -- (15.5885,21);
\draw [gray] (12.9904,20.5) -- (13.8564,20);
\draw [gray] (12.9904,20.5) -- (13.8564,21);
\draw [gray] (12.9904,20.5) -- (12.9904,21.5);
\draw [gray] (14.7224,21.5) -- (15.5885,21);
\draw [gray] (14.7224,21.5) -- (15.5885,22);
\draw [gray] (14.7224,21.5) -- (14.7224,22.5);
\draw [gray] (12.9904,22.5) -- (13.8564,22);
\draw [gray] (12.9904,22.5) -- (13.8564,23);
\draw [gray] (12.9904,22.5) -- (12.9904,23.5);
\draw [gray] (18.1865,22.5) -- (19.0526,22);
\draw [gray] (18.1865,22.5) -- (19.0526,23);
\draw [gray] (18.1865,22.5) -- (18.1865,23.5);
\draw [gray] (12.1244,26) -- (12.9904,25.5);
\draw [gray] (12.1244,26) -- (12.9904,26.5);
\draw [gray] (12.1244,26) -- (12.1244,27);
\draw [gray] (17.3205,23) -- (18.1865,22.5);
\draw [gray] (17.3205,23) -- (18.1865,23.5);
\draw [gray] (17.3205,23) -- (17.3205,24);
\draw [gray] (12.9904,25.5) -- (13.8564,25);
\draw [gray] (12.9904,25.5) -- (13.8564,26);
\draw [gray] (12.9904,25.5) -- (12.9904,26.5);
\draw [gray] (10.3923,26) -- (11.2583,25.5);
\draw [gray] (10.3923,26) -- (11.2583,26.5);
\draw [gray] (10.3923,26) -- (10.3923,27);
\draw [gray] (8.66025,26) -- (9.52628,25.5);
\draw [gray] (8.66025,26) -- (9.52628,26.5);
\draw [gray] (9.52628,28.5) -- (10.3923,28);
\draw [gray] (12.1244,22) -- (12.9904,21.5);
\draw [gray] (12.1244,22) -- (12.9904,22.5);
\draw [gray] (12.1244,22) -- (12.1244,23);
\draw [gray] (14.7224,22.5) -- (15.5885,22);
\draw [gray] (14.7224,22.5) -- (15.5885,23);
\draw [gray] (14.7224,22.5) -- (14.7224,23.5);
\draw [gray] (19.0526,22) -- (19.9186,21.5);
\draw [gray] (19.0526,22) -- (19.0526,23);
\draw [gray] (16.4545,19.5) -- (17.3205,19);
\draw [gray] (16.4545,19.5) -- (17.3205,20);
\draw [gray] (16.4545,19.5) -- (16.4545,20.5);
\draw [gray] (9.52628,23.5) -- (10.3923,23);
\draw [gray] (9.52628,23.5) -- (10.3923,24);
\draw [gray] (9.52628,23.5) -- (9.52628,24.5);
\draw [gray] (12.1244,21) -- (12.9904,20.5);
\draw [gray] (12.1244,21) -- (12.9904,21.5);
\draw [gray] (12.1244,21) -- (12.1244,22);
\draw [gray] (14.7224,25.5) -- (15.5885,25);
\draw [gray] (14.7224,25.5) -- (15.5885,26);
\draw [gray] (9.52628,24.5) -- (10.3923,24);
\draw [gray] (9.52628,24.5) -- (10.3923,25);
\draw [gray] (9.52628,24.5) -- (9.52628,25.5);
\draw [gray] (10.3923,25) -- (11.2583,24.5);
\draw [gray] (10.3923,25) -- (11.2583,25.5);
\draw [gray] (10.3923,25) -- (10.3923,26);
\draw [gray] (12.9904,21.5) -- (13.8564,21);
\draw [gray] (12.9904,21.5) -- (13.8564,22);
\draw [gray] (12.9904,21.5) -- (12.9904,22.5);
\draw [gray] (17.3205,24) -- (18.1865,23.5);
\draw [gray] (17.3205,24) -- (18.1865,24.5);
\draw [gray] (17.3205,24) -- (17.3205,25);
\draw [gray] (10.3923,28) -- (11.2583,27.5);
\draw [gray] (17.3205,22) -- (18.1865,21.5);
\draw [gray] (17.3205,22) -- (18.1865,22.5);
\draw [gray] (17.3205,22) -- (17.3205,23);
\draw [gray] (16.4545,22.5) -- (17.3205,22);
\draw [gray] (16.4545,22.5) -- (17.3205,23);
\draw [gray] (16.4545,22.5) -- (16.4545,23.5);
\draw [gray] (14.7224,23.5) -- (15.5885,23);
\draw [gray] (14.7224,23.5) -- (15.5885,24);
\draw [gray] (14.7224,23.5) -- (14.7224,24.5);
\draw [gray] (8.66025,25) -- (9.52628,24.5);
\draw [gray] (8.66025,25) -- (9.52628,25.5);
\draw [gray] (8.66025,25) -- (8.66025,26);
\draw [gray] (7.79423,25.5) -- (8.66025,25);
\draw [gray] (7.79423,25.5) -- (8.66025,26);
\draw [gray] (17.3205,21) -- (18.1865,20.5);
\draw [gray] (17.3205,21) -- (18.1865,21.5);
\draw [gray] (17.3205,21) -- (17.3205,22);
\draw [gray] (16.4545,25.5) -- (17.3205,25);
\draw [gray] (16.4545,25.5) -- (17.3205,26);
\draw [gray] (16.4545,25.5) -- (16.4545,26.5);
\draw [gray] (15.5885,25) -- (16.4545,24.5);
\draw [gray] (15.5885,25) -- (16.4545,25.5);
\draw [gray] (15.5885,25) -- (15.5885,26);
\draw [gray] (15.5885,26) -- (16.4545,25.5);
\draw [gray] (15.5885,26) -- (16.4545,26.5);
\draw [gray] (15.5885,26) -- (15.5885,27);
\draw [gray] (13.8564,22) -- (14.7224,21.5);
\draw [gray] (13.8564,22) -- (14.7224,22.5);
\draw [gray] (13.8564,22) -- (13.8564,23);
\draw [gray] (10.3923,24) -- (11.2583,23.5);
\draw [gray] (10.3923,24) -- (11.2583,24.5);
\draw [gray] (10.3923,24) -- (10.3923,25);
\draw [gray] (16.4545,26.5) -- (17.3205,26);
\draw [gray] (19.9186,21.5) -- (20.7846,21);
\draw [gray] (19.9186,21.5) -- (20.7846,22);
\draw [gray] (11.2583,24.5) -- (12.1244,24);
\draw [gray] (11.2583,24.5) -- (12.1244,25);
\draw [gray] (11.2583,24.5) -- (11.2583,25.5);
\draw [gray] (13.8564,26) -- (14.7224,25.5);
\draw [gray] (13.8564,26) -- (13.8564,27);
\draw [gray] (15.5885,21) -- (16.4545,20.5);
\draw [gray] (15.5885,21) -- (16.4545,21.5);
\draw [gray] (15.5885,21) -- (15.5885,22);
\draw [gray] (12.1244,23) -- (12.9904,22.5);
\draw [gray] (12.1244,23) -- (12.9904,23.5);
\draw [gray] (12.1244,23) -- (12.1244,24);
\draw [gray] (11.2583,26.5) -- (12.1244,26);
\draw [gray] (11.2583,26.5) -- (12.1244,27);
\draw [gray] (11.2583,26.5) -- (11.2583,27.5);
\draw [gray] (17.3205,25) -- (18.1865,24.5);
\draw [gray] (17.3205,25) -- (18.1865,25.5);
\draw [gray] (17.3205,25) -- (17.3205,26);
\draw [gray] (20.7846,22) -- (21.6506,22.5);
\draw [gray] (11.2583,25.5) -- (12.1244,25);
\draw [gray] (11.2583,25.5) -- (12.1244,26);
\draw [gray] (11.2583,25.5) -- (11.2583,26.5);
\draw [gray] (15.5885,24) -- (16.4545,23.5);
\draw [gray] (15.5885,24) -- (16.4545,24.5);
\draw [gray] (15.5885,24) -- (15.5885,25);
\draw [gray] (12.1244,28) -- (12.9904,27.5);
\draw [gray] (13.8564,27) -- (13.8564,28);
\draw [gray] (12.1244,27) -- (12.9904,26.5);
\draw [gray] (12.1244,27) -- (12.9904,27.5);
\draw [gray] (12.1244,27) -- (12.1244,28);
\draw [gray] (15.5885,22) -- (16.4545,21.5);
\draw [gray] (15.5885,22) -- (16.4545,22.5);
\draw [gray] (15.5885,22) -- (15.5885,23);
\draw [gray] (18.1865,21.5) -- (19.0526,22);
\draw [gray] (18.1865,21.5) -- (18.1865,22.5);
\draw [gray] (13.8564,21) -- (14.7224,20.5);
\draw [gray] (13.8564,21) -- (14.7224,21.5);
\draw [gray] (13.8564,21) -- (13.8564,22);
\draw [gray] (12.1244,25) -- (12.9904,24.5);
\draw [gray] (12.1244,25) -- (12.9904,25.5);
\draw [gray] (12.1244,25) -- (12.1244,26);
\draw [gray] (12.9904,27.5) -- (13.8564,27);
\draw [gray] (12.9904,27.5) -- (13.8564,28);
\draw [gray] (14.7224,24.5) -- (15.5885,24);
\draw [gray] (14.7224,24.5) -- (15.5885,25);
\draw [gray] (14.7224,24.5) -- (14.7224,25.5);
\draw [gray] (15.5885,27) -- (16.4545,26.5);
\draw [gray] (15.5885,27) -- (15.5885,28);
\draw [gray] (16.4545,21.5) -- (17.3205,21);
\draw [gray] (16.4545,21.5) -- (17.3205,22);
\draw [gray] (16.4545,21.5) -- (16.4545,22.5);
\draw [gray] (12.9904,26.5) -- (13.8564,26);
\draw [gray] (12.9904,26.5) -- (13.8564,27);
\draw [gray] (12.9904,26.5) -- (12.9904,27.5);
\draw [gray] (10.3923,27) -- (11.2583,26.5);
\draw [gray] (10.3923,27) -- (11.2583,27.5);
\draw [gray] (10.3923,27) -- (10.3923,28);
\draw [gray] (18.1865,24.5) -- (19.0526,24);
\draw [gray] (18.1865,24.5) -- (18.1865,25.5);
\draw [gray] (19.0526,23) -- (19.0526,24);
\draw [gray] (12.9904,23.5) -- (13.8564,23);
\draw [gray] (12.9904,23.5) -- (13.8564,24);
\draw [gray] (12.9904,23.5) -- (12.9904,24.5);
\draw [gray] (15.5885,19) -- (16.4545,19.5);
\draw [gray] (15.5885,19) -- (15.5885,20);
\draw [black,fill=] (21.6506,22.5) circle [radius = 0.2];
\draw [black,fill=] (16.4545,20.5) circle [radius = 0.2];
\draw [black,fill=] (13.8564,25) circle [radius = 0.2];
\draw [black,fill=] (10.3923,23) circle [radius = 0.2];
\draw [black,fill=] (11.2583,23.5) circle [radius = 0.2];
\draw [black,fill=] (16.4545,24.5) circle [radius = 0.2];
\draw [black,fill=] (9.52628,26.5) circle [radius = 0.2];
\draw [black,fill=] (17.3205,26) circle [radius = 0.2];
\draw [black,fill=] (11.2583,27.5) circle [radius = 0.2];
\draw [black,fill=] (13.8564,24) circle [radius = 0.2];
\draw [black,fill=] (16.4545,23.5) circle [radius = 0.2];
\draw [black,fill=] (17.3205,20) circle [radius = 0.2];
\draw [black,fill=] (20.7846,21) circle [radius = 0.2];
\draw [black,fill=] (18.1865,20.5) circle [radius = 0.2];
\draw [black,fill=] (17.3205,19) circle [radius = 0.2];
\draw [black,fill=] (13.8564,20) circle [radius = 0.2];
\draw [black,fill=] (11.2583,21.5) circle [radius = 0.2];
\draw [black,fill=] (15.5885,23) circle [radius = 0.2];
\draw [black,fill=] (12.9904,24.5) circle [radius = 0.2];
\draw [black,fill=] (14.7224,20.5) circle [radius = 0.2];
\draw [black,fill=] (11.2583,22.5) circle [radius = 0.2];
\draw [black,fill=] (12.1244,24) circle [radius = 0.2];
\draw [black,fill=] (9.52628,25.5) circle [radius = 0.2];
\draw [black,fill=] (18.1865,23.5) circle [radius = 0.2];
\draw [black,fill=] (13.8564,23) circle [radius = 0.2];
\draw [black,fill=] (15.5885,20) circle [radius = 0.2];
\draw [black,fill=] (12.9904,20.5) circle [radius = 0.2];
\draw [black,fill=] (18.1865,25.5) circle [radius = 0.2];
\draw [black,fill=] (14.7224,21.5) circle [radius = 0.2];
\draw [black,fill=] (12.9904,22.5) circle [radius = 0.2];
\draw [black,fill=] (18.1865,22.5) circle [radius = 0.2];
\draw [black,fill=] (12.1244,26) circle [radius = 0.2];
\draw [black,fill=] (17.3205,23) circle [radius = 0.2];
\draw [black,fill=] (12.9904,25.5) circle [radius = 0.2];
\draw [black,fill=] (10.3923,26) circle [radius = 0.2];
\draw [black,fill=] (8.66025,26) circle [radius = 0.2];
\draw [black,fill=] (9.52628,28.5) circle [radius = 0.2];
\draw [black,fill=] (12.1244,22) circle [radius = 0.2];
\draw [black,fill=] (14.7224,22.5) circle [radius = 0.2];
\draw [black,fill=] (19.0526,22) circle [radius = 0.2];
\draw [black,fill=] (16.4545,19.5) circle [radius = 0.2];
\draw [black,fill=] (15.5885,28) circle [radius = 0.2];
\draw [black,fill=] (9.52628,23.5) circle [radius = 0.2];
\draw [black,fill=] (12.1244,21) circle [radius = 0.2];
\draw [black,fill=] (14.7224,25.5) circle [radius = 0.2];
\draw [black,fill=] (9.52628,24.5) circle [radius = 0.2];
\draw [black,fill=] (10.3923,25) circle [radius = 0.2];
\draw [black,fill=] (12.9904,21.5) circle [radius = 0.2];
\draw [black,fill=] (17.3205,24) circle [radius = 0.2];
\draw [black,fill=] (10.3923,28) circle [radius = 0.2];
\draw [black,fill=] (17.3205,22) circle [radius = 0.2];
\draw [black,fill=] (16.4545,22.5) circle [radius = 0.2];
\draw [black,fill=] (14.7224,23.5) circle [radius = 0.2];
\draw [black,fill=] (8.66025,25) circle [radius = 0.2];
\draw [black,fill=] (7.79423,25.5) circle [radius = 0.2];
\draw [black,fill=] (17.3205,21) circle [radius = 0.2];
\draw [black,fill=] (16.4545,25.5) circle [radius = 0.2];
\draw [black,fill=] (15.5885,25) circle [radius = 0.2];
\draw [black,fill=] (15.5885,26) circle [radius = 0.2];
\draw [black,fill=] (13.8564,22) circle [radius = 0.2];
\draw [black,fill=] (10.3923,24) circle [radius = 0.2];
\draw [black,fill=] (16.4545,26.5) circle [radius = 0.2];
\draw [black,fill=] (19.9186,21.5) circle [radius = 0.2];
\draw [black,fill=] (11.2583,24.5) circle [radius = 0.2];
\draw [black,fill=] (13.8564,26) circle [radius = 0.2];
\draw [black,fill=] (15.5885,21) circle [radius = 0.2];
\draw [black,fill=] (12.1244,23) circle [radius = 0.2];
\draw [black,fill=] (11.2583,26.5) circle [radius = 0.2];
\draw [black,fill=] (17.3205,25) circle [radius = 0.2];
\draw [black,fill=] (20.7846,22) circle [radius = 0.2];
\draw [black,fill=] (11.2583,25.5) circle [radius = 0.2];
\draw [black,fill=] (15.5885,24) circle [radius = 0.2];
\draw [black,fill=] (12.1244,28) circle [radius = 0.2];
\draw [black,fill=] (13.8564,27) circle [radius = 0.2];
\draw [black,fill=] (12.1244,27) circle [radius = 0.2];
\draw [black,fill=] (15.5885,22) circle [radius = 0.2];
\draw [black,fill=] (19.0526,24) circle [radius = 0.2];
\draw [black,fill=] (18.1865,21.5) circle [radius = 0.2];
\draw [black,fill=] (13.8564,21) circle [radius = 0.2];
\draw [black,fill=] (12.1244,25) circle [radius = 0.2];
\draw [black,fill=] (13.8564,28) circle [radius = 0.2];
\draw [black,fill=] (12.9904,27.5) circle [radius = 0.2];
\draw [black,fill=] (14.7224,24.5) circle [radius = 0.2];
\draw [black,fill=] (15.5885,27) circle [radius = 0.2];
\draw [black,fill=] (16.4545,21.5) circle [radius = 0.2];
\draw [black,fill=] (12.9904,26.5) circle [radius = 0.2];
\draw [black,fill=] (10.3923,27) circle [radius = 0.2];
\draw [black,fill=] (18.1865,24.5) circle [radius = 0.2];
\draw [black,fill=] (19.0526,23) circle [radius = 0.2];
\draw [black,fill=] (12.9904,23.5) circle [radius = 0.2];
\draw [black,fill=] (15.5885,19) circle [radius = 0.2];
\draw [gray] (23.3827,13.5) -- (23.3827,12.5);
\draw [gray] (24.2487,15) -- (24.2487,14);
\draw [gray] (25.1147,16.5) -- (25.1147,15.5);
\draw [gray] (25.9808,18) -- (25.9808,17);
\draw [gray] (26.8468,19.5) -- (26.8468,18.5);
\draw [gray] (27.7128,21) -- (27.7128,20);
\draw [gray] (28.5788,22.5) -- (28.5788,21.5);
\draw [gray] (22.5167,12) -- (23.3827,12.5);
\draw [gray] (23.3827,13.5) -- (24.2487,14);
\draw [gray] (24.2487,15) -- (25.1147,15.5);
\draw [gray] (25.1147,16.5) -- (25.9808,17);
\draw [gray] (25.9808,18) -- (26.8468,18.5);
\draw [gray] (26.8468,19.5) -- (27.7128,20);
\draw [gray] (27.7128,21) -- (28.5788,21.5);
\draw [red,fill=] (22.5167,12) circle [radius = 0.2];
\draw [red,fill=] (23.3827,12.5) circle [radius = 0.2];
\draw [red,fill=] (23.3827,13.5) circle [radius = 0.2];
\draw [red,fill=] (24.2487,14) circle [radius = 0.2];
\draw [red,fill=] (24.2487,15) circle [radius = 0.2];
\draw [red,fill=] (25.1147,15.5) circle [radius = 0.2];
\draw [red,fill=] (25.1147,16.5) circle [radius = 0.2];
\draw [red,fill=] (25.9808,17) circle [radius = 0.2];
\draw [red,fill=] (25.9808,18) circle [radius = 0.2];
\draw [red,fill=] (26.8468,18.5) circle [radius = 0.2];
\draw [red,fill=] (26.8468,19.5) circle [radius = 0.2];
\draw [red,fill=] (27.7128,20) circle [radius = 0.2];
\draw [red,fill=] (27.7128,21) circle [radius = 0.2];
\draw [red,fill=] (28.5788,21.5) circle [radius = 0.2];
\draw [red,fill=] (28.5788,22.5) circle [radius = 0.2];
\end{tikzpicture}

%% file: hexagonlight_30000000_modified_cropped.txt
\begin{tikzpicture}[scale = 0.14, rotate = -60]
\draw [gray] (12.9904,25.5) -- (13.8564,25);
\draw [gray] (12.9904,25.5) -- (13.8564,26);
\draw [gray] (12.9904,25.5) -- (12.9904,26.5);
\draw [gray] (11.2583,29.5) -- (12.1244,29);
\draw [gray] (11.2583,29.5) -- (12.1244,30);
\draw [gray] (11.2583,29.5) -- (11.2583,30.5);
\draw [gray] (15.5885,26) -- (16.4545,26.5);
\draw [gray] (15.5885,26) -- (15.5885,27);
\draw [gray] (14.7224,25.5) -- (15.5885,26);
\draw [gray] (14.7224,25.5) -- (14.7224,26.5);
\draw [gray] (8.66025,25) -- (9.52628,24.5);
\draw [gray] (8.66025,25) -- (9.52628,25.5);
\draw [gray] (8.66025,25) -- (8.66025,26);
\draw [gray] (12.1244,26) -- (12.9904,25.5);
\draw [gray] (12.1244,26) -- (12.9904,26.5);
\draw [gray] (12.1244,26) -- (12.1244,27);
\draw [gray] (9.52628,28.5) -- (10.3923,28);
\draw [gray] (9.52628,28.5) -- (10.3923,29);
\draw [gray] (9.52628,28.5) -- (9.52628,29.5);
\draw [gray] (6.06218,24.5) -- (6.9282,25);
\draw [gray] (6.06218,24.5) -- (6.06218,25.5);
\draw [gray] (12.9904,27.5) -- (13.8564,27);
\draw [gray] (12.9904,27.5) -- (13.8564,28);
\draw [gray] (12.9904,27.5) -- (12.9904,28.5);
\draw [gray] (11.2583,25.5) -- (12.1244,25);
\draw [gray] (11.2583,25.5) -- (12.1244,26);
\draw [gray] (11.2583,25.5) -- (11.2583,26.5);
\draw [gray] (5.19615,27) -- (6.06218,26.5);
\draw [gray] (5.19615,27) -- (6.06218,27.5);
\draw [gray] (12.9904,30.5) -- (13.8564,30);
\draw [gray] (12.9904,30.5) -- (13.8564,31);
\draw [gray] (8.66025,29) -- (9.52628,28.5);
\draw [gray] (8.66025,29) -- (9.52628,29.5);
\draw [gray] (8.66025,29) -- (8.66025,30);
\draw [gray] (9.52628,29.5) -- (10.3923,29);
\draw [gray] (9.52628,29.5) -- (10.3923,30);
\draw [gray] (9.52628,29.5) -- (9.52628,30.5);
\draw [gray] (7.79423,27.5) -- (8.66025,27);
\draw [gray] (7.79423,27.5) -- (8.66025,28);
\draw [gray] (7.79423,27.5) -- (7.79423,28.5);
\draw [gray] (11.2583,26.5) -- (12.1244,26);
\draw [gray] (11.2583,26.5) -- (12.1244,27);
\draw [gray] (11.2583,26.5) -- (11.2583,27.5);
\draw [gray] (5.19615,26) -- (6.06218,25.5);
\draw [gray] (5.19615,26) -- (6.06218,26.5);
\draw [gray] (5.19615,26) -- (5.19615,27);
\draw [gray] (13.8564,24) -- (14.7224,24.5);
\draw [gray] (13.8564,24) -- (13.8564,25);
\draw [gray] (6.9282,27) -- (7.79423,26.5);
\draw [gray] (6.9282,27) -- (7.79423,27.5);
\draw [gray] (6.9282,27) -- (6.9282,28);
\draw [gray] (10.3923,28) -- (11.2583,27.5);
\draw [gray] (10.3923,28) -- (11.2583,28.5);
\draw [gray] (10.3923,28) -- (10.3923,29);
\draw [gray] (15.5885,29) -- (16.4545,29.5);
\draw [gray] (15.5885,29) -- (15.5885,30);
\draw [gray] (10.3923,29) -- (11.2583,28.5);
\draw [gray] (10.3923,29) -- (11.2583,29.5);
\draw [gray] (10.3923,29) -- (10.3923,30);
\draw [gray] (12.1244,27) -- (12.9904,26.5);
\draw [gray] (12.1244,27) -- (12.9904,27.5);
\draw [gray] (12.1244,27) -- (12.1244,28);
\draw [gray] (9.52628,25.5) -- (10.3923,25);
\draw [gray] (9.52628,25.5) -- (10.3923,26);
\draw [gray] (9.52628,25.5) -- (9.52628,26.5);
\draw [gray] (9.52628,27.5) -- (10.3923,27);
\draw [gray] (9.52628,27.5) -- (10.3923,28);
\draw [gray] (9.52628,27.5) -- (9.52628,28.5);
\draw [gray] (12.1244,28) -- (12.9904,27.5);
\draw [gray] (12.1244,28) -- (12.9904,28.5);
\draw [gray] (12.1244,28) -- (12.1244,29);
\draw [gray] (13.8564,29) -- (14.7224,28.5);
\draw [gray] (13.8564,29) -- (14.7224,29.5);
\draw [gray] (13.8564,29) -- (13.8564,30);
\draw [gray] (13.8564,31) -- (14.7224,30.5);
\draw [gray] (14.7224,28.5) -- (15.5885,29);
\draw [gray] (14.7224,28.5) -- (14.7224,29.5);
\draw [gray] (6.9282,26) -- (7.79423,25.5);
\draw [gray] (6.9282,26) -- (7.79423,26.5);
\draw [gray] (6.9282,26) -- (6.9282,27);
\draw [gray] (14.7224,24.5) -- (14.7224,25.5);
\draw [gray] (7.79423,28.5) -- (8.66025,28);
\draw [gray] (7.79423,28.5) -- (8.66025,29);
\draw [gray] (7.79423,28.5) -- (7.79423,29.5);
\draw [gray] (12.1244,30) -- (12.9904,29.5);
\draw [gray] (12.1244,30) -- (12.9904,30.5);
\draw [gray] (12.1244,30) -- (12.1244,31);
\draw [gray] (8.66025,24) -- (9.52628,24.5);
\draw [gray] (8.66025,24) -- (8.66025,25);
\draw [gray] (14.7224,29.5) -- (15.5885,29);
\draw [gray] (14.7224,29.5) -- (15.5885,30);
\draw [gray] (14.7224,29.5) -- (14.7224,30.5);
\draw [gray] (13.8564,26) -- (14.7224,25.5);
\draw [gray] (13.8564,26) -- (14.7224,26.5);
\draw [gray] (13.8564,26) -- (13.8564,27);
\draw [gray] (12.9904,24.5) -- (13.8564,24);
\draw [gray] (12.9904,24.5) -- (13.8564,25);
\draw [gray] (12.9904,24.5) -- (12.9904,25.5);
\draw [gray] (11.2583,28.5) -- (12.1244,28);
\draw [gray] (11.2583,28.5) -- (12.1244,29);
\draw [gray] (11.2583,28.5) -- (11.2583,29.5);
\draw [gray] (15.5885,27) -- (16.4545,26.5);
\draw [gray] (11.2583,32.5) -- (12.1244,32);
\draw [gray] (6.06218,27.5) -- (6.9282,27);
\draw [gray] (6.06218,27.5) -- (6.9282,28);
\draw [gray] (11.2583,27.5) -- (12.1244,27);
\draw [gray] (11.2583,27.5) -- (12.1244,28);
\draw [gray] (11.2583,27.5) -- (11.2583,28.5);
\draw [gray] (11.2583,24.5) -- (12.1244,24);
\draw [gray] (11.2583,24.5) -- (12.1244,25);
\draw [gray] (11.2583,24.5) -- (11.2583,25.5);
\draw [gray] (6.06218,26.5) -- (6.9282,26);
\draw [gray] (6.06218,26.5) -- (6.9282,27);
\draw [gray] (6.06218,26.5) -- (6.06218,27.5);
\draw [gray] (12.1244,29) -- (12.9904,28.5);
\draw [gray] (12.1244,29) -- (12.9904,29.5);
\draw [gray] (12.1244,29) -- (12.1244,30);
\draw [gray] (7.79423,24.5) -- (8.66025,24);
\draw [gray] (7.79423,24.5) -- (8.66025,25);
\draw [gray] (7.79423,24.5) -- (7.79423,25.5);
\draw [gray] (16.4545,29.5) -- (16.4545,30.5);
\draw [gray] (12.9904,28.5) -- (13.8564,28);
\draw [gray] (12.9904,28.5) -- (13.8564,29);
\draw [gray] (12.9904,28.5) -- (12.9904,29.5);
\draw [gray] (7.79423,26.5) -- (8.66025,26);
\draw [gray] (7.79423,26.5) -- (8.66025,27);
\draw [gray] (7.79423,26.5) -- (7.79423,27.5);
\draw [gray] (6.9282,25) -- (7.79423,24.5);
\draw [gray] (6.9282,25) -- (7.79423,25.5);
\draw [gray] (6.9282,25) -- (6.9282,26);
\draw [gray] (10.3923,24) -- (11.2583,24.5);
\draw [gray] (10.3923,24) -- (10.3923,25);
\draw [gray] (6.9282,29) -- (7.79423,28.5);
\draw [gray] (6.9282,29) -- (7.79423,29.5);
\draw [gray] (7.79423,29.5) -- (8.66025,29);
\draw [gray] (7.79423,29.5) -- (8.66025,30);
\draw [gray] (7.79423,29.5) -- (7.79423,30.5);
\draw [gray] (13.8564,28) -- (14.7224,28.5);
\draw [gray] (13.8564,28) -- (13.8564,29);
\draw [gray] (7.79423,25.5) -- (8.66025,25);
\draw [gray] (7.79423,25.5) -- (8.66025,26);
\draw [gray] (7.79423,25.5) -- (7.79423,26.5);
\draw [gray] (6.06218,25.5) -- (6.9282,25);
\draw [gray] (6.06218,25.5) -- (6.9282,26);
\draw [gray] (6.06218,25.5) -- (6.06218,26.5);
\draw [gray] (11.2583,30.5) -- (12.1244,30);
\draw [gray] (11.2583,30.5) -- (12.1244,31);
\draw [gray] (11.2583,30.5) -- (11.2583,31.5);
\draw [gray] (15.5885,30) -- (16.4545,29.5);
\draw [gray] (15.5885,30) -- (16.4545,30.5);
\draw [gray] (10.3923,26) -- (11.2583,25.5);
\draw [gray] (10.3923,26) -- (11.2583,26.5);
\draw [gray] (10.3923,26) -- (10.3923,27);
\draw [gray] (9.52628,26.5) -- (10.3923,26);
\draw [gray] (9.52628,26.5) -- (10.3923,27);
\draw [gray] (9.52628,26.5) -- (9.52628,27.5);
\draw [gray] (10.3923,30) -- (11.2583,29.5);
\draw [gray] (10.3923,30) -- (11.2583,30.5);
\draw [gray] (10.3923,30) -- (10.3923,31);
\draw [gray] (13.8564,27) -- (14.7224,26.5);
\draw [gray] (13.8564,27) -- (13.8564,28);
\draw [gray] (12.1244,24) -- (12.9904,23.5);
\draw [gray] (12.1244,24) -- (12.9904,24.5);
\draw [gray] (12.1244,24) -- (12.1244,25);
\draw [gray] (8.66025,30) -- (9.52628,29.5);
\draw [gray] (8.66025,30) -- (9.52628,30.5);
\draw [gray] (8.66025,27) -- (9.52628,26.5);
\draw [gray] (8.66025,27) -- (9.52628,27.5);
\draw [gray] (8.66025,27) -- (8.66025,28);
\draw [gray] (16.4545,31.5) -- (17.3205,32);
\draw [gray] (8.66025,28) -- (9.52628,27.5);
\draw [gray] (8.66025,28) -- (9.52628,28.5);
\draw [gray] (8.66025,28) -- (8.66025,29);
\draw [gray] (9.52628,24.5) -- (10.3923,24);
\draw [gray] (9.52628,24.5) -- (10.3923,25);
\draw [gray] (9.52628,24.5) -- (9.52628,25.5);
\draw [gray] (7.79423,30.5) -- (8.66025,30);
\draw [gray] (6.9282,28) -- (7.79423,27.5);
\draw [gray] (6.9282,28) -- (7.79423,28.5);
\draw [gray] (6.9282,28) -- (6.9282,29);
\draw [gray] (10.3923,31) -- (11.2583,30.5);
\draw [gray] (10.3923,31) -- (11.2583,31.5);
\draw [gray] (10.3923,31) -- (10.3923,32);
\draw [gray] (13.8564,25) -- (14.7224,24.5);
\draw [gray] (13.8564,25) -- (14.7224,25.5);
\draw [gray] (13.8564,25) -- (13.8564,26);
\draw [gray] (8.66025,26) -- (9.52628,25.5);
\draw [gray] (8.66025,26) -- (9.52628,26.5);
\draw [gray] (8.66025,26) -- (8.66025,27);
\draw [gray] (10.3923,25) -- (11.2583,24.5);
\draw [gray] (10.3923,25) -- (11.2583,25.5);
\draw [gray] (10.3923,25) -- (10.3923,26);
\draw [gray] (12.1244,25) -- (12.9904,24.5);
\draw [gray] (12.1244,25) -- (12.9904,25.5);
\draw [gray] (12.1244,25) -- (12.1244,26);
\draw [gray] (12.1244,31) -- (12.9904,30.5);
\draw [gray] (12.1244,31) -- (12.1244,32);
\draw [gray] (10.3923,32) -- (11.2583,31.5);
\draw [gray] (10.3923,32) -- (11.2583,32.5);
\draw [gray] (11.2583,31.5) -- (12.1244,31);
\draw [gray] (11.2583,31.5) -- (12.1244,32);
\draw [gray] (11.2583,31.5) -- (11.2583,32.5);
\draw [gray] (12.9904,23.5) -- (13.8564,24);
\draw [gray] (12.9904,23.5) -- (12.9904,24.5);
\draw [gray] (14.7224,26.5) -- (15.5885,26);
\draw [gray] (14.7224,26.5) -- (15.5885,27);
\draw [gray] (16.4545,30.5) -- (16.4545,31.5);
\draw [gray] (12.9904,26.5) -- (13.8564,26);
\draw [gray] (12.9904,26.5) -- (13.8564,27);
\draw [gray] (12.9904,26.5) -- (12.9904,27.5);
\draw [gray] (10.3923,27) -- (11.2583,26.5);
\draw [gray] (10.3923,27) -- (11.2583,27.5);
\draw [gray] (10.3923,27) -- (10.3923,28);
\draw [gray] (12.1244,23) -- (12.9904,23.5);
\draw [gray] (12.1244,23) -- (12.1244,24);
\draw [gray] (9.52628,30.5) -- (10.3923,30);
\draw [gray] (9.52628,30.5) -- (10.3923,31);
\draw [gray] (13.8564,30) -- (14.7224,29.5);
\draw [gray] (13.8564,30) -- (14.7224,30.5);
\draw [gray] (13.8564,30) -- (13.8564,31);
\draw [gray] (14.7224,30.5) -- (15.5885,30);
\draw [gray] (12.9904,29.5) -- (13.8564,29);
\draw [gray] (12.9904,29.5) -- (13.8564,30);
\draw [gray] (12.9904,29.5) -- (12.9904,30.5);
\draw [black,fill=] (12.9904,25.5) circle [radius = 0.2];
\draw [black,fill=] (11.2583,29.5) circle [radius = 0.2];
\draw [black,fill=] (15.5885,26) circle [radius = 0.2];
\draw [black,fill=] (14.7224,25.5) circle [radius = 0.2];
\draw [black,fill=] (8.66025,25) circle [radius = 0.2];
\draw [black,fill=] (12.1244,26) circle [radius = 0.2];
\draw [black,fill=] (9.52628,28.5) circle [radius = 0.2];
\draw [black,fill=] (6.06218,24.5) circle [radius = 0.2];
\draw [black,fill=] (12.9904,27.5) circle [radius = 0.2];
\draw [black,fill=] (11.2583,25.5) circle [radius = 0.2];
\draw [black,fill=] (5.19615,27) circle [radius = 0.2];
\draw [black,fill=] (12.9904,30.5) circle [radius = 0.2];
\draw [black,fill=] (8.66025,29) circle [radius = 0.2];
\draw [black,fill=] (9.52628,29.5) circle [radius = 0.2];
\draw [black,fill=] (7.79423,27.5) circle [radius = 0.2];
\draw [black,fill=] (11.2583,26.5) circle [radius = 0.2];
\draw [black,fill=] (5.19615,26) circle [radius = 0.2];
\draw [black,fill=] (13.8564,24) circle [radius = 0.2];
\draw [black,fill=] (6.9282,27) circle [radius = 0.2];
\draw [black,fill=] (10.3923,28) circle [radius = 0.2];
\draw [black,fill=] (15.5885,29) circle [radius = 0.2];
\draw [black,fill=] (10.3923,29) circle [radius = 0.2];
\draw [black,fill=] (12.1244,27) circle [radius = 0.2];
\draw [black,fill=] (9.52628,25.5) circle [radius = 0.2];
\draw [black,fill=] (9.52628,27.5) circle [radius = 0.2];
\draw [black,fill=] (12.1244,28) circle [radius = 0.2];
\draw [black,fill=] (13.8564,29) circle [radius = 0.2];
\draw [black,fill=] (13.8564,31) circle [radius = 0.2];
\draw [black,fill=] (14.7224,28.5) circle [radius = 0.2];
\draw [black,fill=] (6.9282,26) circle [radius = 0.2];
\draw [black,fill=] (14.7224,24.5) circle [radius = 0.2];
\draw [black,fill=] (7.79423,28.5) circle [radius = 0.2];
\draw [black,fill=] (12.1244,30) circle [radius = 0.2];
\draw [black,fill=] (8.66025,24) circle [radius = 0.2];
\draw [black,fill=] (14.7224,29.5) circle [radius = 0.2];
\draw [black,fill=] (13.8564,26) circle [radius = 0.2];
\draw [black,fill=] (12.9904,24.5) circle [radius = 0.2];
\draw [black,fill=] (11.2583,28.5) circle [radius = 0.2];
\draw [black,fill=] (15.5885,27) circle [radius = 0.2];
\draw [black,fill=] (11.2583,32.5) circle [radius = 0.2];
\draw [black,fill=] (6.06218,27.5) circle [radius = 0.2];
\draw [black,fill=] (11.2583,27.5) circle [radius = 0.2];
\draw [black,fill=] (11.2583,24.5) circle [radius = 0.2];
\draw [black,fill=] (6.06218,26.5) circle [radius = 0.2];
\draw [black,fill=] (12.1244,29) circle [radius = 0.2];
\draw [black,fill=] (7.79423,24.5) circle [radius = 0.2];
\draw [black,fill=] (16.4545,29.5) circle [radius = 0.2];
\draw [black,fill=] (12.9904,28.5) circle [radius = 0.2];
\draw [black,fill=] (7.79423,26.5) circle [radius = 0.2];
\draw [black,fill=] (6.9282,25) circle [radius = 0.2];
\draw [black,fill=] (10.3923,24) circle [radius = 0.2];
\draw [black,fill=] (6.9282,29) circle [radius = 0.2];
\draw [black,fill=] (7.79423,29.5) circle [radius = 0.2];
\draw [black,fill=] (13.8564,28) circle [radius = 0.2];
\draw [black,fill=] (7.79423,25.5) circle [radius = 0.2];
\draw [black,fill=] (6.06218,25.5) circle [radius = 0.2];
\draw [black,fill=] (11.2583,30.5) circle [radius = 0.2];
\draw [black,fill=] (15.5885,30) circle [radius = 0.2];
\draw [black,fill=] (10.3923,26) circle [radius = 0.2];
\draw [black,fill=] (9.52628,26.5) circle [radius = 0.2];
\draw [black,fill=] (10.3923,30) circle [radius = 0.2];
\draw [black,fill=] (13.8564,27) circle [radius = 0.2];
\draw [black,fill=] (12.1244,24) circle [radius = 0.2];
\draw [black,fill=] (8.66025,30) circle [radius = 0.2];
\draw [black,fill=] (8.66025,27) circle [radius = 0.2];
\draw [black,fill=] (16.4545,31.5) circle [radius = 0.2];
\draw [black,fill=] (8.66025,28) circle [radius = 0.2];
\draw [black,fill=] (9.52628,24.5) circle [radius = 0.2];
\draw [black,fill=] (7.79423,30.5) circle [radius = 0.2];
\draw [black,fill=] (6.9282,28) circle [radius = 0.2];
\draw [black,fill=] (10.3923,31) circle [radius = 0.2];
\draw [black,fill=] (13.8564,25) circle [radius = 0.2];
\draw [black,fill=] (8.66025,26) circle [radius = 0.2];
\draw [black,fill=] (10.3923,25) circle [radius = 0.2];
\draw [black,fill=] (12.1244,25) circle [radius = 0.2];
\draw [black,fill=] (12.1244,31) circle [radius = 0.2];
\draw [black,fill=] (16.4545,26.5) circle [radius = 0.2];
\draw [black,fill=] (10.3923,32) circle [radius = 0.2];
\draw [black,fill=] (11.2583,31.5) circle [radius = 0.2];
\draw [black,fill=] (12.9904,23.5) circle [radius = 0.2];
\draw [black,fill=] (14.7224,26.5) circle [radius = 0.2];
\draw [black,fill=] (16.4545,30.5) circle [radius = 0.2];
\draw [black,fill=] (12.9904,26.5) circle [radius = 0.2];
\draw [black,fill=] (10.3923,27) circle [radius = 0.2];
\draw [black,fill=] (12.1244,23) circle [radius = 0.2];
\draw [black,fill=] (9.52628,30.5) circle [radius = 0.2];
\draw [black,fill=] (12.1244,32) circle [radius = 0.2];
\draw [black,fill=] (13.8564,30) circle [radius = 0.2];
\draw [black,fill=] (17.3205,32) circle [radius = 0.2];
\draw [black,fill=] (14.7224,30.5) circle [radius = 0.2];
\draw [black,fill=] (12.9904,29.5) circle [radius = 0.2];
\draw [gray] (24.2487,15) -- (24.2487,14);
\draw [gray] (25.1147,16.5) -- (25.1147,15.5);
\draw [gray] (25.9808,18) -- (25.9808,17);
\draw [gray] (26.8468,19.5) -- (26.8468,18.5);
\draw [gray] (27.7128,21) -- (27.7128,20);
\draw [gray] (28.5788,22.5) -- (28.5788,21.5);
\draw [gray] (29.4449,24) -- (29.4449,23);
\draw [gray] (23.3827,13.5) -- (24.2487,14);
\draw [gray] (24.2487,15) -- (25.1147,15.5);
\draw [gray] (25.1147,16.5) -- (25.9808,17);
\draw [gray] (25.9808,18) -- (26.8468,18.5);
\draw [gray] (26.8468,19.5) -- (27.7128,20);
\draw [gray] (27.7128,21) -- (28.5788,21.5);
\draw [gray] (28.5788,22.5) -- (29.4449,23);
\draw [red,fill=] (23.3827,13.5) circle [radius = 0.2];
\draw [red,fill=] (24.2487,14) circle [radius = 0.2];
\draw [red,fill=] (24.2487,15) circle [radius = 0.2];
\draw [red,fill=] (25.1147,15.5) circle [radius = 0.2];
\draw [red,fill=] (25.1147,16.5) circle [radius = 0.2];
\draw [red,fill=] (25.9808,17) circle [radius = 0.2];
\draw [red,fill=] (25.9808,18) circle [radius = 0.2];
\draw [red,fill=] (26.8468,18.5) circle [radius = 0.2];
\draw [red,fill=] (26.8468,19.5) circle [radius = 0.2];
\draw [red,fill=] (27.7128,20) circle [radius = 0.2];
\draw [red,fill=] (27.7128,21) circle [radius = 0.2];
\draw [red,fill=] (28.5788,21.5) circle [radius = 0.2];
\draw [red,fill=] (28.5788,22.5) circle [radius = 0.2];
\draw [red,fill=] (29.4449,23) circle [radius = 0.2];
\draw [red,fill=] (29.4449,24) circle [radius = 0.2];
\end{tikzpicture}

%% file: SWARM_submission.bbl
\begin{thebibliography}{10}

\bibitem{AndresArroyo2017}
M.~Andr{\'{e}}s~Arroyo, S.~Cannon, J.~J. Daymude, D.~Randall, and A.~W. Richa.
\newblock A stochastic approach to shortcut bridging in programmable matter.
\newblock Submitted to the 23rd International Conference on {DNA} Computing and
  Molecular Programming ({DNA}23).

\bibitem{Angluin2006}
D.~Angluin, J.~Aspnes, Z.~Diamadi, M.~J. Fischer, and R.~Peralta.
\newblock Computation in networks of passively mobile finite-state sensors.
\newblock {\em Distributed Computing}, 18(4):235--253, 2006.

\bibitem{Berg1983}
H.~Berg.
\newblock {\em Random walks in Biology}.
\newblock Princeton, 1983.

\bibitem{Cannon2016}
S.~Cannon, J.~J. Daymude, D.~Randall, and A.~W. Richa.
\newblock A {M}arkov chain algorithm for compression in self-organizing
  particle systems.
\newblock In {\em Proc. of the 2016 {ACM} Symposium on Principles of
  Distributed Computing (PODC '16)}, pages 279--288, 2016.

\bibitem{Chazelle2009}
B.~Chazelle.
\newblock Natural algorithms.
\newblock In {\em Proceedings of the 2009 {ACM-SIAM} Symposium on Discrete
  Algorithms ({SODA}09)}, pages 422--431, 2009.

\bibitem{Cheung2011}
K.~C. Cheung, E.~D. Demaine, J.~R. Bachrach, and S.~Griffith.
\newblock Programmable assembly with universally foldable strings (moteins).
\newblock {\em {IEEE} Transactions on Robotics}, 27(4):718--729, 2011.

\bibitem{Cieliebak2012}
M.~Cieliebak, P.~Flocchini, G.~Prencipe, and N.~Santoro.
\newblock Distributed computing by mobile robots: Gathering.
\newblock {\em {SIAM} Journal on Computing}, 41(4):829--879, 2012.

\bibitem{Derakhshandeh2015}
Z.~Derakhshandeh, R.~Gmyr, T.~Strothmann, R.~A. Bazzi, A.~W. Richa, and
  C.~Scheideler.
\newblock Leader election and shape formation with self-organizing programmable
  matter.
\newblock In {\em Proc. of the 21st International Conference on {DNA} Computing
  and Molecular Programming (DNA '15)}, pages 117--132, 2015.

\bibitem{Hastings1970}
W.~K. Hastings.
\newblock Monte carlo sampling methods using {M}arkov chains and their
  applications.
\newblock {\em Biometrika}, 57(1):97--109, 1970.

\bibitem{Dauchot2017}
G.~Junot, G.~Briand, R.~Ledesma-Alonso, and O.~Dauchot.
\newblock Active vs. passive hard disks against a membrane : Mechanical
  pressure and instability.
\newblock {\em arXiv}, 2017.

\bibitem{lynch96}
N.~Lynch.
\newblock {\em Distributed Algorithms}.
\newblock Morgan Kauffman, 1996.

\bibitem{Metropolis1953}
N.~Metropolis, A.~W. Rosenbluth, M.~N. Rosenbluth, A.~H. Teller, and E.~Teller.
\newblock Equation of state calculations by fast computing machines.
\newblock {\em Journal of Chemical Physics}, 21:1087--1092, 1953.

\bibitem{Mlot2011}
N.~J. Mlot, C.~A. Tovey, and D.~L. Hu.
\newblock Fire ants self-assemble into waterproof rafts to survive floods.
\newblock {\em Proceedings of the National Academy of Sciences},
  108(19):7669--7673, 2011.

\bibitem{Ramaswamy2010}
S.~Ramaswamy.
\newblock The mechanics and statistics of active matter.
\newblock {\em Annual Review of Condensed Matter Physics}, 1:323--345, 2010.

\bibitem{Reid2015}
C.~R. Reid, M.~J. Lutz, S.~Powell, A.~B. Kao, I.~D. Couzin, and S.~Garnier.
\newblock Army ants dynamically adjust living bridges in response to a
  cost--benefit trade-off.
\newblock {\em Proceedings of the National Academy of Sciences},
  112(49):15113--15118, 2015.

\bibitem{Rubenstein2014}
M.~Rubenstein, A.~Cornejo, and R.~Nagpal.
\newblock Programmable self-assembly in a thousand-robot swarm.
\newblock {\em Science}, 345(6198):795--799, 2014.

\bibitem{Kardar2015}
A.~P. Solon, J.~Stenhammar, R.~Wittkowski, and M~Kardar.
\newblock Pressure and phase equilibria in interacting active brownian spheres.
\newblock {\em Phys. Rev. Lett}, 114(19), 2015.

\bibitem{Tarantino2014}
N.~Tarantino, J.-Y. Tinevez, E.F. Crowell, B.~Boisson, R.~Henriques,
  M.~Mhlanga, F.~Agou, A.~Isra{\"e}l, and E.~Laplantine.
\newblock {TNF} and {IL-}1 exhibit distinct ubiquitin requirements for inducing
  {NEMO}{\textendash}{IKK} supramolecular structures.
\newblock {\em The Journal of Cell Biology}, 204(2):231--245, 2014.

\bibitem{Woods2013}
D.~Woods.
\newblock Intrinsic universality and the computational power of self-assembly.
\newblock In {\em Proceedings of Machines, Computations and Universality 2013
  ({MCU} '13)}, pages 16--22, 2013.

\bibitem{Woods2013-nubot}
D.~Woods, H.-L. Chen, S.~Goodfriend, N.~Dabby, E.~Winfree, and P.~Yin.
\newblock Active self-assembly of algorithmic shapes and patterns in
  polylogarithmic time.
\newblock In {\em Proceedings of the 4th Innovations in Theoretical Computer
  Science Conference (ITCS '13)}, pages 353--354, 2013.

\bibitem{Yim2007}
M.~Yim, W.-M. Shen, B.~Salemi, D.~Rus, M.~Moll, H.~Lipson, E.~Klavins, and
  G.~S. Chirikjian.
\newblock Modular self-reconfigurable robot systems.
\newblock {\em IEEE Robotics Automation Magazine}, 14(1):43--52, 2007.

\end{thebibliography}
